\documentclass[letterpaper]{article} %
\usepackage{aaai2026} 

\usepackage{times}  %
\usepackage{helvet}  %
\usepackage{courier}  %
\usepackage[hyphens]{url}  %
\usepackage{graphicx} %
\usepackage{natbib}  %
\usepackage{caption} %
\usepackage{newfloat}
\usepackage{listings}
\DeclareCaptionStyle{ruled}{labelfont=normalfont,labelsep=colon,strut=off} %
\usepackage[table,svgnames,x11names]{xcolor}
\usepackage[utf8]{inputenc} %
\usepackage[T1]{fontenc}    %
\usepackage{url}            %
\usepackage{booktabs}       %
\usepackage{amsfonts}       %
\usepackage{nicefrac}       %
\usepackage{microtype}      %
\usepackage{enumitem}

\usepackage{amsmath}
\usepackage{amssymb}
\usepackage{mathtools}
\usepackage{amsthm}

\usepackage{algorithmic}
\usepackage[linesnumbered,ruled,vlined]{algorithm2e}
\usepackage[linesnumbered,ruled,vlined]{algorithm2e}

\SetCommentSty{mycommfont}
\SetKwInput{KwInput}{Input}                %
\SetKwInput{KwOutput}{Output}              %
\SetKwInput{KwRequire}{Require}              %
\usepackage{array}
\usepackage{soul} %
\usepackage{multirow}
\usepackage{xtab,rotating}
\usepackage{subfig}
\usepackage[export]{adjustbox}
\usepackage[final]{changes} %
\usepackage{minitoc}
\usepackage{xspace}

\definecolor{rui_colour}{RGB}{140, 197, 250}

\definecolor{stelios_colour}{RGB}{200, 238, 200}

\newcolumntype{s}{>{\columncolor{Lavender}} c}
\newcolumntype{d}{>{\columncolor{Thistle3}} c}
\newcolumntype{f}{>{\columncolor{LightPink1}} c}

\newlength\mylenin
\newcommand\myinput[1]{%
\settowidth\mylenin{\KwIn{}}%
\setlength\hangindent{\mylenin}%
\hspace*{\mylenin}#1\\}

\let\oldnl\nl%
\newcommand{\nonl}{\renewcommand{\nl}{\let\nl\oldnl}}%

\SetKwComment{Comment}{$\triangleright$\ }{}

\newlength\mylenout

\usepackage[capitalize,noabbrev]{cleveref}

\usepackage{multirow}

\newcommand{\sysname}{\textit{HierarchicalPrune}\xspace}

\theoremstyle{plain}

\theoremstyle{definition}

\theoremstyle{remark}

\title{HierarchicalPrune: Position-Aware Compression for Large-Scale Diffusion Models}
\author{
Young D. Kwon$^{1*}$, Rui Li$^{1*}$ \\Sijia Li$^{2}$, Da Li$^{1}$, Sourav Bhattacharya$^{1}$, Stylianos I. Venieris$^{1}$\\
}

\affiliations{

    $^1$ Samsung AI Center-Cambridge, UK \\
    $^2$Independent Researcher\\
    \{yd.kwon, rui.li, s.venieris\}@samsung.com
}

\begin{document}

\maketitle

\renewcommand*{\thefootnote}{\fnsymbol{footnote}}
\footnotetext{*Co-first authors}
\renewcommand*{\thefootnote}{\arabic{footnote}}

\begin{abstract}

State-of-the-art text-to-image diffusion models (DMs) achieve remarkable quality, yet their massive parameter scale (8-11B) poses significant challenges for inferences on resource-constrained devices. In this paper, we present \sysname, a novel compression framework grounded in a key observation: DM blocks exhibit distinct functional hierarchies, where early blocks establish semantic structures while later blocks handle texture refinements. \sysname synergistically combines three techniques: (1) Hierarchical Position Pruning, which identifies and removes less essential later blocks based on position hierarchy; (2) Positional Weight Preservation, which systematically protects early model portions that are essential for semantic structural integrity; and (3) Sensitivity-Guided Distillation, which adjusts knowledge-transfer intensity based on our discovery of block-wise sensitivity variations. As a result, our framework brings billion-scale diffusion models into a range more suitable for on-device inference, while preserving the quality of the output images. Specifically, combined with INT4 weight quantisation, \sysname achieves 77.5-80.4\% memory footprint reduction (\textit{e.g.}, from 15.8 GB to 3.2 GB) and 27.9-38.0\% latency reduction, measured on server and consumer grade GPUs, with the minimum drop of 2.6\% in GenEval score and 7\% in HPSv2 score compared to the original model. Finally, our comprehensive user study with 85 participants demonstrates that \sysname maintains perceptual quality comparable to the original model while significantly outperforming prior works.

\end{abstract}

\section{Introduction}\label{sec:intro}

Diffusion-based text-to-image (T2I) synthesis~\citep{2021ICLRPxTIG12RRHS_SDE,stability_ai_sdxl_turbo,2022arXiv221002747L_FlowMatching,2022arXiv220600364K_karras,2022arXiv221209748P_dit,2024arXiv240212376L_fit} has emerged as a powerful tool for a wide variety of applications, such as the generation of educational content, creative artwork, and UX/UI design prototyping, intensifying demand for deployable billion-parameter diffusion models (DMs). While recent advances, such as Stable Diffusion 3.5 (SD3.5)~\citep{esser2024mmdit} and FLUX~\citep{blackforestlabs_flux1}, significantly outperform previous generations (SDXL~\citep{2023arXiv230701952P_sdxl}, SD1.5~\citep{2022CVPRLDM}, and DALLE-2) in image quality and text alignment, they also come with excessive model sizes (8-11B parameters) and compute demands, limiting the accessibility of such advanced models.

At the same time, despite quantitative metrics, \textit{e.g.},~GenEval~\citep{ghosh2023geneval} and DPG-Bench~\citep{hu2024ellaequipdiffusionmodels}, suggesting smaller models with approx. 2B parameters like SANA-Sprint~\citep{chen2025sanasprintonestepdiffusioncontinuoustime} perform better than their large-scale counterparts, user-based evaluations on the Artificial Analysis Leaderboard\footnote{\url{https://artificialanalysis.ai/image/leaderboard/text-to-image}} reveal a significant gap in the \textit{perceived quality} between compact models and their larger counterparts, which quantitative metrics fail to capture. As such, there is a longstanding need for deploying large models even in resource-constrained settings in order to provide users with high-quality T2I capabilities.

\begin{figure}[t]
\centering
\includegraphics[trim={0 0.5cm 0 0},clip, width=1.01\linewidth]{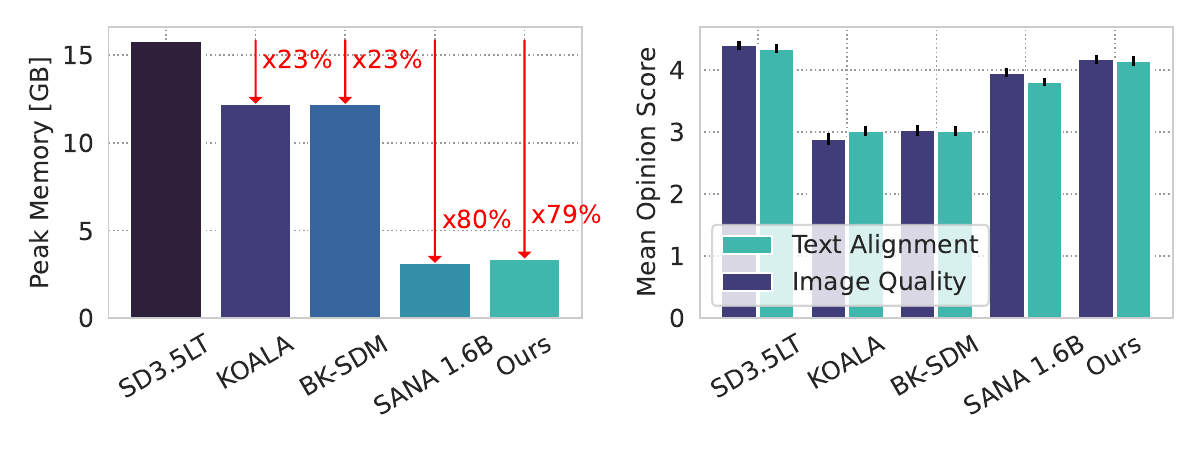}
\caption{
\sysname achieves 79.5\% memory reduction (left) while maintaining image quality. User study with 85 participants (right) demonstrates minimal quality drop (4.8-5.3\%) with 95\% confidence intervals, contrary to the excessive degradation (11.1-52.2\%) of prior methods.
}
\label{fig:teaser}
\end{figure}

\newcommand\imgintroA{\adjustbox{valign=m,vspace=0pt,margin=0pt}{\includegraphics[width=.33\linewidth]{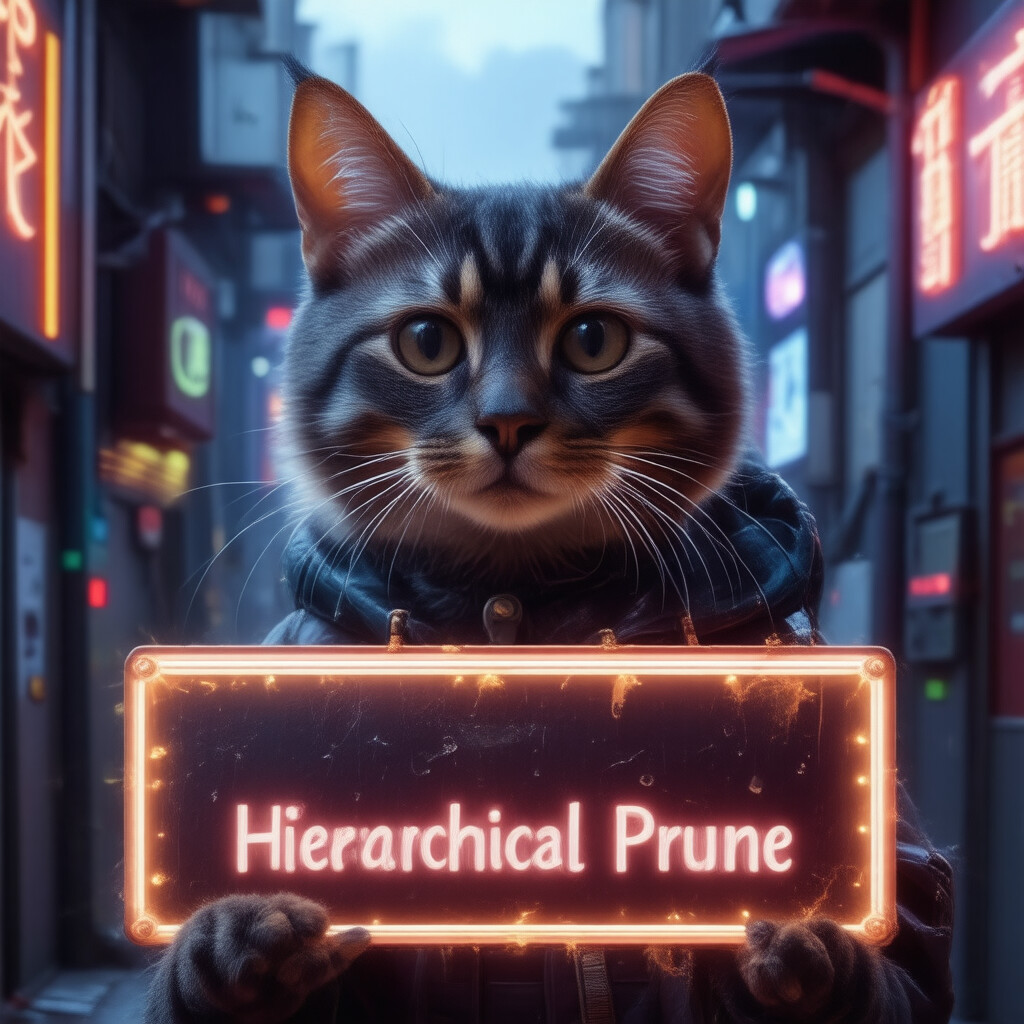}}}
\newcommand\imgintroB{\adjustbox{valign=m,vspace=0pt,margin=0pt}{\includegraphics[width=.33\linewidth]{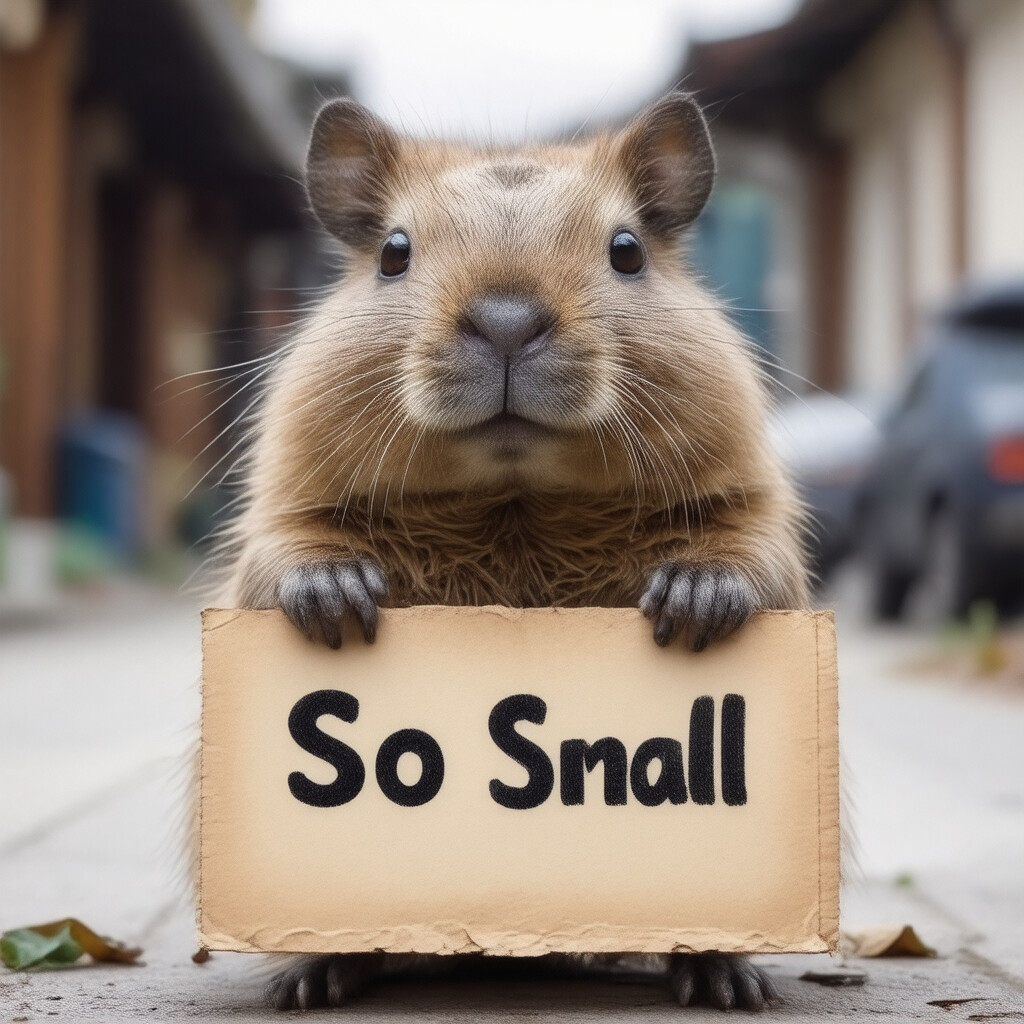}}}
\newcommand\imgintroC{\adjustbox{valign=m,vspace=0pt,margin=0pt}{\includegraphics[width=.33\linewidth]{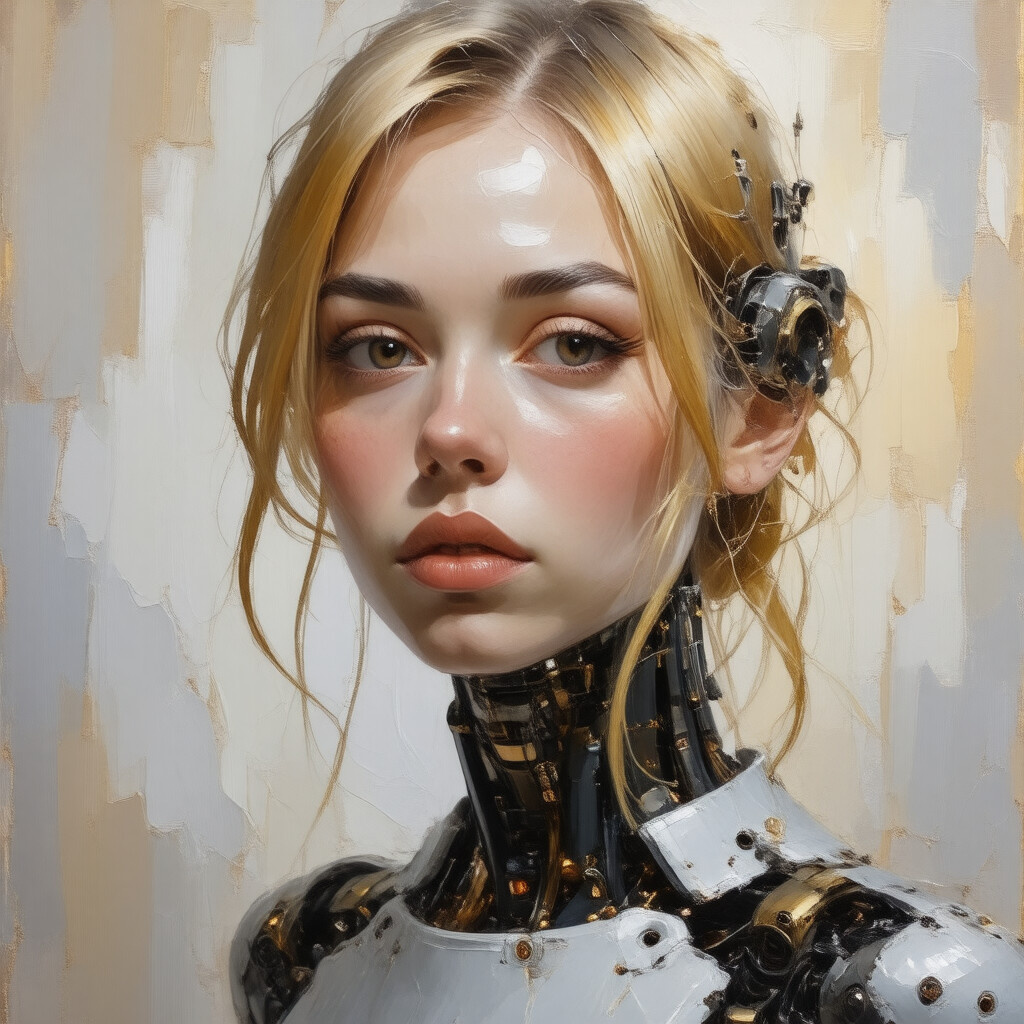}}}
\newcommand\imgintroD{\adjustbox{valign=m,vspace=0pt,margin=0pt}{\includegraphics[width=.33\linewidth]{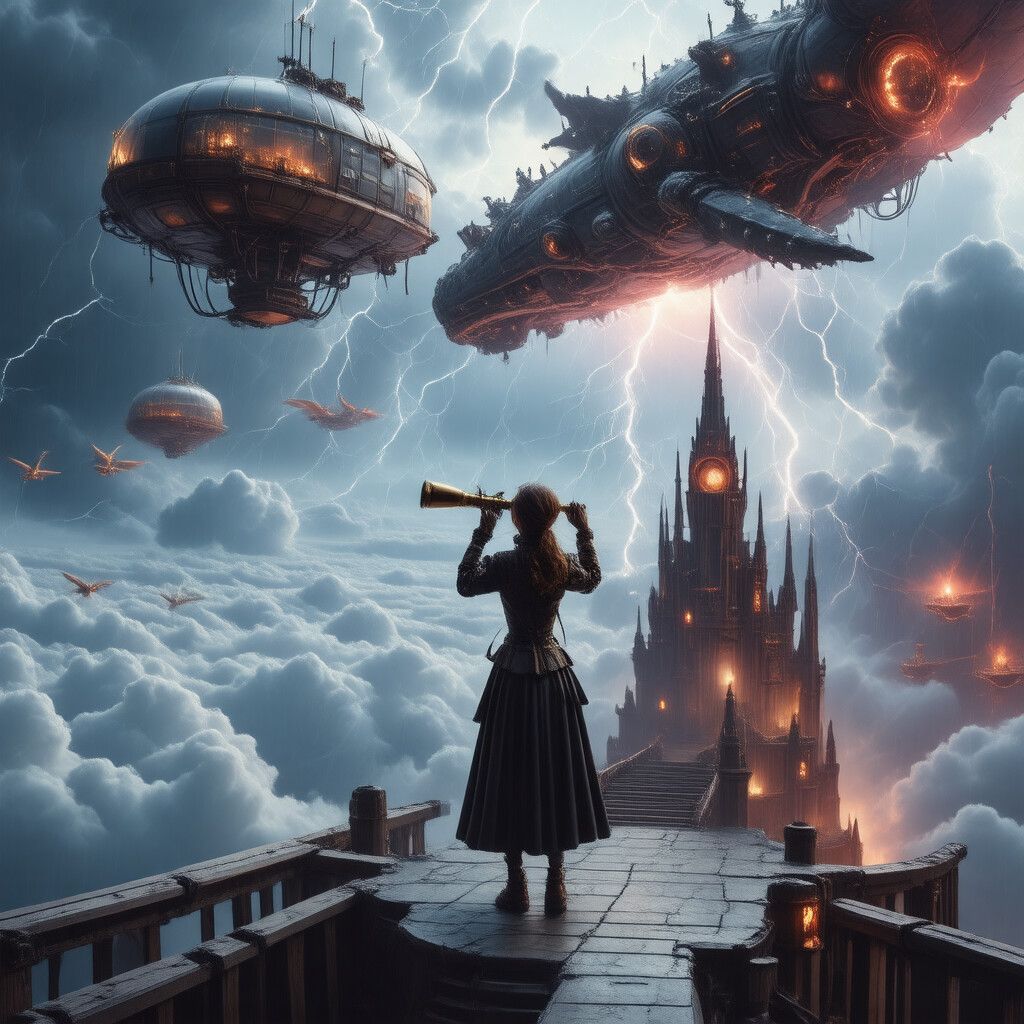}}}

\newcommand\imgintroAA{\adjustbox{valign=m,vspace=0pt,margin=0pt}{\includegraphics[width=.33\linewidth]{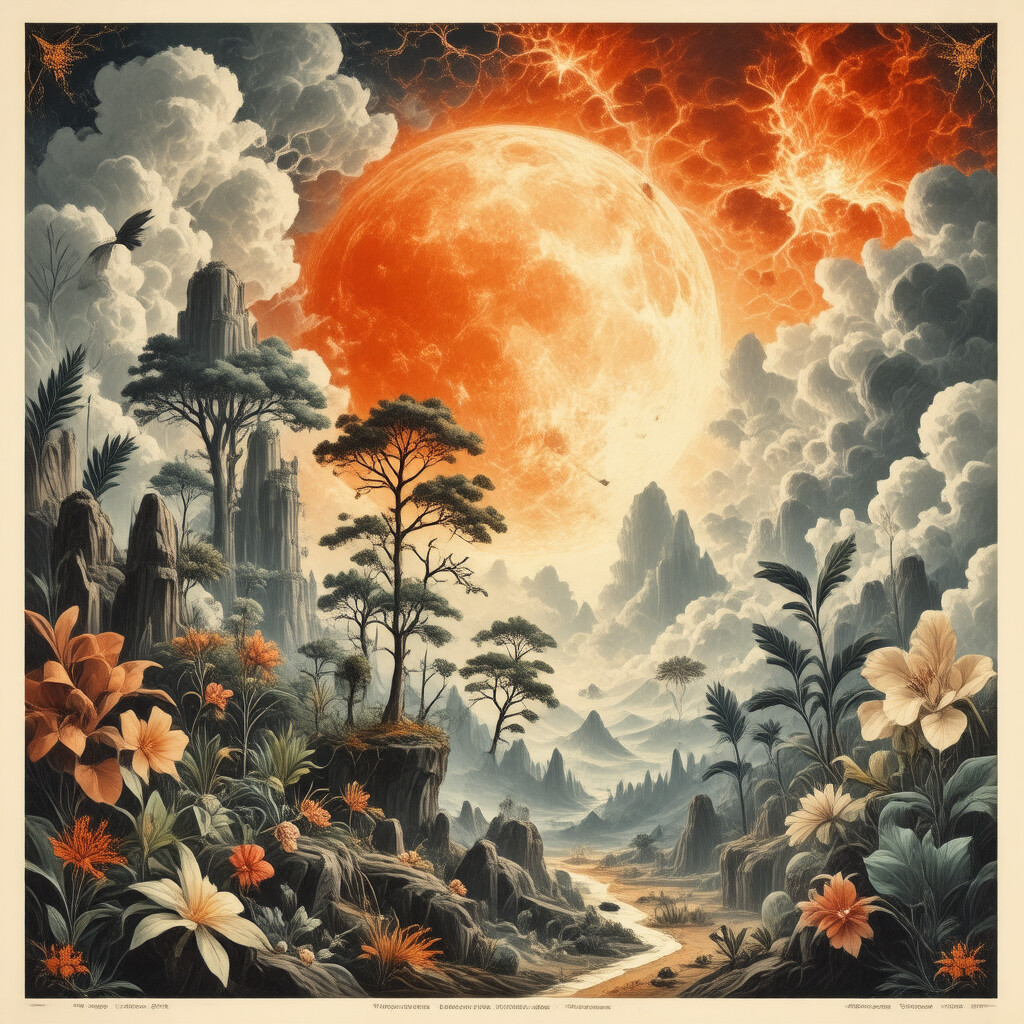}}}
\newcommand\imgintroBB{\adjustbox{valign=m,vspace=0pt,margin=0pt}{\includegraphics[width=.33\linewidth]{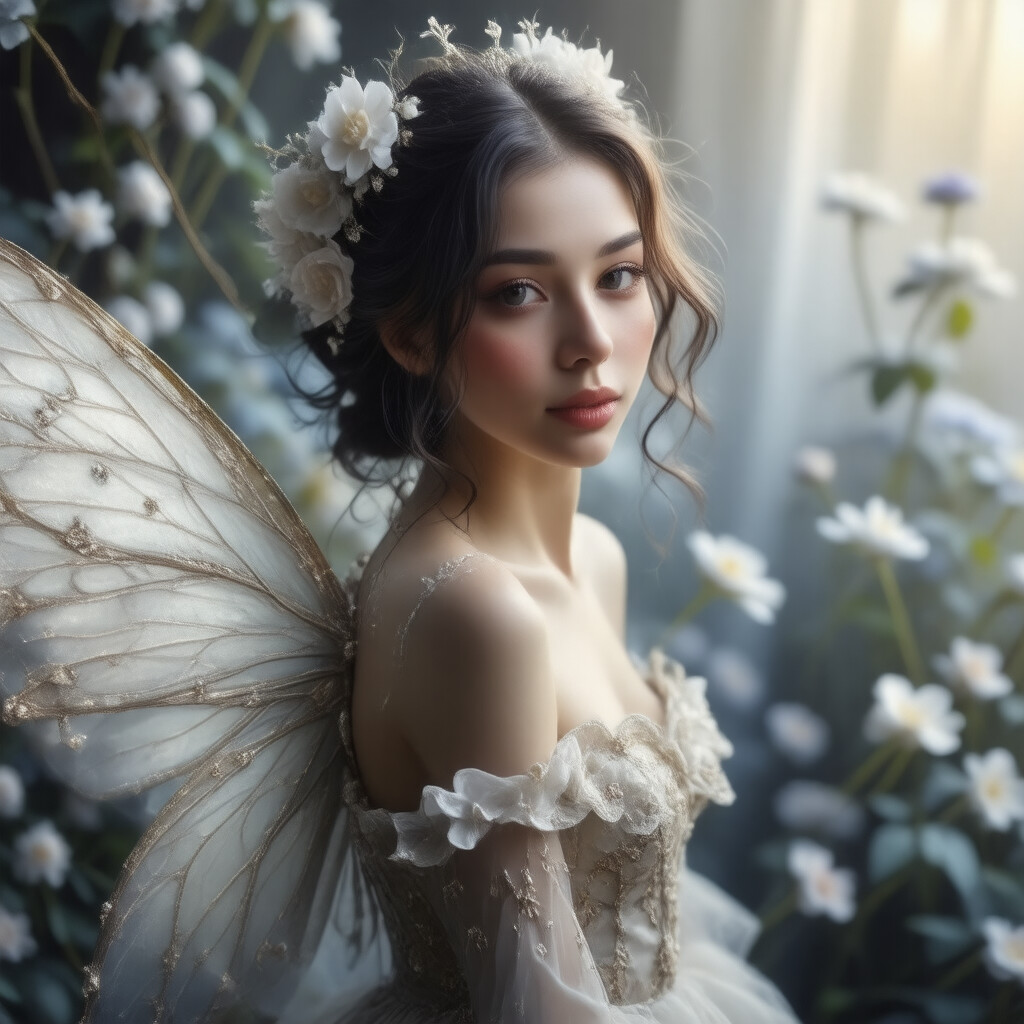}}}
\newcommand\imgintroCC{\adjustbox{valign=m,vspace=0pt,margin=0pt}{\includegraphics[width=.33\linewidth]{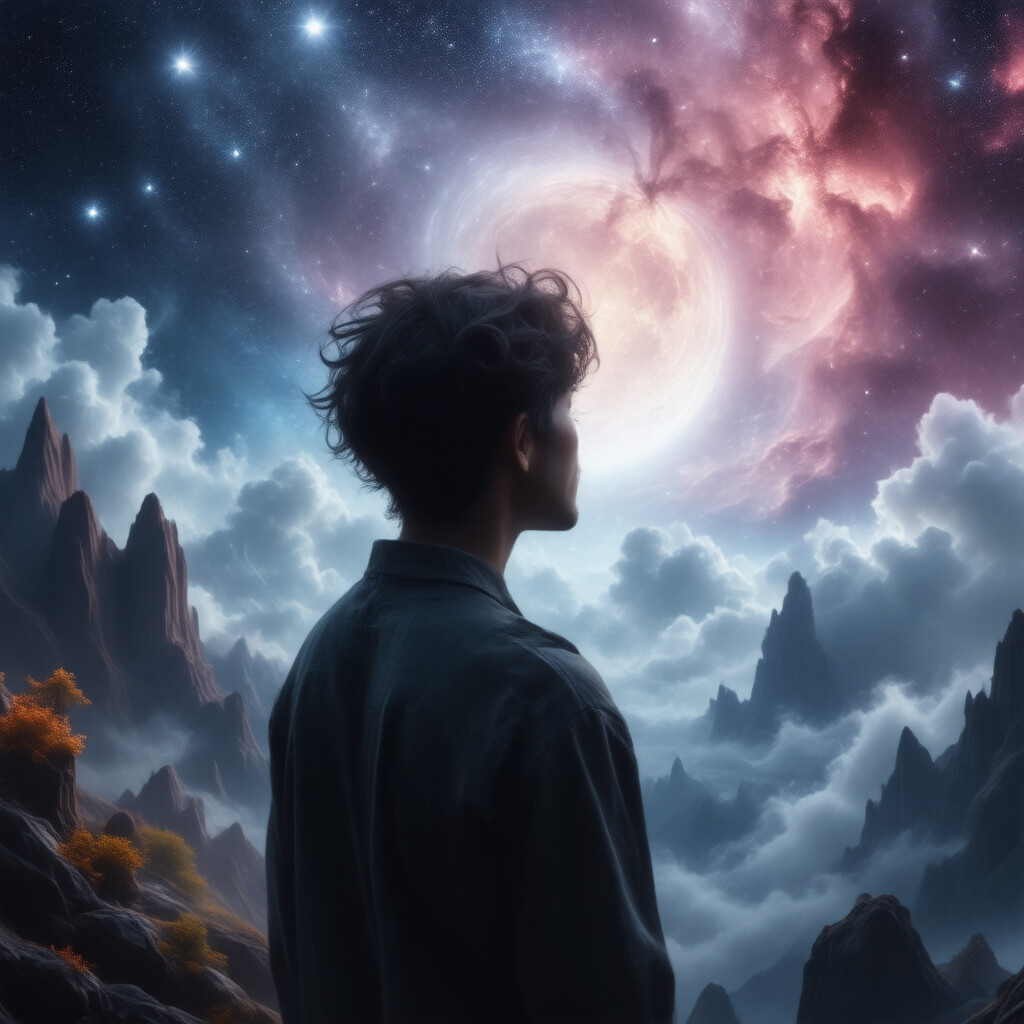}}}
\newcommand\imgintroDD{\adjustbox{valign=m,vspace=0pt,margin=0pt}{\includegraphics[width=.33\linewidth]{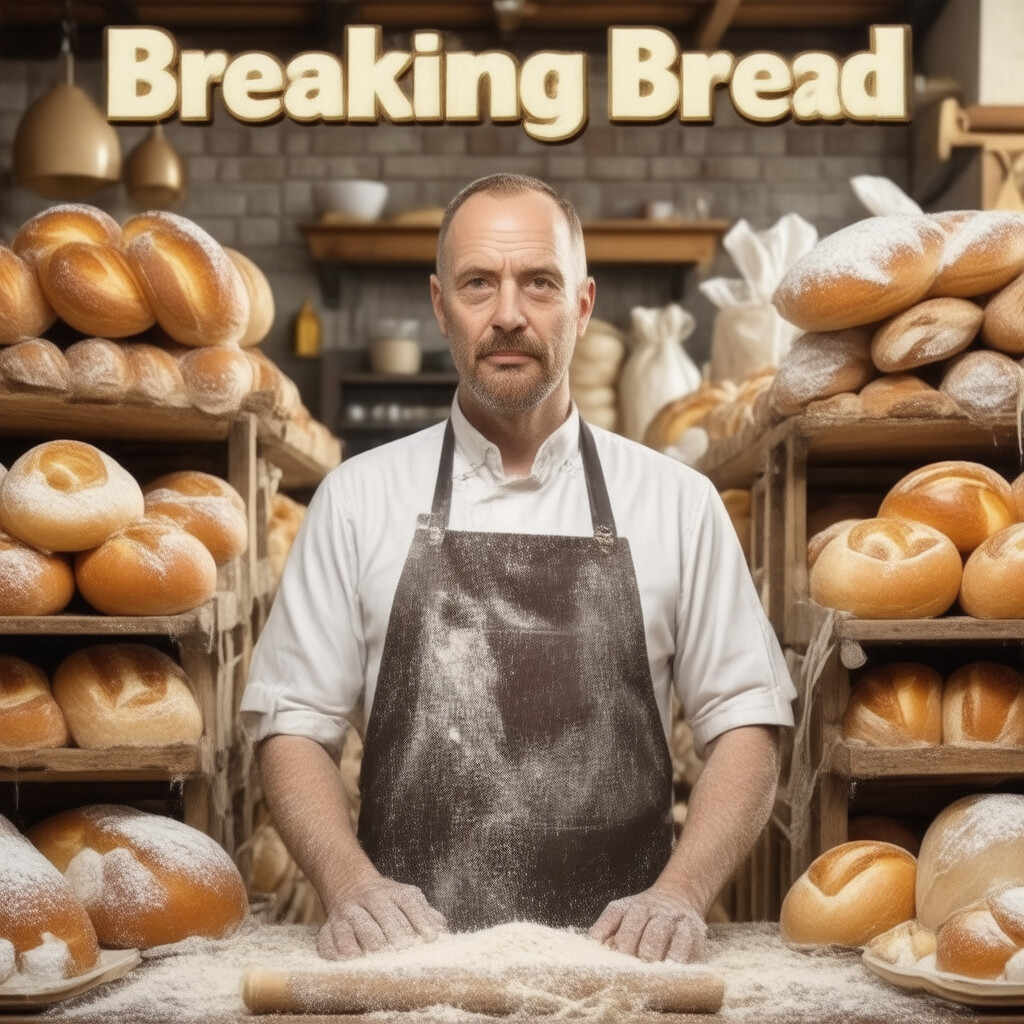}}}

\begin{figure*}[h!]
    \centering
    \resizebox{1.01\linewidth}{!}{%
    \begin{tabular}{@{}c @{\hspace{0.05cm}}c @{\hspace{0.05cm}}c @{\hspace{0.05cm}}c@{}}
        \imgintroA & \imgintroB & \imgintroC &
        \imgintroD \vspace{0.1cm} \\
        \imgintroAA & \imgintroBB & \imgintroCC &
        \imgintroDD \vspace{0.1cm} \\
    \end{tabular}
    }
    \caption{High-resolution image samples generated by compressed model using \sysname, showcasing its superior visual quality across various visual styles, precisely following text prompts, and preserving the ability to draw typography.}
    \label{fig:intro}
\end{figure*}

Nonetheless, this endeavour constitutes a very challenging task due to the high memory and compute intensity~\citep{li2025svdquant} of large DMs, thereby often requiring cloud solutions equipped with high-end GPUs and a minimum of 80GB VRAM. Concurrently, most of the renowned recent releases of foundational DMs, including SD3.5~\citep{esser2024mmdit}, FLUX~\citep{blackforestlabs_flux1}, and Seedream~3.0~\citep{2025seedream3}, are built upon multi-modal diffusion Transformer (MMDiT) backbones~\citep{esser2024mmdit} instead of U-Net, revealing new compression opportunities (Section~\ref{subsec:motivation}).
At the moment, existing efforts to improve efficiency face important limitations.
\textbf{Firstly,} \textit{sampling step reduction}~\citep{2023arXiv230600980L_snapfusion,stability_ai_sdxl_turbo,2023arXiv231117042S_add} and \textit{efficient operator design}~\citep{xie2025sana,FlashAttention} are tailored to improving the speed of DMs rather than reducing memory requirements, leaving the important task of DM deployment in memory-constrained devices unresolved.
\textbf{Secondly,} existing depth-pruning methods~\citep{lee2024koala,kim2024bksdm,fang2024tinyfusion} show promising results in both memory and computation reduction~\citep{kim2024bksdm,fang2024tinyfusion}, outperforming width pruning~\citep{fang2023structural,Castells_2024_CVPR_ldpruner}, but face critical scalability challenges. While they achieve reasonable compression on U-Net-based small DMs (2.6B or less), they fail to compress large-scale, state-of-the-art (SOTA) DMs such as SD3.5 Large (8B) and FLUX (11B) without experiencing significant degradation at 20-30\% memory reduction as demonstrated in Table~\ref{tab:quality_metric}. The full-block-removal methods employed therein cannot capture the fine-grained impact of the subcomponents within each block, and more importantly, the different roles of blocks in different positions across the network's hierarchy, leading to excessive performance drop at high pruning ratios.
\textbf{Lastly,} orthogonal to pruning, \textit{reduced-precision computation}~\citep{he2023ptqd,li2025svdquant,2024arXiv240203666W_quest} has been proven effective, but it has not been combined with prior block-removal DM compression methods.

By observing the limitations of prior work, this paper identifies a novel insight for DMs that consist of MMDiT blocks: Such DMs form a two-fold hierarchy, spanning the inter- and intra-block levels. Inter-block hierarchy reflects the contribution of different blocks to disparate aspects of the output image (\textit{e.g.}~semantic structure, finer visual details), which is determined by their position in the overall architecture. Intra-block hierarchy highlights the subcomponents that compose each MMDiT block and their diverse patterns of importance to the overall quality.

\begin{figure}[t!]
\centering
\includegraphics[trim={0 0 0 0},clip, width=0.979\linewidth]{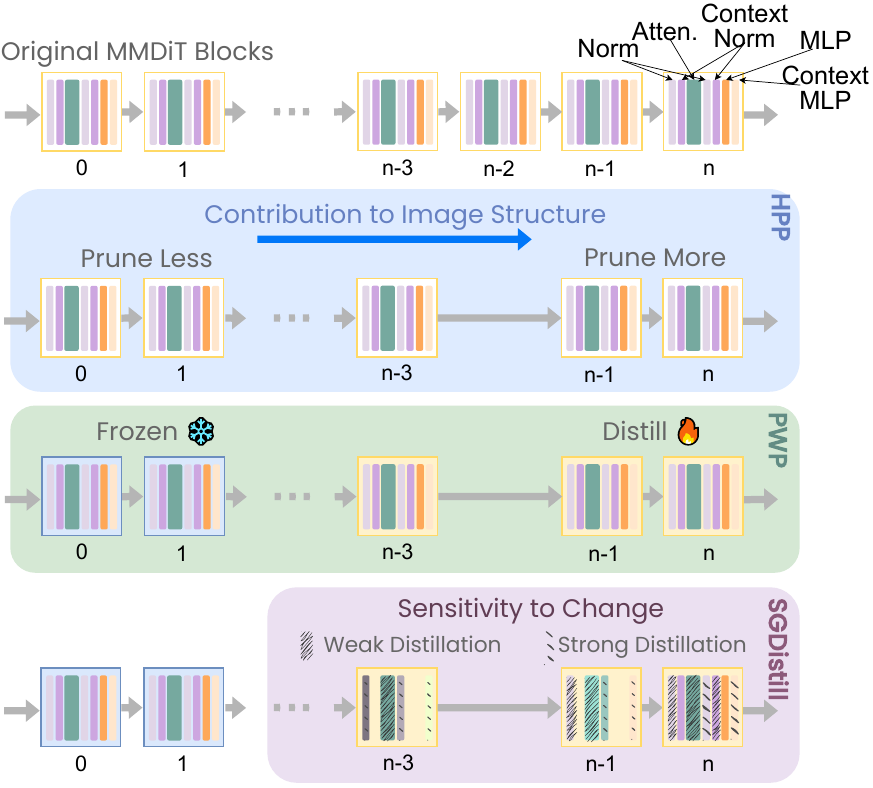}
\caption{
\sysname's compression framework leverages MMDiT's two-fold hierarchy (\emph{inter-block}: early blocks establish semantics, later blocks refine; \emph{intra-block}: varying subcomponent importance). It comprises (1)~Hierarchical Position Pruning (HPP), maintaining early blocks while pruning later ones, (2)~Positional Weight Preservation (PWP), freezing critical early blocks during distillation, and (3)~Sensitivity-Guided Distillation (SGDistill), applying inverse weights—minimal updates to sensitive blocks and subcomponents. The resulting framework enables effective compression while preserving model capabilities.
}
\label{fig:framework}
\vspace{-0.4cm}
\end{figure}

Building on this new viewpoint, we propose \textbf{\sysname}, a principled compression methodology that overcomes the limitations of existing methods using three hierarchy-informed techniques (Fig.~\ref{fig:framework}). First, we introduce \textbf{Hierarchical Position Pruning (HPP)}, a method that leverages our empirical insight that later MMDiT blocks contribute less to fundamental image structure, by strategically maintaining early blocks that form core image structures while pruning later blocks that primarily handle refinements. Second, we incorporate \textbf{Positional Weight Preservation (PWP)}, which freezes the non-pruned and earlier portions of the model during the distillation process. This
approach maintains the integrity of early blocks, which are essential for image formation, while allowing later, less critical blocks to be updated. Finally, we propose \textbf{Sensitivity-Guided Distillation (SGDistill)}, which operates with a counterintuitive yet effective principle: blocks with higher importance are also more sensitive to change. Our analysis reveals that \textit{in aggressive pruning settings, attempting to update these highly important blocks often proves detrimental to model performance.} As such, we enforce inverse distillation weights—assigning minimal or zero update weights to the most important blocks, while concentrating updates on less sensitive components.

We extensively benchmark \sysname on both server and desktop-grade GPUs, demonstrating superior results over SOTA methods. Our contributions include:

\newlist{myitemize}{itemize}{5}
\setlist[myitemize,1]{label=\textbullet,leftmargin=5mm}
\setlist[myitemize,2]{label=$\rightarrow$,leftmargin=1em}
\setlist[myitemize,3]{label=$\diamond$}

\begin{myitemize}
    \item We identify a dual hierarchical structure in MMDiT DMs: an inter-block hierarchy (earlier blocks establish semantics, later blocks handle refinements) and an intra-block hierarchy (varying importance patterns of subcomponents within each MMDiT block).
    \item \sysname establishes a comprehensive, position-aware pruning and distillation framework for large-scale DMs, for the first time, by combining HPP, PWP, and SGDistill with INT4 quantisation that achieves 77.5-80.4\% memory reduction with minimal quality loss (3.2-4.8\% ours vs. 15.3-41.2\% degradation for prior works).
    \item An extensive user study with 85 participants demonstrates that \sysname significantly outperforms all the baselines (Fig.~\ref{fig:teaser}): KOALA and BK-SDM suffer from a substantial 44.0-52.2\% user-perceived quality degradation, and the SOTA small-scale DM, namely SANA-Sprint-1.6B, shows a 11.1-14.2\% drop. In contrast, \sysname observes a mere 4.8-5.3\% degradation.
\end{myitemize}

\section{Methodology}\label{sec:method}

\subsection{Motivation}
\label{subsec:motivation}

Examining MMDiT-based DMs, we identify a two-fold hierarchy: \textit{i)}~\textit{inter-block} and \textit{ii)}~\textit{intra-block hierarchy}. Inter-block hierarchy refers to the organisation of blocks along the overall architecture and implies a functional hierarchy where blocks have different responsibilities based on their position. Intra-block hierarchy focuses on the fact that individual blocks are internally composed of subcomponents, which in the case of MMDiT consist of Norm, Context Norm, Attention, MLP, and Context MLP modules~\citep{esser2024mmdit}, with each subcomponent having a varying impact on performance based on its type and position.

Given this dual hierarchy, we conjectured that each block is responsible for different aspects of the generated images and that its position in the DM architecture largely determines these aspects. This also holds for subcomponents and is affected by their type. 
Concretely, we hypothesised that the contribution of different blocks to the output image is not uniform, regarding importance to performance and influence on specific image traits (\textit{e.g.},~structure, texture or finer details). 

\begin{figure*}[t!]
  \centering
  \subfloat[Block Removal]{
    \includegraphics[width=0.325\textwidth]{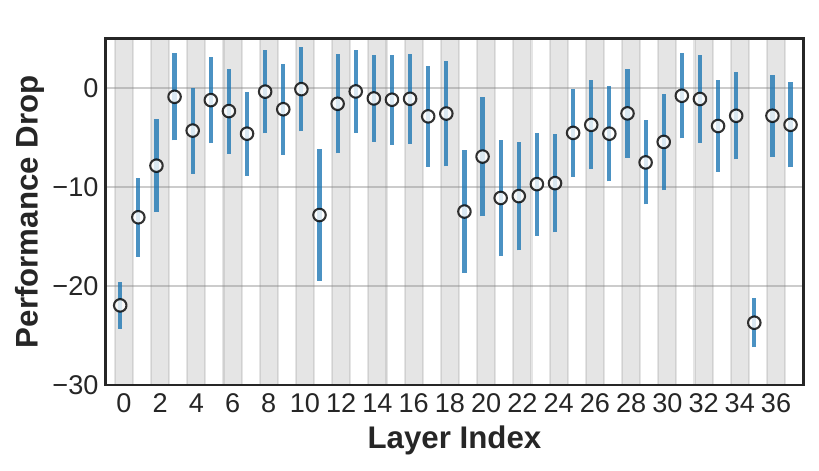}
  \label{subfig:block_removal}}
  \subfloat[Multi-modal Attention]{
    \includegraphics[width=0.325\textwidth]{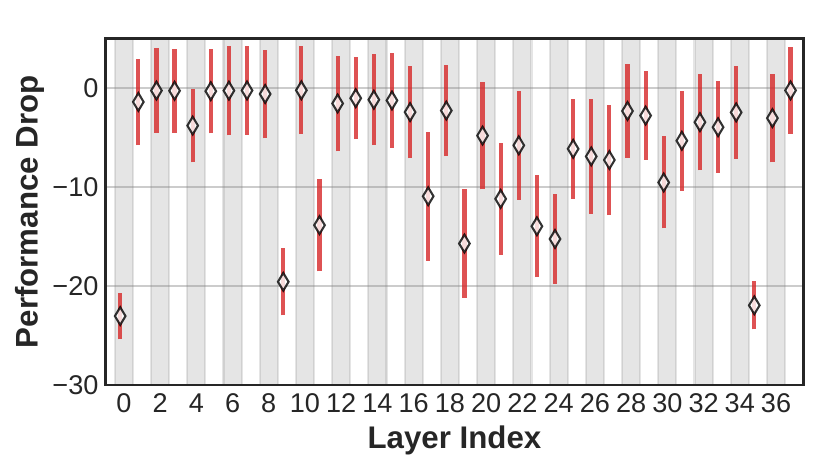}
  \label{subfig:attn_removal}}
  \subfloat[MLP]{
    \includegraphics[width=0.325\textwidth]{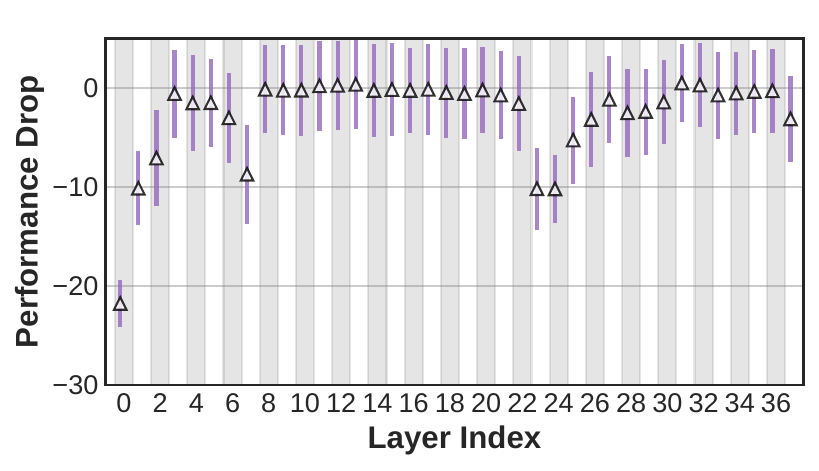}
  \label{subfig:mlp_removal}}
  \hfill
  \caption{
  Fine-grained contribution analysis of SD3.5 Large Turbo on the HPSv2 dataset by removing either an entire MMDiT block (a), following prior depth pruning approaches~\citep{lee2024koala,kim2024bksdm,fang2024tinyfusion}, or an intra-block subcomponent (b, c and see Fig.~\ref{fig:contribution_analysis_full} in Appendix~\ref{app:additional_analysis} for full set of analysis). We report the performance drop compared to the original model. The discrepancy in performance drop patterns reveals the different patterns of importance of each subcomponent. 
  }
  \label{fig:contribution_analysis}
  \vspace{-0.3cm}
\end{figure*}

To investigate this, we conducted a contribution analysis on SD3.5 Large Turbo over the HPSv2 dataset~\citep{Wu_2023_HPS}, by removing both individual MMDiT blocks (Fig.~\ref{subfig:block_removal}) and subcomponents (Fig.~\ref{subfig:attn_removal},~\ref{subfig:mlp_removal} and Fig.~\ref{subfig:attn_removal_supp}--\ref{subfig:norm_context_removal_supp}~in Appendix~\ref{app:additional_analysis}), and comparing the performance before and after. We observe that each of the subcomponents, as well as whole-block removal, demonstrates \textit{different patterns of impact at different locations}. This was further highlighted when analysing the joint removal of multiple subcomponent types, where different subcomponent combinations exhibited significantly different effects on the final performance (see Appendix~\ref{app:additional_analysis}).

\begin{figure}[t]
\centering
\includegraphics[width=1.0\columnwidth, trim={0 0.5cm 0 0},clip]{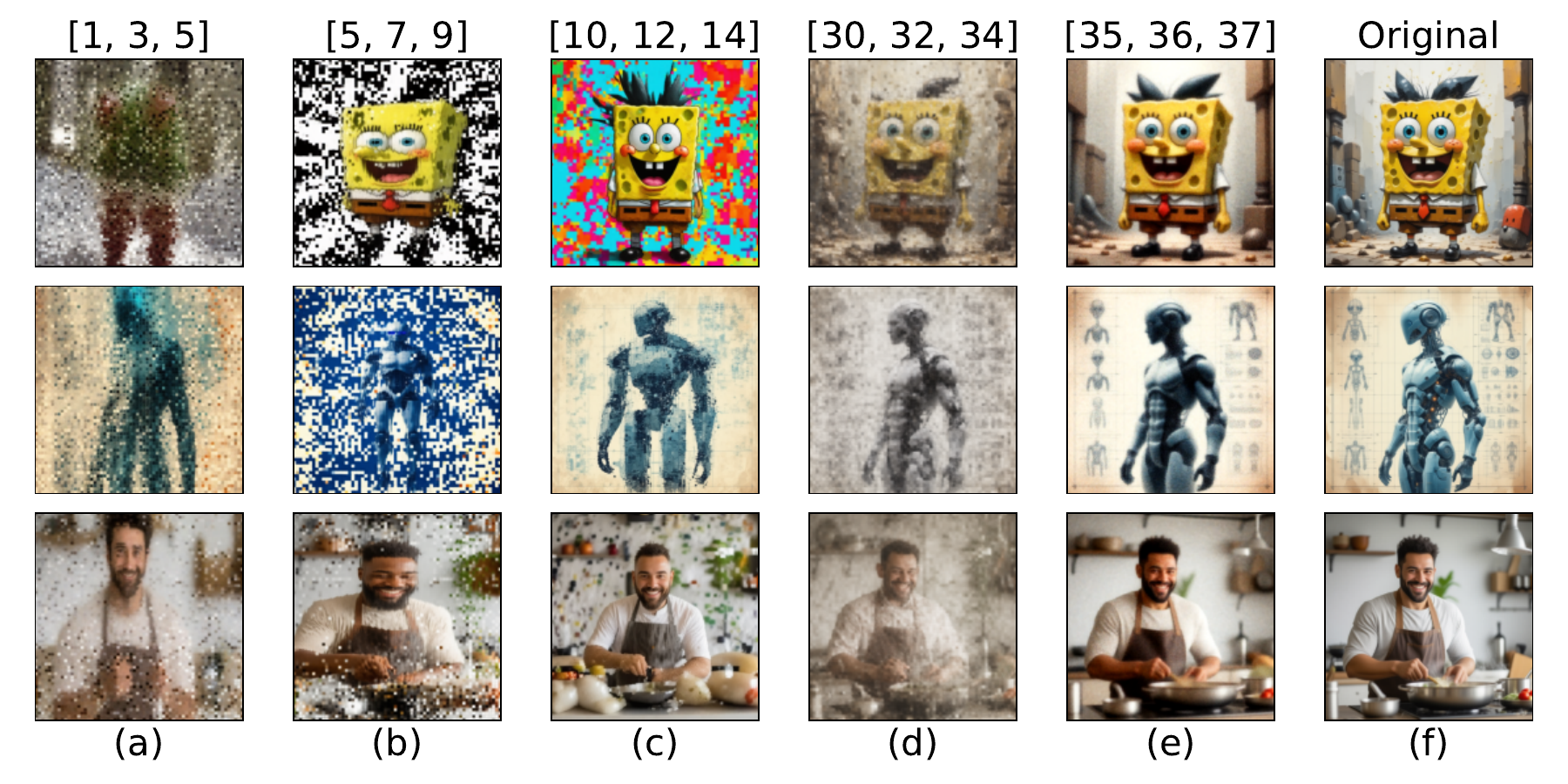}
\caption{Impact of removing MMDiT blocks at different positions. Compared to original outputs (f), removing earlier layers leads to high impact on image structure (a-c), whereas removing later blocks affects mainly fine details (d, e).}
\label{fig:image_rm_layers}
\vspace{-0.4cm}
\end{figure}

We further observe that the performance drop is typically more severe in earlier stages of the network. Given the inter-block hierarchy hypothesis, we attribute this to earlier blocks contributing more to core elements of the output image that affect its quality, such as semantic structure, whereas later blocks are more important for finer visual details. To verify this, we examined the generated images when removing a number of MMDiT blocks at different locations throughout the network. Specifically, we randomly selected a few examples from the HPSv2 dataset and performed T2I generation with SD3.5 Large Turbo, removing three non-consecutive blocks each time. Fig.~\ref{fig:image_rm_layers} shows that the overall image structure changes dramatically when layers are removed before and up to layer 10, while after layer 30, removal of the same number of blocks has minimal impact on the structure, while finer details, such as style, are still affected, providing evidence for the inter-block hierarchy of MMDiT DMs.

Despite this variability, existing full-block removal methods~\citep{lee2024koala,kim2024bksdm} treat MMDiT blocks homogeneously, compromising performance when targeting high compression ratios. Our observations reveal a critical shortcoming: by ignoring the inter-block differences, these methods inadvertently prune blocks that are disproportionately important to visual quality while retaining less impactful ones. Moreover, by coarsely removing whole blocks, existing methods not only discard redundant layers but also eliminate subcomponents that might be essential for capturing fine-grain features. As shown in Section~\ref{subsec:results}, this lack of differentiation leads to a steep drop in model performance under aggressive pruning rates. These insights motivate us to design hierarchy-informed techniques for the effective compression of large MMDiT-based DMs.

\subsection{HierarchicalPrune}

\begin{algorithm}[!t]
\caption{HierarchicalPrune}
\label{alg:hierarchical_prune}
    \setcounter{AlgoLine}{0}
    \small
    \SetAlgoLined
    \LinesNumbered
    \DontPrintSemicolon
    
    \KwIn{DM $m$ with $|\mathcal{B}|$ MMDiT blocks and parameters $\boldsymbol{\theta}$}
    \nonl
    \myinput{Subcomponent set $\mathcal{C}$ of each MMDiT block}
    \nonl
    \myinput{Quality threshold $q$, Compression ratio $r$}
    \nonl
    \myinput{Dataset $\mathcal{D}$ and calibration set $\mathcal{D}_{\text{calib}} \subset \mathcal{D}$}
    \KwOut{Pruned and distilled DM $m'$ with parameters $\boldsymbol{\theta'}$}
    \SetAlgoVlined
    \nonl

    \texttt{/*} - - - \textit{Stage 1 - HPP / PWP} - - - \texttt{*/}\

    $\Delta P(i,c) \leftarrow \text{ContributionAnalysis}(m, \mathcal{D}_{\text{calib}}, i, c)$, \nonl\ $\forall i \in [0, |\mathcal{B}|-1], \forall c \in \mathcal{C}$ \\ \Comment*[f]{\scriptsize \textrm{Estimate importance based on performance drop}}
    
    $Score(i,c) \leftarrow \text{PrunabilityScore}(\Delta P, i, c)$ \nonl\ $\forall i \in [0, |\mathcal{B}|-1], \forall c \in \mathcal{C}$ \\ \Comment*[f]{\scriptsize \textrm{Calculate prunability as per Eq.~(\ref{eq:score}) and (\ref{eq:pos_weight_func})}}
    
    $m_{\text{pruned}}, \boldsymbol{\theta}_{\text{pruned}} \leftarrow \text{ApplyHPP}(m, \boldsymbol{\theta}, Score, r)$ \\ \Comment*[f]{\scriptsize \textrm{Given the prunability per block, remove blocks until compression ratio}}

    \nonl\
    \nonl
    
    \texttt{/*} - - - \textit{Stage 2 - Distillation} - - - \texttt{*/}\

    $m_{\text{pruned,frz}}, \boldsymbol{\theta}_{\text{pruned,frz}} \leftarrow \text{ApplyPWP}(m_{\text{pruned}}, \boldsymbol{\theta}_{\text{pruned}})$ \nonl\ \Comment*[f]{\scriptsize \textrm{Freeze the non-pruned and earlier parts of the model}}

    \If{$r \geq r_\textup{thres}$}{
        $m', \boldsymbol{\theta}' \leftarrow \text{SGDistill}(m'_{\text{pruned,frz}}, \boldsymbol{\theta}'_{\text{pruned,frz}}, \mathcal{D}, q)$ \\ \Comment*[f]{\scriptsize \textrm{If aggressive pruning, use sensitivity-guided distillation}}
    }\Else{
        $m', \boldsymbol{\theta}' \leftarrow \text{Distill}(m'_{\text{pruned,frz}}, \boldsymbol{\theta}'_{\text{pruned,frz}}, \mathcal{D}, q)$ \\ \Comment*[f]{\scriptsize \textrm{If moderate pruning, use normal distillation}}
    }

    \nonl\
    \nonl

    \texttt{/*} - - - \textit{Stage 3 - Post-Training Quantisation} - - - \texttt{*/}\

    $m', \boldsymbol{\theta}' \leftarrow \text{PTQ}(m', \boldsymbol{\theta}', \text{precision=W4A16})$ \\ \Comment*[f]{\scriptsize \textrm{Apply PTQ for further compression}}
    
    \KwRet $m'$, $\boldsymbol{\theta'}$    
\end{algorithm}

Motivated by our findings in Section~\ref{subsec:motivation}, we propose \sysname (Fig.~\ref{fig:framework}), a cohesive, multi-stage pruning and distillation approach that respects the hierarchical nature of DMs. \sysname introduces three key techniques: \textit{i)}~Hierarchical Position Pruning (HPP), \textit{ii)}~Positional Weight Preservation (PWP), and \textit{iii)}~Sensitivity-Guided Distillation (SGDistill). These techniques operate complementarily with the objective to maximise the model compression rate while maintaining output quality throughout the T2I process. Algorithm~\ref{alg:hierarchical_prune} presents the complete methodology. Given a pre-trained model $m$, the first stage (line~1) is responsible for producing a pruned model $m_{\text{pruned}}$ with a compression ratio $r$ using HPP.

In the second stage (lines~2-5), \sysname first prepares the pruned model for distillation, by preserving its most sensitive blocks by means of PWP (line~2), yielding the selectively frozen model $m_{\text{pruned,frz}}$. Then, distillation is applied in order to improve the attainable image quality. Specifically, if the target compression ratio $r$ indicates aggressive compression by surpassing a threshold $r_{\text{thres}}$, \sysname applies SGDistill (line~3), our proposed sensitivity-guided distillation method. For moderate compression, a simpler distillation process is followed (line~5). As a final step, the resulting model is optionally quantised with 4-bit weights and 16-bit activations (line~6) to obtain additional compression.

\textbf{Hierarchical Position Pruning:} At its foundation, \textit{Hierarchical Position Pruning (HPP)} leverages our insight that deeper MMDiT blocks contribute less to core visual structure. As such, HPP strategically targets blocks for removal based on their hierarchical position within the model, maintaining early blocks that form image structures while pruning deeper blocks that mainly handle visual details.

Formally, for a DM with $|\mathcal{B}|$ MMDiT blocks, we first introduce a position weight function $W_{\text{pos}}(i)$ (Eq.~(\ref{eq:pos_weight_func})) that captures the inter-block hierarchy by mapping the more critical earlier blocks to lower values and guiding the pruning process towards the later layers. Next, we calculate a prunability score $Score$ (Eq.~(\ref{eq:score})) for each block in $\mathcal{B}$ and subcomponent(s), consisting of the performance drop from the original model scaled by the position weight function. Concretely: 
\begin{equation}
Score(i, c) = -|\Delta P(i, c)| \times W_\text{pos}(i) \quad  \forall i \in [0, |\mathcal{B}|-1] 
\label{eq:score}
\end{equation}
\begin{equation}
W_\text{pos}(i) = e^{(i - |\mathcal{B}|)/|\mathcal{B}|}
\label{eq:pos_weight_func}
\end{equation}
where $i$ is the block index and $\mathcal{B}$ is the number of blocks, $c \in \mathcal{C}$ is the subcomponent type, $\Delta P(i, c)$ is the importance score quantified as performance drop from the original model when block $i$ and/or subcomponents $c$ are removed, $W_{\text{pos}}(i)$ is the position weight function that favours later layers.

\textbf{Positional Weight Preservation:} Building upon this position-aware foundation, we incorporate \textit{Positional Weight Preservation (PWP)}, which freezes the non-pruned and earlier portions of the model during the distillation process. This straightforward yet effective approach maintains the integrity of early blocks which are essential for image formation, while allowing later, less critical blocks to be updated. By systematically keeping weights static based on their position in the network, PWP significantly outperforms basic position-based pruning, HPP, for moderate pruning scenarios (\textit{e.g.}, 25\% parameter reduction), as it ensures that the most structurally important parts of the model remain intact. Note that even at moderate compression levels, prior methods suffer substantial quality degradation (15.3-41.2\% in Table~\ref{tab:quality_metric}), while our approach maintains near-original performance.

\textbf{Sensitivity-Guided Distillation:} For aggressive parameter reduction (\textit{e.g.},~$\geq$30\%), however, we discovered that even with careful pruning and preservation, excessive block-level pruning leads to unacceptable quality degradation. To address this challenge, we introduce \textit{Sensitivity-Guided Distillation \mbox{(SGDistill)}}, which operates with a counterintuitive yet effective principle: blocks with higher importance are also more sensitive to change.

Our analysis reveals that \textit{in aggressive pruning settings, attempting to update these highly important blocks often proves detrimental to model performance} (showing 31.9\% average quality reduction even with PWP, see Table~\ref{tab:quality_metric_ablation}). Consequently, SGDistill applies inverse distillation weights—assigning minimal or zero update weights to the most important blocks, while concentrating updates on less sensitive components. This approach, combined with subcomponent (\textit{e.g.}, Normalisation and MLP within an MMDiT block) pruning for the remaining portion of the DM after pruning up to $r_{\text{thres}}$ (line~2) with HPP and PWP, extends our hierarchical approach from the block-level to the intra-block subcomponents, preserving the carefully tuned parameters of critical blocks while allowing adaptation in less sensitive regions. 
Our experimental results confirm this hypothesis, showing that protecting sensitive blocks from significant updates during distillation is key to maintaining quality at high compression ratios, reducing average quality degradation from 31.9\% to 10.1\% (Table~\ref{tab:quality_metric_ablation}).

Concretely, the objective of the distillation process of \sysname leverages a combination of the feature loss, $\mathcal{L}_{\text{feat}}$, \textit{i.e.}, a feature distillation loss, and the standard knowledge distillation (KD) loss to minimise the difference between the final output of the compressed (\textit{i.e.},~student) model $\boldsymbol{\theta'}$ and the original (\textit{i.e.},~teacher) model $\boldsymbol{\theta}$, leading to the overall loss $\mathcal{L} = \mathcal{L}_{\text{feat}} + \mathcal{L}_{\text{KD}}$, with
$\mathcal{L}_{\text{KD}}$ expressed as:
\begin{equation}\label{eq:loss_kd}
\mathcal{L}_{\text{KD}} = \mathbb{E} \left[ ||v_{\boldsymbol{\theta'}}(x_t, t) - v_{\boldsymbol{\theta}}(x_t, t) ||^2 \right]
\end{equation}
where $t$ is the time step, and $x_t$ is the noisy diffusion sample, started from the clean latent $x_0$ from the autoencoder (VAE). For the selected set of blocks to update $\mathcal{B^*} \in \mathcal{B}$ each associated with an importance score $\Delta P(i, c)$ (Eq.~(\ref{eq:score})), by denoting feature output of block $i$ in teacher and student models as $f_{\boldsymbol{\theta}}^l$ and $f_{\boldsymbol{\theta'}}^{l'}$, respectively, we have: 
\begin{equation}\label{eq:loss_SGDistil}
\scriptsize
\mathcal{L}_{\text{feat}} = \mathbb{E} \left[ \sum_{i \in [ 0, |\mathcal{B^*}|-1 ] } ||f_{\boldsymbol{\theta'}}^{i'}(x_t, t) - f_{\boldsymbol{\theta}}^i(x_t, t) ||^2 \right]
\end{equation} 
SGDistill specifies that, for each block, we scale the final parameter update by its sensitivity, \textit{i.e.}, $\frac{1}{\Delta P(i, c)}$, regulating in this way the rate of change of each block during distillation.

\section{Evaluation}\label{sec:evaluation}
To evaluate \sysname against existing methods, we conducted both quantitative and qualitative comparisons (Section~\ref{subsec:results}). We also perform an ablation study to assess each proposed component, the impact of quantisation, and the robustness of text-drawing capability (Section~\ref{subsec:ablation}).

\subsection{Experimental Setup}\label{subsec:experimental_setup}

\textbf{Architectures, Datasets, and Implementation:}
We target SD3.5 Large Turbo (8B) and FLUX.1-Schnell (12B), two SOTA models designed to perform diffusion tasks within a small number of steps (\textit{e.g.}, 4). Following prior work~\citep{lee2024koala}, we use the YE-POP dataset~\citep{yepop} consisting of 500K images. We implemented the \sysname pipeline in PyTorch and \texttt{Diffusers}~\citep{von-platen-etal-2022-diffusers}, utilising SD3.5 Large Turbo and FLUX.1-Schnell model checkpoints. For 4-bit weight quantisation, we adopt \texttt{bitsandbytes}~\citep{dettmers2022llmint8}. %

\newcommand\appImgAAA{\adjustbox{valign=m,vspace=0pt,margin=0pt}{\includegraphics[width=.33\linewidth]{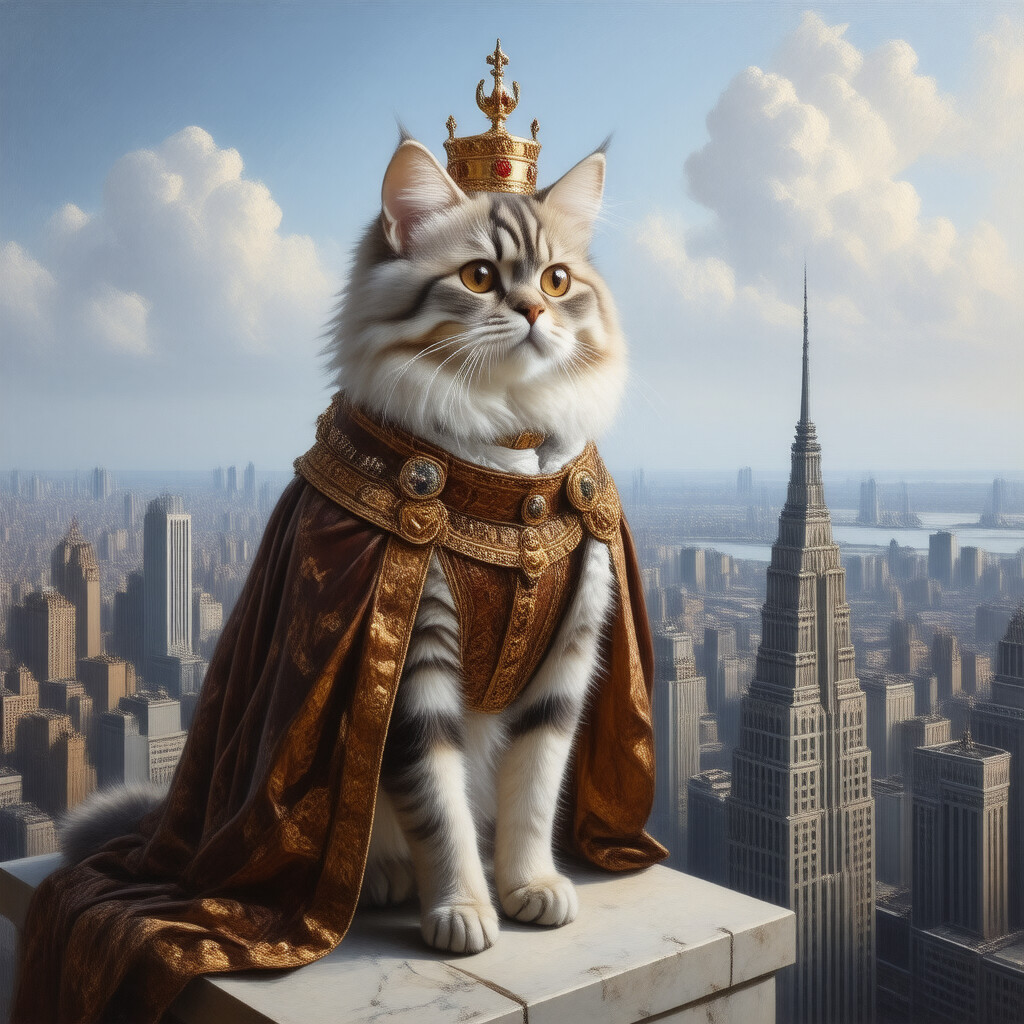}}}
\newcommand\appImgBBB{\adjustbox{valign=m,vspace=0pt,margin=0pt}{\includegraphics[width=.33\linewidth]{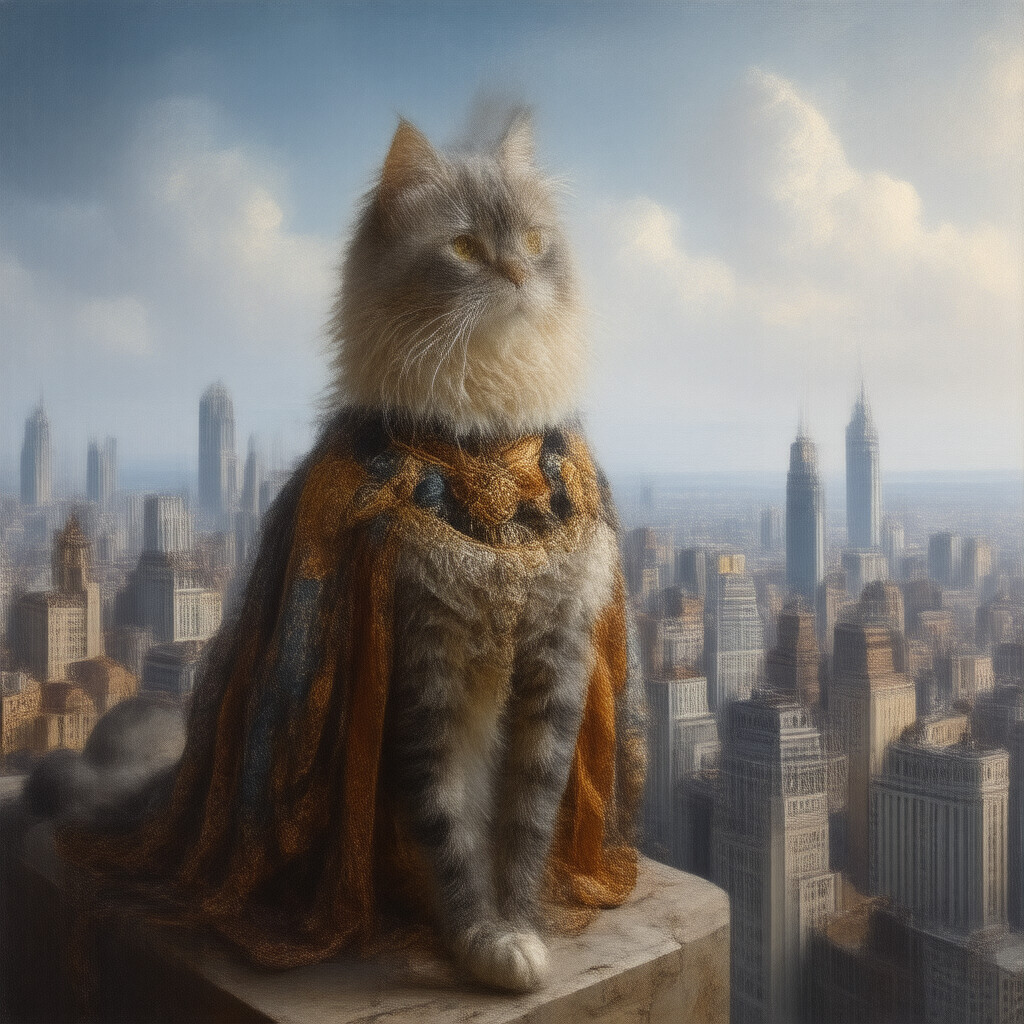}}}
\newcommand\appImgCCC{\adjustbox{valign=m,vspace=0pt,margin=0pt}{\includegraphics[width=.33\linewidth]{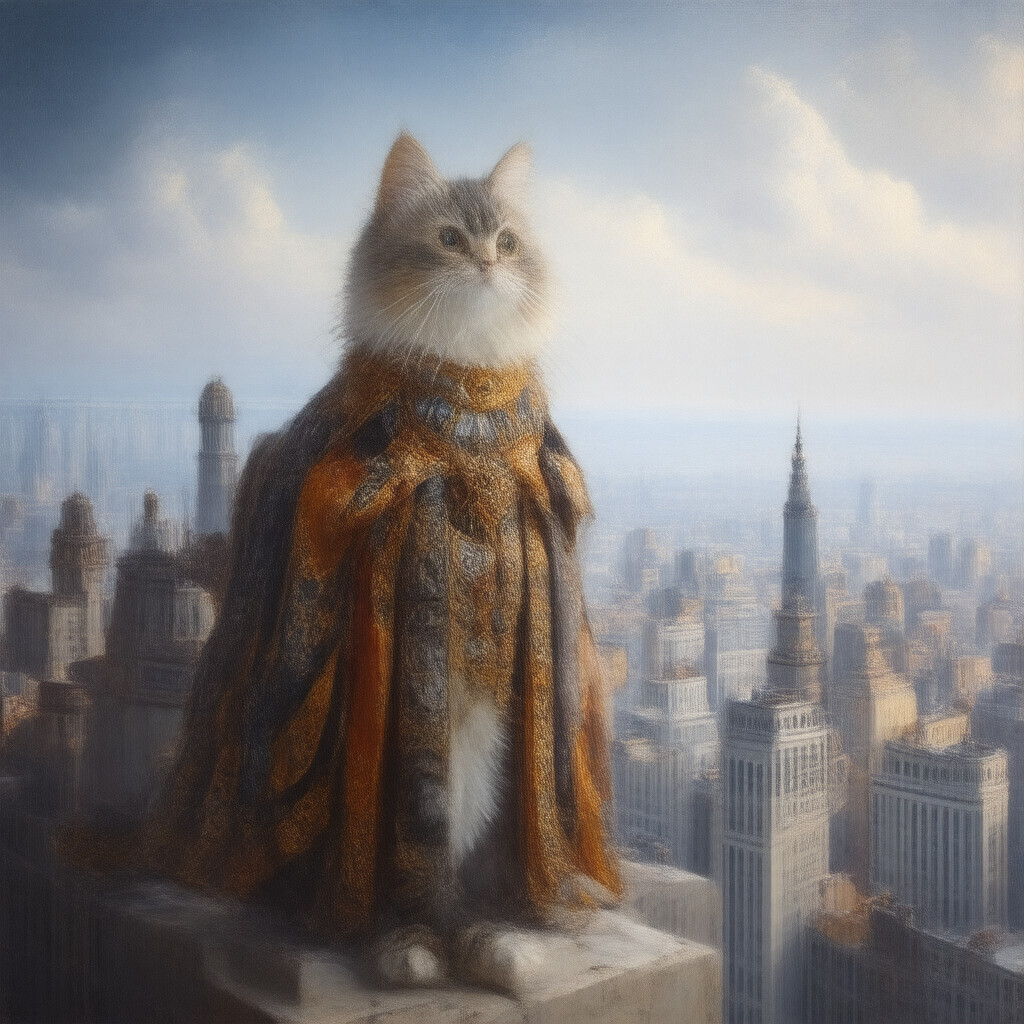}}}
\newcommand\appImgDDD{\adjustbox{valign=m,vspace=0pt,margin=0pt}{\includegraphics[width=.33\linewidth]{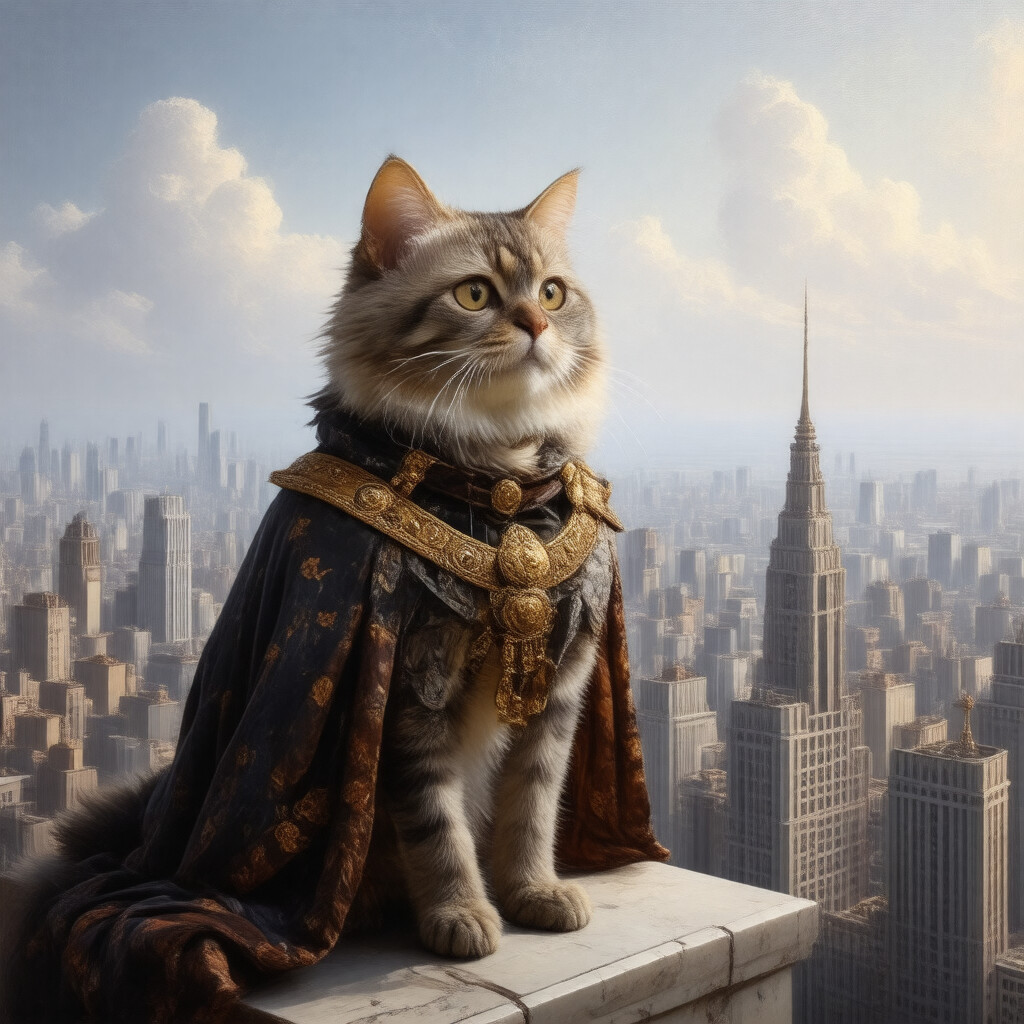}}}
\newcommand\appImgEEE{\adjustbox{valign=m,vspace=0pt,margin=0pt}{\includegraphics[width=.33\linewidth]{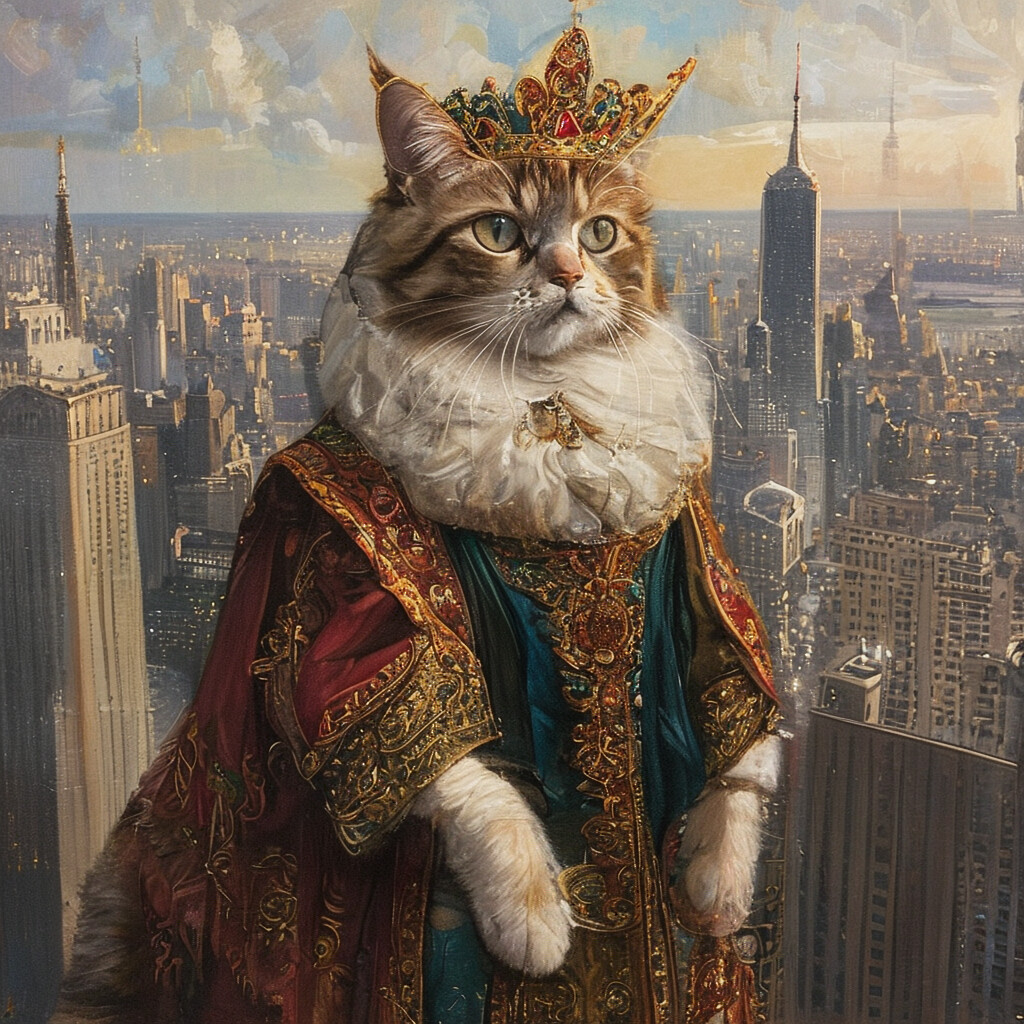}}}

\newcommand\imgAA{\adjustbox{valign=m,vspace=0pt,margin=0pt}{\includegraphics[width=.33\linewidth]{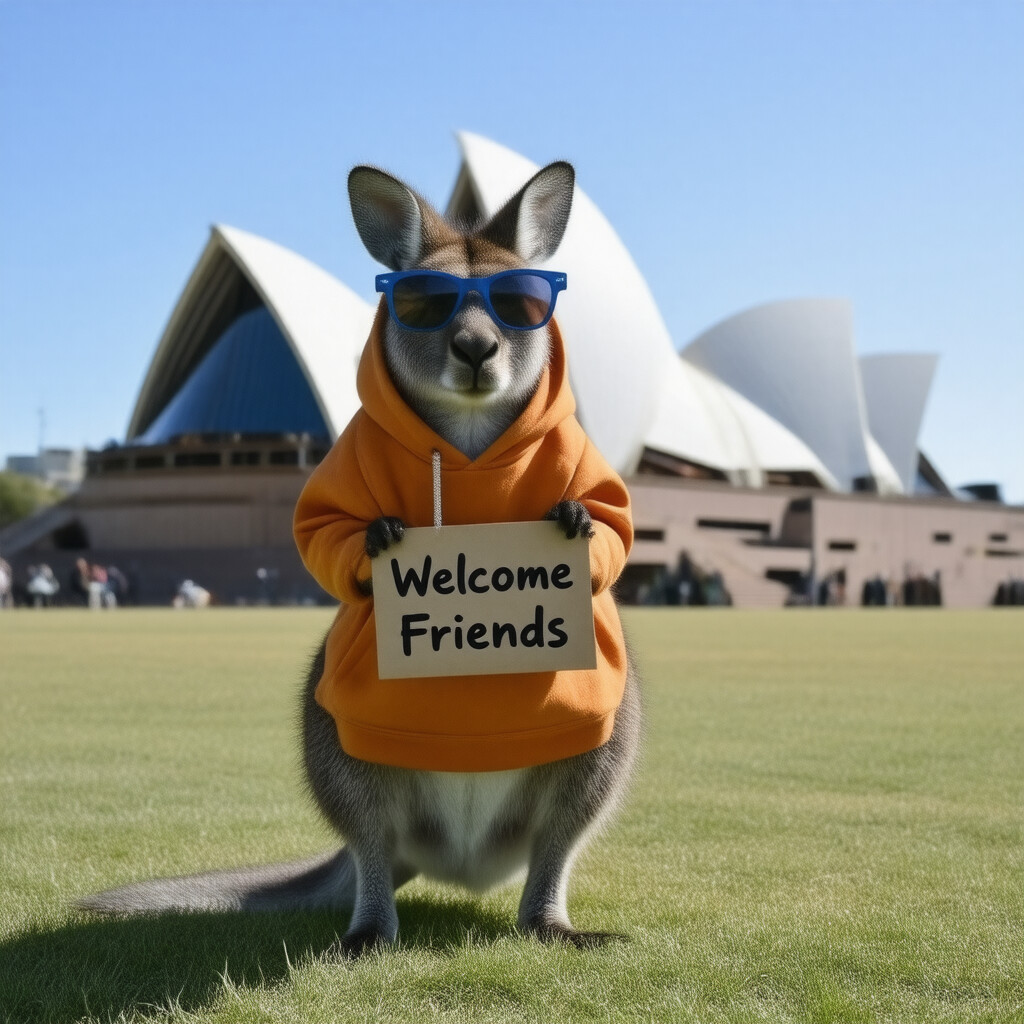}}}
\newcommand\imgBB{\adjustbox{valign=m,vspace=0pt,margin=0pt}{\includegraphics[width=.33\linewidth]{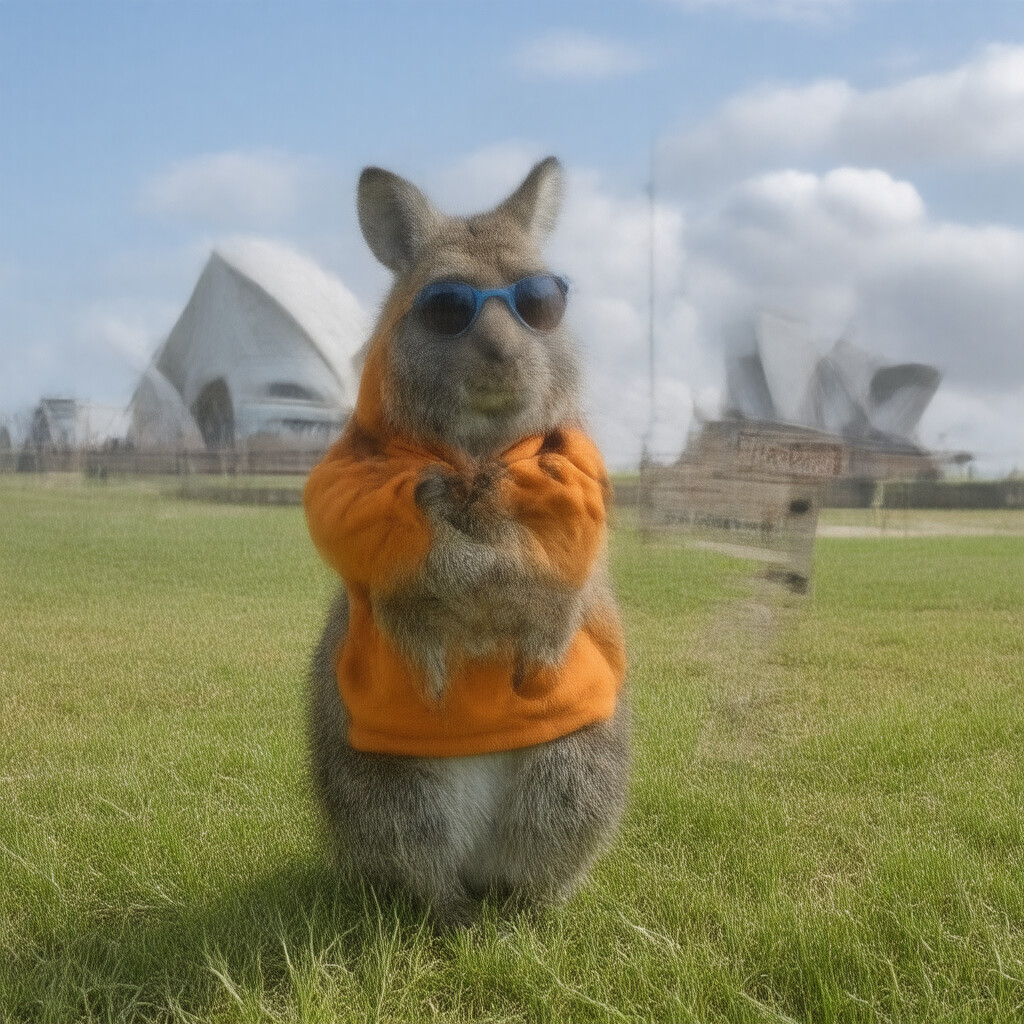}}}
\newcommand\imgCC{\adjustbox{valign=m,vspace=0pt,margin=0pt}{\includegraphics[width=.33\linewidth]{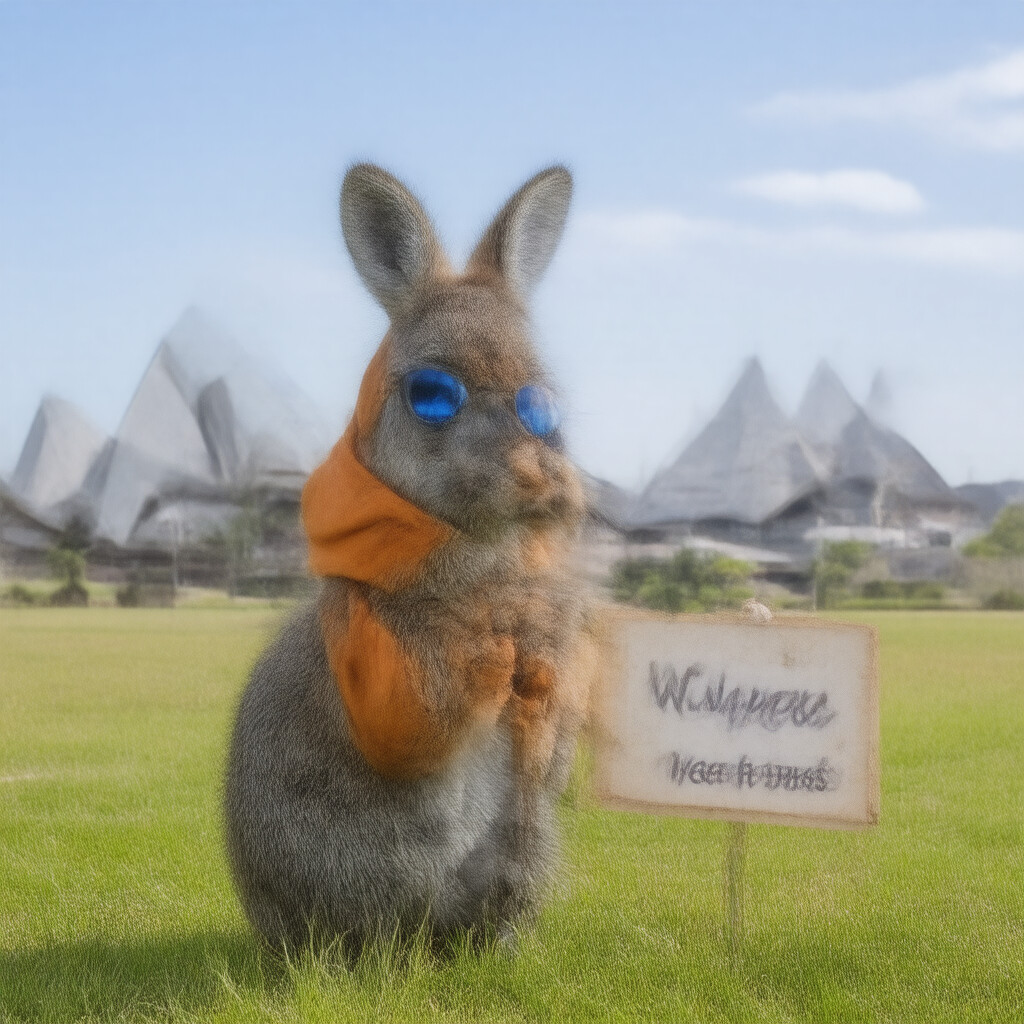}}}
\newcommand\imgDD{\adjustbox{valign=m,vspace=0pt,margin=0pt}{\includegraphics[width=.33\linewidth]{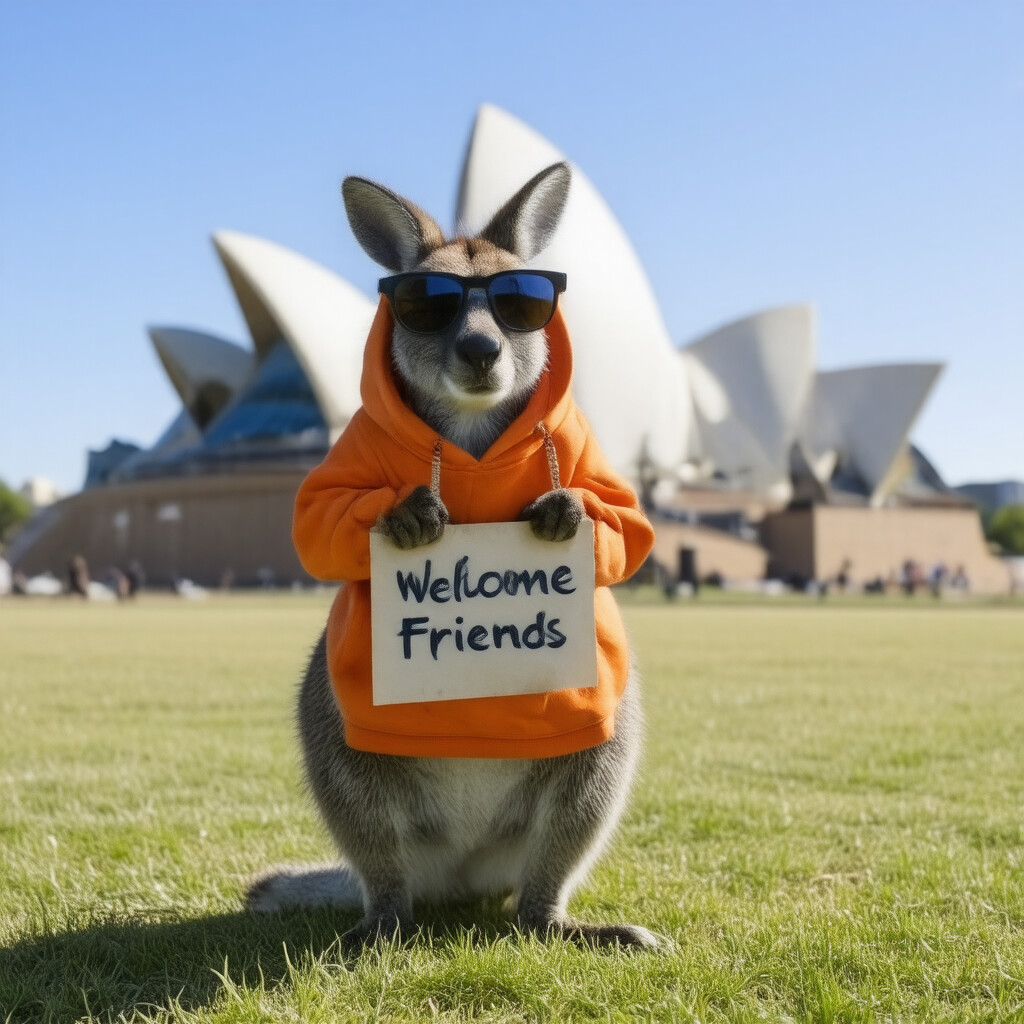}}}
\newcommand\imgEE{\adjustbox{valign=m,vspace=0pt,margin=0pt}{\includegraphics[width=.33\linewidth]{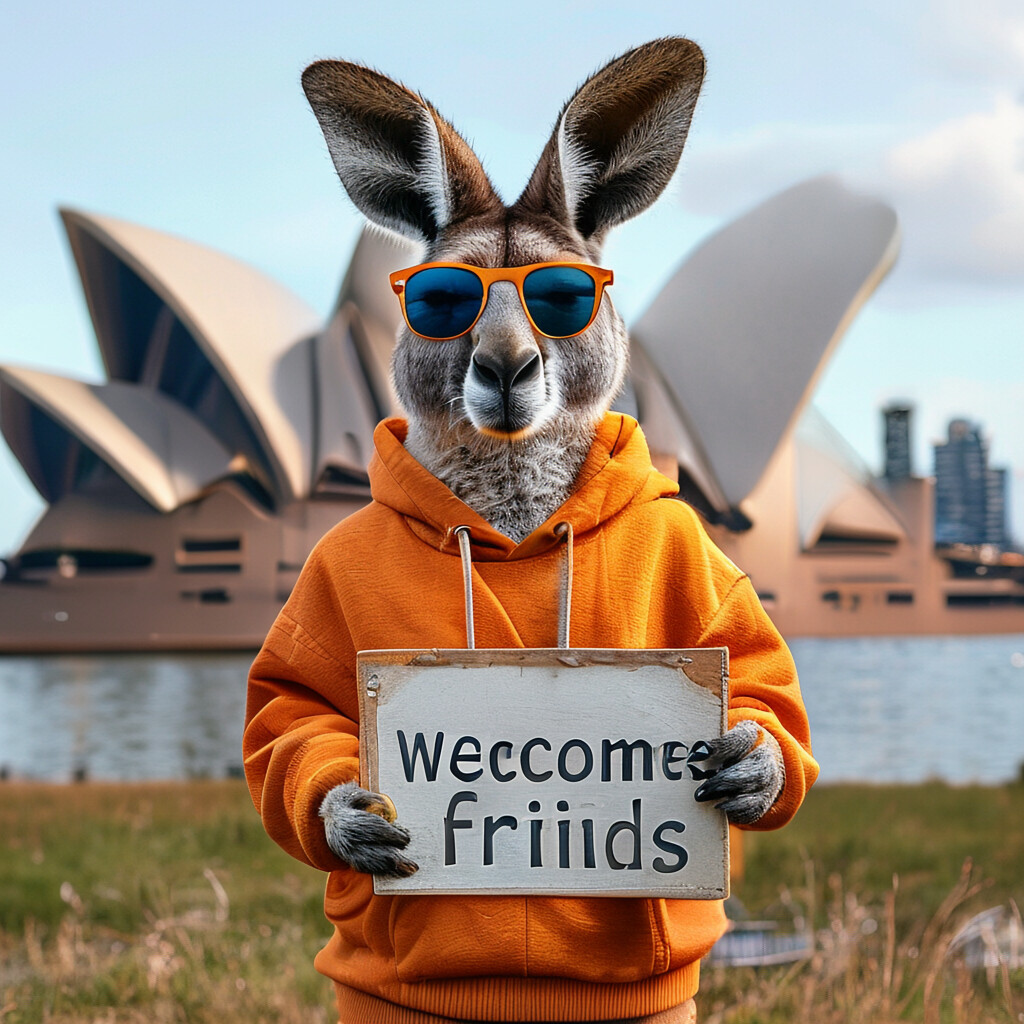}}}

\begin{figure*}[t]
    \centering
    \resizebox{1.01\linewidth}{!}{%
    \begin{tabular}{@{}c| @{\hspace{0.2cm}}c @{\hspace{0.1cm}}c @{\hspace{0.1cm}}c@{} @{\hspace{0.2cm}} |c}
    \LARGE \textbf{Original Model} &\multicolumn{3}{c|}{\LARGE \textbf{Compression Methods}} & \LARGE \textbf{Non-Compression} \\
        \toprule
                \LARGE SD3.5 Large Turbo & \LARGE BK-SDM & \LARGE KOALA  & \LARGE HierarchicalPrune (Ours) & \LARGE SANA-Sprint-1.6B \\
        \appImgAAA & \appImgBBB & \appImgCCC &
        \appImgDDD & \appImgEEE \vspace{0.2cm} \\
        \multicolumn{5}{c}{\textit{\LARGE "A painting of a Persian cat dressed as a Renaissance king, standing on a skyscraper overlooking a city."}} \vspace{0.2cm} \\
        \imgAA & \imgBB & \imgCC &
        \imgDD & \imgEE \vspace{0.2cm} \\
        \multicolumn{5}{c}{\textit{\LARGE "A kangaroo in an orange hoodie and blue sunglasses stands on the grass in front of the Sydney Opera House}} \vspace{0.1cm} \\
    \end{tabular}
    }
    \caption{Visual comparison demonstrating the quality difference between the original model (column~1), depth pruning based on BK-SDM (column~2), KOALA (column~3), and our proposed \sysname (column~4), as well as the SOTA small-scale diffusion model, SANA-Sprint-1.6B (column~5). Our approach successfully maintains visual quality while delivering 79.5\% memory reduction over the original model. Notably, our method preserves the text drawing capability of the original SD3.5 Large Turbo model, where SANA-Sprint-1.6B is limited.}
    \label{fig:image_quality}
    \vspace{-0.4cm}
\end{figure*}

\textbf{Baselines:}
We compare \sysname with two prior works related to depth pruning and distillation:
\textit{i)}~KOALA~\citep{lee2024koala}, and \textit{ii)}~BK-SDM~\citep{kim2024bksdm}. BK-SDM proposed block pruning of U-Net-based models using the CLIP score~\citep{hessel-etal-2021-clipscore}, followed by distilling the pruned model using knowledge from the original model. KOALA follows a similar method, replacing CLIP-score-based importance ranking with scores from each block's input-output cosine similarity. Also, we compare against SANA-Sprint-1.6B~\citep{chen2025sanasprintonestepdiffusioncontinuoustime},
a SOTA compact model optimised for efficient on-device deployment.

\textbf{Metrics:}
To quantitatively evaluate all methods, we employ the two most recent and representative image quality metrics, GenEval~\citep{ghosh2023geneval} (used in~\citep{xie2025sana,esser2024mmdit}) and HPSv2~\citep{wu2023human} (used in KOALA~\citep{lee2024koala}). Furthermore, we conduct a user study to assess human preferences for the generated images that are difficult to capture with the quantitative quality metrics. Similar to~\cite{sheynin2023emueditpreciseimage} and \cite{dai2023emuenhancingimagegeneration}, we evaluate two aspects: (i)~Text Alignment: How well a generated image follows the description of the text prompt, (ii)~Image Quality: The overall visual quality considering clarity, colour, composition, and other factors. On the system resource side, we report the measured peak memory usage and latency of running the target models on diverse GPUs, including A6000 (Table~\ref{tab:latency_measurements}), GTX 3090 and A100 GPUs (see Appendix Table~\ref{app:tab:latency_full}).

\subsection{Main Results}\label{subsec:results}
\textbf{Image Quality under Memory Compression:} Fig.~\ref{fig:image_quality} presents a qualitative comparison of example outputs. \sysname shows better visual outputs, outperforming all the baselines, and maintaining high fidelity to the original DMs, both in structure and fine details. More examples can be found in Fig.~\ref{fig:intro} and Appendix~\ref{app:additional_results}. Table~\ref{tab:quality_metric} shows that \sysname achieves comparable quality to the original model based on both HPSv2 and GenEval. While baselines induce substantial quality degradation of 38.2-45.1\%, our approach yields only a minimal drop ($\sim$3.2-4.8\%), given the same range of memory compression ratio, showcasing the effectiveness of our hierarchical compression strategy.

Importantly, \sysname achieves 79.5\% lower peak memory, a reduction from 15.8 GB to 3.24 GB for SD3.5 Large Turbo model. Notably, the peak memory of our compressed SD3.5 model (3.24 GB) is comparable to SANA-Sprint (3.14 GB), but with substantially better image quality. For FLUX.1-Schnell, our method achieves 80.4\% peak memory reduction, indicating its generalisability to larger DMs.

\renewcommand{\arraystretch}{0.84}
\begin{table}[t]
  \vspace{-0.1cm}
  \centering
  \scriptsize
  \setlength{\tabcolsep}{1.4pt}
  \setlength{\aboverulesep}{1.0pt}
  \setlength{\belowrulesep}{1.3pt}
  \begin{tabular}{ l l | c | c c | c }
    \toprule
     \multicolumn{1}{c}{\textbf{Model}} & \multicolumn{1}{c|}{\textbf{Method}} & \textbf{Memory} & \textbf{GenEval $\Uparrow$} & \textbf{HPSv2 $\Uparrow$} & \textbf{Reduction $\Downarrow$} \\
        \cmidrule(l){1-6}
   Linear DiT & SANA-Sprint & 3.14 (100\%)             & 0.77 & 29.61	& - \\
        \cmidrule(l){1-6}
    \multirow{7}{*}{ \parbox[l]{0.8cm}{SD3.5 \\ Large \\ Turbo }} & Original                       & 15.8 (100\%) & 0.71 & 30.29	& - \\
        \cmidrule(l){2-6}
          & KOALA                    & 12.6 (79.4\%)	 & 0.37 & 19.99	& 41.2\% \\
          & KOALA (+Quant)              & 3.56 (22.5\%)	 & 0.33 & 18.44	& 46.4\% \\
           & BK-SDM & 12.6 (79.4\%)	 & 0.38 & 21.21	& 38.2\% \\
           & BK-SDM (+Quant) & 3.56 (22.5\%)	 & 0.34 & 19.83	& 43.3\% \\
        \cmidrule(l){2-6}
          & Ours  (HPP+PWP+Q) & 3.56 (22.5\%)  & \textbf{0.69} & \textbf{28.15}	& \textbf{4.8}\% \\
     & Ours (All)         & \textbf{3.24 (20.5\%)}	           & 0.62 & 26.29	& 13.3\% \\
        \cmidrule(l){1-6}
    \multirow{4}{*}{ \parbox[l]{0.8cm}{ FLUX.1 \\ Schnell }} & Original     & 22.6 (100\%)             & 0.66 & 29.71	& - \\
        \cmidrule(l){2-6}
     & KOALA          & 15.9 (70.5\%)	     & 0.38 & 25.24	& 28.7\% \\
     & BK-SDM         & 15.9 (70.5\%)	     & 0.45 & 27.38	& 19.8\% \\
        \cmidrule(l){2-6}
     & Ours (All)             & \textbf{4.44 (19.6\%)}        & \textbf{0.64} & \textbf{28.69} & \textbf{3.2}\% \\
    \bottomrule
  \end{tabular}
  \vspace{-0.2cm}
    \caption{Image quality measured in GenEval and HPSv2 scores and the corresponding peak memory measurements in GB and remaining ratios (\%) after compression. KOALA and BK-SDM experience substantial degradation of image quality when reducing memory usage by 20-30\%. \sysname achieves significantly reduced memory usage while maintaining the image quality close to the original models.}
  \label{tab:quality_metric}
  \vspace{-0.5cm}
\end{table}

\textbf{User Study:}
Our user study with 85 participants reaffirms the effectiveness of \sysname. Fig.~\ref{fig:teaser} presents the mean opinion score (MOS) across all methods. \sysname achieves remarkably close MoS to the original SD3.5 Large Turbo model with only minimal reduction (4.8\% for text alignment, 5.3\% for image quality). In contrast, SANA-Sprint-1.6B shows a noticeable quality drop (14.2\% for text alignment, 11.1\% for image quality) compared to the original model, while other baselines (\textit{i.e.}, BK-SDM, KOALA) show a substantial 44.0-52.2\% degradation.

\textbf{Training Cost Comparison:}
Beyond quality preservation, \sysname requires only 615--1,287 A100 GPU hours for architectural profiling and distillation (Appendix~\ref{app:additional_results:cost_analysis}), compared to 140k--200k A100 GPU hours required to pre-train small-scale DMs like SD1.4 and SD2.1~\cite{2024ICLRgU58d5QeGv_cascade}. This demonstrates that compressing existing large models is far more cost-efficient than training compact alternatives from scratch, while maintaining superior quality.

\renewcommand{\arraystretch}{0.86}
\begin{table}[t]
  \vspace{-0.2cm}
  \centering
  \scriptsize
  \setlength{\aboverulesep}{1.1pt}
  \setlength{\belowrulesep}{1.1pt}
\begin{tabular}{ll|cc}
\toprule
\textbf{Model} & \multicolumn{1}{c|}{\textbf{Method}} & \textbf{Latency $\Downarrow$}    & \textbf{Reduction $\Uparrow$}    \\ \cmidrule{1-4}
Linear DiT     & SANA-Sprint            & 54 ms                            & -                                \\ \cmidrule{1-4}
               & Original               & 823 ms                           & -                                \\ \cmidrule{2-4}
               & KOALA                  & 642 ms                           & 22.0\%                           \\
SD3.5          & BK-SDM                 & 642 ms                           & 22.0\%                           \\ \cmidrule{2-4}
Large Turbo    & Ours (HPP, PWP, Quant) & \multirow{2}{*}{\textbf{593 ms}} & \multirow{2}{*}{\textbf{27.9\%}} \\
               & Ours (All)             &                                  &                                  \\ \cmidrule{1-4}
               & Original               & 756 ms                           & -            \\ \cmidrule{2-4}
FLUX.1         & KOALA                  & \textbf{432 ms}                  & \textbf{42.9\%}                           \\
Schnell        & BK-SDM                 & \textbf{432 ms}                  & \textbf{42.9\%}                           \\ \cmidrule{2-4}
               & Ours (All)             & 469 ms                           & 38.0\%                           \\
    \bottomrule
  \end{tabular}
  \vspace{-0.1cm}
    \caption{Comparison of per-step latency of the DM Transformer measured on an A6000 GPU with 48 GB of VRAM.}
  \label{tab:latency_measurements}
  \vspace{-0.4cm}
\end{table}

\textbf{Latency:} We measure per-step latency of \sysname and baselines both server- and desktop-grade GPUs with different specifications (A100, A6000, and GTX 3090 with 80, 48, and 25 GB VRAM, respectively, reported in Appendix Table~\ref{app:tab:latency_full}). While we report the measurement results from A6000 in Table~\ref{tab:latency_measurements}, the reduction rates are similar across GPUs. Our compression pipeline achieves 27.9\% and 38.0\% latency reduction compared to the original SD3.5 Large Turbo and FLUX.1-Schnell, respectively.

\textit{Overall, our findings reveal that respecting the hierarchical sensitivity of diffusion model components enables more effective model compression than approaches that treat all blocks as equally important, establishing a new paradigm for the efficient deployment of large-scale generative models.}

\subsection{Ablation Study and Analysis}\label{subsec:ablation}
\textbf{Impact of Each Component of \sysname:}
We conducted an ablation study of \sysname~to investigate the contribution of each component: (1)~HPP, (2)~PWP, (3)~HPP+PWP+SGDistill, and (4) our final form, HPP+PWP+SGDistill+Quant. Table~\ref{tab:quality_metric_ablation} shows that using HPP+PWP drastically improves image quality compared to HPP only and prior works (\textit{i.e.}, BK-SDM, KOALA). Moreover, by leveraging intra-block sensitivity, SGDistill substantially reduces image quality degradation from 31.0\% to 10.1\% compared to HPP+PWP at an aggressive paramter reduction rate of 30\%\footnote{We set $r_{thres}=0.25$ for both models, determined where quality degradation exceeded 15\% without SGDistill.}. \textit{Our method outperforms all baselines with superior image quality, lower memory, and faster execution.}

\textbf{Impact of Quantisation:}
We investigate how much W4A16 quantisation in our pipeline affects the final image quality. Table~\ref{tab:quality_metric_ablation} shows the quality metrics with and without quantisation applied. The impact of quantisation is as small as 2.4-3.5\% in both GenEval and HPSv2 scores.

\textbf{Robustness of Text Generation:}
As shown in Fig.~\ref{fig:image_quality}, prior works (BK-SDM and KOALA in pruning) and small-scale DMs (SANA-Sprint-1.6B) are limited in synthesising legible texts in the generated images. However, \sysname demonstrates the superior quality in text generation in Fig.~\ref{fig:image_quality} and~\ref{fig:intro}.
\textit{These results represent the effectiveness of our proposed method in preserving the innate capability of the original model, not only in the aesthetic quality but also in other aspects like text generation.}

\begin{table}[t]
  \vspace{-0.2cm}
  \centering
  \scriptsize
  \setlength{\tabcolsep}{2.5pt}
  \setlength{\aboverulesep}{1.0pt}
  \setlength{\belowrulesep}{1.3pt}
\begin{tabular}{c|l|c|cc|c}
\toprule
\textbf{\begin{tabular}[c]{@{}c@{}}Prunning \\ Ratio\end{tabular}} & \multicolumn{1}{c|}{\textbf{Method}}  & \textbf{\begin{tabular}[c]{@{}c@{}}Remaining \\ Memory\end{tabular}}  & \textbf{GenEval $\Uparrow$} & \textbf{HPSv2 $\Uparrow$} & \textbf{Reduction $\Downarrow$} \\ \midrule
None (0\%) & Original          & 100\%                & 0.71                        & 30.29                     & -                               \\ \midrule
\multirow{3}{*}{ \parbox[c]{1.04cm}{\centering Moderate \\ Pruning \\ (20\%) }} & Ours (HPP)        & 79.4\%               & 0.03                        & 11.08                     & 79.4\%                          \\
& Ours (+PWP)       & 79.4\%               & \textbf{0.71}               & \textbf{28.97}            & \textbf{2.5\%}                           \\
 & Ours (+Quant)     & \textbf{22.5\%}      & 0.69                        & 28.15                     & 4.8\%                           \\ \midrule
\multirow{4}{*}{ \parbox[c]{1.04cm}{\centering Aggressive \\ Pruning \\ (30\%) }} & Ours (HPP)        & 71.5\%               & 0.0                         & 7.00                      & 88.4\%                          \\
 & Ours (+PWP)       & 71.5\%               & 0.46                        & 21.74                     & 31.9\% \\
 & Ours (+SGDistill) & 71.5\%               & \textbf{0.64}               & \textbf{27.29}            & \textbf{10.1\%} \\
 & Ours (+Quant)     & \textbf{20.5\%}      & 0.62                        & 26.29                     & 13.3\%                          \\ \bottomrule
\end{tabular}
  \vspace{-0.1cm}
  \caption{Ablation study of each component and quantisation in \sysname on SD3.5 Large Turbo.}
  \label{tab:quality_metric_ablation}
  \vspace{-0.4cm}
\end{table}
\renewcommand{\arraystretch}{1.0}

\section{Related Work}\label{sec:related}

\textbf{T2I Diffusion Models.}~Since Stable Diffusion (SD)~\citep{2022CVPRLDM}, there has been rapid community adoption and iterative updates in the field. SDv1.4 and v1.5 were released in 2022, enhancing efficiency and enabling specialised fine-tuning (\textit{e.g.}, DreamBooth). By mid-2023, Stable Diffusion XL (SDXL) significantly advanced resolution (1024$\times$1024), text comprehension, and image quality. By the end of 2024, the adoption of MMDiT~\citep{esser2024mmdit} as a backbone in SD3, SD3.5, and FLUX brought a significant boost in image generation quality and alignment with long text input. At the same time, the increasing parameter count, particularly in the backbone Transformer blocks, significantly improved image quality but with excessive resource demands. As SOTA models like SD3.5 scale up to as many as 8B parameters, efficient inference on resource-constrained devices becomes impractical. This underlines the significance of approaches such as our \sysname framework, to enable DM deployment outside of high-end compute setups.

\textbf{Compression Techniques.}~Traditional model compression methods, including  distillation~\citep{hinton2015distilling}, pruning~\citep{han2015deepcompression} (such as depth pruning via block removal \citep{ghiasi2018dropblock, kim2024layermerge}, width pruning~\citep{kwon2024tinytrain}), and quantisation~\citep{jacob2018quantization} have been applied to DMs~\citep{lee2024koala,kim2024bksdm, castells2024edgefusion,hu2024snapgen, fang2024tinyfusion,li2025svdquant,kim2024IDcompression,zhang2024laptopdiff}, resulting in variable parameter efficiency gains. Specifically, \cite{kim2024bksdm} and \cite{lee2024koala} explore block removal in the U-Net backbone of SD1.5 and SDXL models, respectively. More recently, \cite{li2025svdquant} explored the effectiveness of quantisation on DMs for memory reduction, hence our \sysname employs adjunct post-training quantisation as part of its design. Orthogonal to compression, $\Delta$-DiT~\cite{chen2024deltaDit} identifies hierarchical patterns in DiT (MMDiT predecessor) for inference caching. Our analysis confirms that this generalises to modern MMDiT and further leverages these insights for the compression of SOTA multi-billion-scale DMs, under a unified framework.

\section{Conclusion}\label{sec:conclusion}
In this work, we proposed \sysname, pushing the limit of compressing MMDiT-based large-scale DMs through hierarchical insights. By combining HPP, PWP, and SGDistil with quantisation, \sysname achieves 77.5-80.4\% memory reduction and 27.9-38.0\% latency improvements with minimal quality loss, bringing SOTA large-scale DMs (SD3.5 Large Turbo, FLUX.1-Schnell) within reach of resource-constrained environments and democratising access to high-quality T2I generation.

\bibliography{main}

@String(CVPR= {IEEE Conf. Comput. Vis. Pattern Recog.})

@String(ICCV= {Int. Conf. Comput. Vis.})

@String(ECCV= {Eur. Conf. Comput. Vis.})

@String(ICLR = {Int. Conf. Learn. Represent.})

@String(CVPRW= {IEEE Conf. Comput. Vis. Pattern Recog. Worksh.})

@String(CVPR  = {CVPR})

@String(ICCV  = {ICCV})

@String(ECCV  = {ECCV})

@String(ICLR  = {ICLR})

@String(CVPRW= {CVPRW})

@misc{chen2025sanasprintonestepdiffusioncontinuoustime,
      title={{SANA-Sprint: One-Step Diffusion with Continuous-Time Consistency Distillation}}, 
      author={Junsong Chen and Shuchen Xue and Yuyang Zhao and Jincheng Yu and Sayak Paul and Junyu Chen and Han Cai and Enze Xie and Song Han},
      year={2025},
      eprint={2503.09641},
      archivePrefix={arXiv},
      primaryClass={cs.GR},
      url={https://arxiv.org/abs/2503.09641}, 
}

@misc{hu2024ellaequipdiffusionmodels,
      title={{ELLA: Equip Diffusion Models with LLM for Enhanced Semantic Alignment}}, 
      author={Xiwei Hu and Rui Wang and Yixiao Fang and Bin Fu and Pei Cheng and Gang Yu},
      year={2024},
      eprint={2403.05135},
      archivePrefix={arXiv},
      primaryClass={cs.CV},
      url={https://arxiv.org/abs/2403.05135}, 
}

@inproceedings{2024ICLRgU58d5QeGv_cascade,
  title={{{W\"uerstchen: An Efficient Architecture for Large-Scale Text-to-Image Diffusion Models}}},
  author={{Pernias}, Pablo and {Rampas}, Dominic and {Richter}, Mats L. and {Pal}, Christopher J. and {Aubreville}, Marc},
  year={2024},
  booktitle={https://openreview.net/forum?id=gU58d5QeGv}
}

@misc{stability_ai_sdxl_turbo,
  author = {{Stability AI}},
  title = {{SDXL-Turbo}},
  howpublished = "\url{https://huggingface.co/stabilityai/sdxl-turbo}",
  year = {2023},
}

@inproceedings{2025seedream3,
  author={ByteDance},
  title={{Seedream 3.0 Technical Report}},
  year={2025},
  booktitle={https://arxiv.org/pdf/2504.11346}
}

@inproceedings{2023arXiv231117042S_add,
  author={{Sauer}, Axel and {Lorenz}, Dominik and {Blattmann}, Andreas and {Rombach}, Robin},
  title={{Adversarial Diffusion Distillation}},
  year={2023},
  booktitle={https://arxiv.org/abs/2311.17042}
}

@inproceedings{2023arXiv230600980L_snapfusion,
  author={{Li}, Yanyu and {Wang}, Huan and {Jin}, Qing and {Hu}, Ju and {Chemerys}, Pavlo and {Fu}, Yun and {Wang}, Yanzhi and {Tulyakov}, Sergey and {Ren}, Jian},
  title={{SnapFusion: Text-to-Image Diffusion Model on Mobile Devices within Two Seconds}},
  year={2023},
  booktitle={https://arxiv.org/abs/2306.00980}
}

@article{hu2024snapgen,
  title={SnapGen: Taming High-Resolution Text-to-Image Models for Mobile Devices with Efficient Architectures and Training},
  author={Hu, Dongting and Chen, Jierun and Huang, Xijie and Coskun, Huseyin and Sahni, Arpit and Gupta, Aarush and Goyal, Anujraaj and Lahiri, Dishani and Singh, Rajesh and Idelbayev, Yerlan and others},
  journal={arXiv preprint arXiv:2412.09619},
  year={2024}
}

@inproceedings{2024arXiv240203666W_quest,
  author={{Wang}, Haoxuan and {Shang}, Yuzhang and {Yuan}, Zhihang and {Wu}, Junyi and {Yan}, Yan},
  title={{QuEST: Low-bit Diffusion Model Quantization via Efficient Selective Finetuning}},
  year={2024},
  booktitle={https://arxiv.org/abs/2402.03666}
}

@inproceedings{2022arXiv221002747L_FlowMatching,
  author={{Lipman}, Yaron and {Chen}, Ricky T.~Q. and {Ben-Hamu}, Heli and {Nickel}, Maximilian and {Le}, Matt},
  title={{Flow Matching for Generative Modeling}},
  year={2023},
  booktitle={International Conference on Learning Representations (ICLR)}
}

@inproceedings{2021ICLRPxTIG12RRHS_SDE,
  title={{{Score-Based Generative Modeling through Stochastic Differential Equations}}},
  author={{Song}, Yang and {Sohl-Dickstein}, Jascha and {Kingma}, Diederik P. and {Kumar}, Abhishek and {Ermon}, Stefano and {Poole}, Ben},
  year={2021},
  booktitle={International Conference on Learning Representations (ICLR)}
}

@inproceedings{2022arXiv220600364K_karras,
  author={{Karras}, Tero and {Aittala}, Miika and {Aila}, Timo and {Laine}, Samuli},
  title={{Elucidating the Design Space of Diffusion-Based Generative Models}},
  year={2022},
  booktitle={Advances in Neural Processing Systems (NeurIPS)}
}

@inproceedings{2022arXiv221209748P_dit,
  author={{Peebles}, William and {Xie}, Saining},
  title={{Scalable Diffusion Models with Transformers}},
  year={2022},
  booktitle={International Conference on Computer Vision (ICCV)}
}

@inproceedings{2024arXiv240212376L_fit,
  author={{Lu}, Zeyu and {Wang}, Zidong and {Huang}, Di and {Wu}, Chengyue and {Liu}, Xihui and {Ouyang}, Wanli and {Bai}, Lei},
  title={{{FiT}: Flexible Vision Transformer for Diffusion Model}},
  year={2024},
  booktitle={International Conference on Machine Learning (ICML)}
}

@inproceedings{2022CVPRLDM,
  title={{{High-resolution image synthesis with latent diffusion models}}},
  author={Rombach, Robin and Blattmann, Andreas and Lorenz, Dominik and Esser, Patrick and Ommer, Bj{\"o}rn},
  year={2022},
  booktitle={{IEEE/CVF Conference on Computer Vision and Pattern Recognition (CVPR)}}
}

@article{2022NeurIPSLAION5B,
  title={{{LAION-5B: An open large-scale dataset for training next generation image-text models}}},
  author={{Schuhmann}, Christoph and {Beaumont}, Romain and {Vencu}, Richard and {Gordon}, Cade and {Wightman}, Ross and {Cherti}, Mehdi and {Coombes}, Theo and {Katta}, Aarush and {Mullis}, Clayton and {Wortsman}, Mitchell and {Schramowski}, Patrick and {Kundurthy}, Srivatsa and {Crowson}, Katherine and {Schmidt}, Ludwig and {Kaczmarczyk}, Robert and {Jitsev}, Jenia},
  year={2022},
  journal={Advances in Neural Information Processing Systems (NeurIPS)},
}

@inproceedings{2023arXiv230701952P_sdxl,
  author={{Podell}, Dustin and {English}, Zion and {Lacey}, Kyle and {Blattmann}, Andreas and {Dockhorn}, Tim and {M{\"u}ller}, Jonas and {Penna}, Joe and {Rombach}, Robin},
  title={{{SDXL}: Improving Latent Diffusion Models for High-Resolution Image Synthesis}},
  year={2023},
  booktitle={https://arxiv.org/abs/2307.01952}
}

@misc{dai2023emuenhancingimagegeneration,
      title={{Emu: Enhancing Image Generation Models Using Photogenic Needles in a Haystack}}, 
      author={Xiaoliang Dai and Ji Hou and Chih-Yao Ma and Sam Tsai and Jialiang Wang and Rui Wang and Peizhao Zhang and Simon Vandenhende and Xiaofang Wang and Abhimanyu Dubey and Matthew Yu and Abhishek Kadian and Filip Radenovic and Dhruv Mahajan and Kunpeng Li and Yue Zhao and Vladan Petrovic and Mitesh Kumar Singh and Simran Motwani and Yi Wen and Yiwen Song and Roshan Sumbaly and Vignesh Ramanathan and Zijian He and Peter Vajda and Devi Parikh},
      year={2023},
      eprint={2309.15807},
      archivePrefix={arXiv},
      primaryClass={cs.CV},
      url={https://arxiv.org/abs/2309.15807}, 
}

@misc{sheynin2023emueditpreciseimage,
      title={{Emu Edit: Precise Image Editing via Recognition and Generation Tasks}}, 
      author={Shelly Sheynin and Adam Polyak and Uriel Singer and Yuval Kirstain and Amit Zohar and Oron Ashual and Devi Parikh and Yaniv Taigman},
      year={2023},
      eprint={2311.10089},
      archivePrefix={arXiv},
      primaryClass={cs.CV},
      url={https://arxiv.org/abs/2311.10089}, 
}

@article{wu2023human,
  title={{Human Preference Score v2: A Solid Benchmark for Evaluating Human Preferences of Text-to-Image Synthesis}},
  author={Wu, Xiaoshi and Hao, Yiming and Sun, Keqiang and Chen, Yixiong and Zhu, Feng and Zhao, Rui and Li, Hongsheng},
  journal={arXiv preprint arXiv:2306.09341},
  year={2023}
}

@article{lee2024koala,
  title={{Koala: Empirical Lessons toward Memory-Efficient and Fast Diffusion Models for Text-to-Image Synthesis}},
  author={Lee, Youngwan and Park, Kwanyong and Cho, Yoorhim and Lee, Yong-Ju and Hwang, Sung Ju},
  journal={Advances in Neural Information Processing Systems (NeurIPS)},
  year={2024}
}

@article{ghosh2023geneval,
  title={{Geneval: An object-focused framework for evaluating text-to-image alignment}},
  author={Ghosh, Dhruba and Hajishirzi, Hannaneh and Schmidt, Ludwig},
  journal={Advances in Neural Information Processing Systems (NeurIPS)},
  year={2023}
}

@inproceedings{esser2024mmdit,
  title={{Scaling Rectified Flow Transformers for High-Resolution Image Synthesis}},
  author={Esser, Patrick and Kulal, Sumith and Blattmann, Andreas and Entezari, Rahim and M{\"u}ller, Jonas and Saini, Harry and Levi, Yam and Lorenz, Dominik and Sauer, Axel and Boesel, Frederic and others},
  booktitle={International Conference on Machine Learning (ICML)},
  year={2024}
}

@article{hinton2015distilling,
  title={{Distilling the Knowledge in a Neural Network}},
  author={Hinton, Geoffrey and Vinyals, Oriol and Dean, Jeff},
  journal={arXiv preprint arXiv:1503.02531},
  year={2015}
}

@article{zhang2024laptopdiff,
  title={Laptop-diff: Layer pruning and normalized distillation for compressing diffusion models},
  author={Zhang, Dingkun and Li, Sijia and Chen, Chen and Xie, Qingsong and Lu, Haonan},
  journal={arXiv preprint arXiv:2404.11098},
  year={2024}
}

@inproceedings{kim2024IDcompression,
  title={Diffusion Model Compression for Image-to-Image Translation},
  author={Kim, Geonung and Kim, Beomsu and Park, Eunhyeok and Cho, Sunghyun},
  booktitle={Proceedings of the Asian Conference on Computer Vision},
  pages={2105--2123},
  year={2024}
}

@article{chen2024deltaDit,
  title={$Delta$-DiT: A Training-Free Acceleration Method Tailored for Diffusion Transformers},
  author={Chen, Pengtao and Shen, Mingzhu and Ye, Peng and Cao, Jianjian and Tu, Chongjun and Bouganis, Christos-Savvas and Zhao, Yiren and Chen, Tao},
  journal={arXiv preprint arXiv:2406.01125},
  year={2024}
}

@inproceedings{jacob2018quantization,
  title={{Quantization and Training of Neural Networks for Efficient Integer-arithmetic-only Inference}},
  author={Jacob, Benoit and Kligys, Skirmantas and Chen, Bo and Zhu, Menglong and Tang, Matthew and Howard, Andrew and Adam, Hartwig and Kalenichenko, Dmitry},
  booktitle={IEEE Conference on Computer Vision and Pattern Recognition (CVPR)},
  year={2018}
}

@inproceedings{kim2024bksdm,
  title={{BK-SDM: A Lightweight, Fast, and Cheap Version of Stable Diffusion}},
  author={Kim, Bo-Kyeong and Song, Hyoung-Kyu and Castells, Thibault and Choi, Shinkook},
  booktitle={European Conference on Computer Vision (ECCV)},
  year={2024},
}

@inproceedings{
    xie2025sana,
    title={{{SANA}: Efficient High-Resolution Text-to-Image Synthesis with Linear Diffusion Transformers}},
    author={Enze Xie and Junsong Chen and Junyu Chen and Han Cai and Haotian Tang and Yujun Lin and Zhekai Zhang and Muyang Li and Ligeng Zhu and Yao Lu and Song Han},
    booktitle={International Conference on Learning Representations (ICLR)},
    year={2025},
}

@inproceedings{
    li2025svdquant,
    title={{{SVDQ}uant: Absorbing Outliers by Low-Rank Component for 4-Bit Diffusion Models}},
    author={Muyang Li and Yujun Lin and Zhekai Zhang and Tianle Cai and Junxian Guo and Xiuyu Li and Enze Xie and Chenlin Meng and Jun-Yan Zhu and Song Han},
    booktitle={International Conference on Learning Representations (ICLR)},
    year={2025},
}

@article{han2015deepcompression,
  title={{Deep Compression: Compressing Deep Neural Networks with Pruning, Trained Quantization and Huffman Coding}},
  author={Han, Song and Mao, Huizi and Dally, William J},
  journal={International Conference on Learning Representations (ICLR)},
  year={2016}
}

@article{fang2024tinyfusion,
  title={{TinyFusion: Diffusion Transformers Learned Shallow}},
  author={Fang, Gongfan and Li, Kunjun and Ma, Xinyin and Wang, Xinchao},
  journal={arXiv preprint arXiv:2412.01199},
  year={2024}
}

@article{castells2024edgefusion,
  publtype={informal},
  author={Thibault Castells and Hyoung-Kyu Song and Tairen Piao and Shinkook Choi and Bo-Kyeong Kim and Hanyoung Yim and Changgwun Lee and Jae Gon Kim and Tae-Ho Kim},
  title={EdgeFusion: On-Device Text-to-Image Generation},
  year={2024},
  cdate={1704067200000},
  journal={CoRR},
  volume={abs/2404.11925},
  url={https://doi.org/10.48550/arXiv.2404.11925}
}

@inproceedings{kwon2024tinytrain,
    title     = {TinyTrain: Resource-Aware Task-Adaptive Sparse Training of DNNs at the Data-Scarce Edge},
    author    = {Kwon, Young D. and Li, Rui and Venieris, Stylianos I. and Chauhan, Jagmohan and Lane, Nicholas D. and Mascolo, Cecilia},
    booktitle = {International Conference on Machine Learning (ICML)},
    year      = {2024}
}

@article{ghiasi2018dropblock,
  title={{DropBlock: A Regularization Method for Convolutional Networks}},
  author={Ghiasi, Golnaz and Lin, Tsung-Yi and Le, Quoc V},
  journal={Advances in Neural Information Processing Systems (NeurIPS)},
  year={2018}
}

@misc{yepop,
  author={HuggingFace},
  year={2024},
  title = {YE-POP Dataset},
  howpublished = {\url{https://huggingface.co/datasets/Ejafa/ye-pop}},
  note = {Accessed: 2025-March}
}

@misc{blackforestlabs_flux1,
  author = {{Black Forest Labs}},
  title = {{Flux.1 Model Family}},
  howpublished = "\url{https://blackforestlabs.ai/announcing-black-forest-labs/}",
  year = {2024},
}

@InProceedings{Wu_2023_HPS,
    author    = {Wu, Xiaoshi and Sun, Keqiang and Zhu, Feng and Zhao, Rui and Li, Hongsheng},
    title     = {{Human Preference Score: Better Aligning Text-to-Image Models with Human Preference}},
    booktitle = {Proceedings of the IEEE/CVF International Conference on Computer Vision (ICCV)},
    year      = {2023},
}

@misc{von-platen-etal-2022-diffusers,
  author = {Patrick von Platen and Suraj Patil and Anton Lozhkov and Pedro Cuenca and Nathan Lambert and Kashif Rasul and Mishig Davaadorj and Dhruv Nair and Sayak Paul and William Berman and Yiyi Xu and Steven Liu and Thomas Wolf},
  title = {{Diffusers: State-of-the-art diffusion models}},
  year = {2022},
  publisher = {GitHub},
  journal = {GitHub repository},
  howpublished = {\url{https://github.com/huggingface/diffusers}}
}

@inproceedings{he2023ptqd,
    title={{PTQD: Accurate Post-Training Quantization for Diffusion Models}},
    author={Yefei He and Luping Liu and Jing Liu and Weijia Wu and Hong Zhou and Bohan Zhuang},
    booktitle={Thirty-seventh Conference on Neural Information Processing Systems (NeurIPS)},
    year={2023},
}

@inproceedings{FlashAttention,
    author = {Dao, Tri and Fu, Dan and Ermon, Stefano and Rudra, Atri and R\'{e}, Christopher},
    booktitle = {Advances in Neural Information Processing Systems (NeurIPS)},
    title = {{FlashAttention: Fast and Memory-Efficient Exact Attention with IO-Awareness}},
     year = {2022}
}

@inproceedings{
    fang2023structural,
    title={{Structural Pruning for Diffusion Models}},
    author={Gongfan Fang and Xinyin Ma and Xinchao Wang},
    booktitle={Thirty-seventh Conference on Neural Information Processing Systems (NeurIPS)},
    year={2023},
}

@InProceedings{Castells_2024_CVPR_ldpruner,
    author    = {Castells, Thibault and Song, Hyoung-Kyu and Kim, Bo-Kyeong and Choi, Shinkook},
    title     = {{LD-Pruner: Efficient Pruning of Latent Diffusion Models using Task-Agnostic Insights}},
    booktitle = {IEEE/CVF Conference on Computer Vision and Pattern Recognition Workshops (CVPRW)},
    year      = {2024},
}

@inproceedings{kim2024layermerge,
  title={{LayerMerge: Neural Network Depth Compression through Layer Pruning and Merging}},
  author={Kim, Jinuk and Halabi, Marwa El and Ji, Mingi and Song, Hyun Oh},
  booktitle={International Conference on Machine Learning (ICML)},
  year={2024}
}

@inproceedings{hessel-etal-2021-clipscore,
    title = "{{CLIPS}core: A Reference-free Evaluation Metric for Image Captioning}",
    author = "Hessel, Jack  and
      Holtzman, Ari  and
      Forbes, Maxwell  and
      Le Bras, Ronan  and
      Choi, Yejin",
    editor = "Moens, Marie-Francine  and
      Huang, Xuanjing  and
      Specia, Lucia  and
      Yih, Scott Wen-tau",
    booktitle = "Proceedings of the 2021 Conference on Empirical Methods in Natural Language Processing (EMNLP)",
    month = nov,
    year = "2021",
    address = "Online and Punta Cana, Dominican Republic",
    publisher = "Association for Computational Linguistics",
    url = "https://aclanthology.org/2021.emnlp-main.595/",
    doi = "10.18653/v1/2021.emnlp-main.595",
    pages = "7514--7528",
}

@inproceedings{dettmers2022llmint8,
  title={{GPT3.int8(): 8-bit Matrix Multiplication for Transformers at Scale}},
  author={Dettmers, Tim and Lewis, Mike and Belkada, Younes and Zettlemoyer, Luke},
  booktitle={Advances in Neural Information Processing Systems (NeurIPS)},
  year={2022}
}

@inproceedings{dettmers2023qlora,
  title={{QLoRA: Efficient Finetuning of Quantized LLMs}},
  author={Dettmers, Tim and Pagnoni, Artidoro and Holtzman, Ari and Zettlemoyer, Luke},
  booktitle={Advances in Neural Information Processing Systems (NeurIPS)},
  year={2023}
}

@inproceedings{chen2025deep,
title={{Deep Compression Autoencoder for Efficient High-Resolution Diffusion Models}},
author={Junyu Chen and Han Cai and Junsong Chen and Enze Xie and Shang Yang and Haotian Tang and Muyang Li and Song Han},
booktitle={International Conference on Learning Representations (ICLR)},
year={2025},
}

@inproceedings{linear_attn_katharopoulos2020,
author = {Katharopoulos, Angelos and Vyas, Apoorv and Pappas, Nikolaos and Fleuret, Fran\c{c}ois},
title = {{Transformers are RNNs: Fast Autoregressive Transformers with Linear Attention}},
year = {2020},
booktitle = {International Conference on Machine Learning (ICML)},
}
\setcounter{secnumdepth}{2}
\newpage
\appendix
\onecolumn

\vbox{
\hsize\textwidth
\linewidth\hsize
\vskip 0.1in
\hrule height 4pt
\vskip 0.25in%
\vskip -\parskip%
\centering \LARGE\bf Supplementary Material \\
\Large \sysname: Position-Aware Compression for Large-Scale Diffusion Models \par
\vskip 0.29in
  \vskip -\parskip
  \hrule height 1pt
  \vskip 0.09in%
}

\renewcommand \thepart{}
\renewcommand \partname{}

\doparttoc %
\faketableofcontents %

\addcontentsline{toc}{section}{Appendix}
\part{}
\parttoc

\section{Detailed Experimental Setup}\label{app:detailed_exp_setup}

\subsection{Datasets}
We use the YE-POP dataset from HuggingFace for distillation training of the target models, similar to prior work by~\cite{lee2024koala}. The YE-POP dataset is comprised of 500,000 high-quality images sampled from the LAION-5B dataset~\citep{2022NeurIPSLAION5B}, incorporating a variety of visual modalities across multiple disciplines, including images of natural scenes, objects, people, and artistic images, providing extensive breadth for the training and evaluation of diffusion models (DMs). Our dataset preprocessing pipeline includes standard normalisation processes and low-resolution image and damage filtering, and ensures consistent image resolutions and aspect ratios to provide a fair data comparison for our model configurations and baseline methods.

\subsection{Architectures and Implementation}
\textbf{Target Models:} We examine the two most recent DMs: SD3.5 Large Turbo (8B parameters) and FLUX.1-Schnell (12B). Both models are state-of-the-art (SOTA) for few-step, diffusion-based T2I generation, producing high-quality high-resolution images in an average of 4 steps.

\noindent\textbf{\sysname Hyperparameters:} The pruning process is configured through hyperparameters tuned for the target architecture. For SD3.5 Large Turbo and FLUX.1-Schnell, we set $\alpha$ to 0.55 and 0.01, respectively. Hyperparameter $\alpha$ in Eq.~(\ref{eq:pos_weight_func}) controls the strength of positional importance weighting within the network. Higher values of $\alpha$ emphasise positional weighting more strongly, prioritising the removal of components based on their network position, while lower $\alpha$ values reduce this positional weighting, allowing importance scores to be determined based more on component-specific contributions rather than architectural location.

\noindent\textbf{Implementation:} The \sysname pipeline is implemented in PyTorch. In particular, we built on top of checkpoints and inference pipelines for SD3.5 Large Turbo and FLUX.1-Schnell from the \texttt{Diffusers} library ~\citep{von-platen-etal-2022-diffusers}. This allows us both consistency with pre- and post-processing steps, and also reproducible experiments. The distillation training process is conducted on a single node of 8 A100 GPUs with 80 GB of VRAM.

\noindent\textbf{Quantization Strategy:} To further reduce the memory footprint of the models, we employed 4-bit weight quantization from using \texttt{bitsandbytes}~\citep{dettmers2022llmint8,dettmers2023qlora}. %

\noindent\textbf{Open-Sourcing:} We plan to release our trained checkpoints, distillation and inference pipelines upon acceptance of the paper to assist the community to reproduce our work as well as to further advance this line of research.

\subsection{Baseline Comparisons}
We compare \sysname with (1) BK-SDM and (2) KOALA, two prior works, employing depth pruning and knowledge distillation, and (3) SANA-Sprint-1.6B, the SOTA small-scale DM.

\noindent\textbf{BK-SDM~\citep{kim2024bksdm}:} BK-SDM performs block pruning for U-Net-based diffusion using CLIP~\citep{hessel-etal-2021-clipscore} scores as importance metric, where it ranks blocks based on the reduction of CLIP scores of the output images from the model before and after pruning each block. After pruning, BK-SDM distils the compressed model using knowledge from the original unpruned model, to recover performance from the previous steps.

\noindent\textbf{KOALA~\citep{lee2024koala}:} KOALA uses an importance ranking approach by calculating input-output cosine similarity to inform the block pruning strategy for U-Net based DMs. The blocks of the transformer have intrinsic importance relative to input-output transformations by evaluating the cosine similarity between input and output representations/features rather than utilising an external quality metric of transformation. %

\noindent\textbf{SANA-Sprint-1.6B~\citep{chen2025sanasprintonestepdiffusioncontinuoustime}:} We use SANA-Sprint-1.6B as the state-of-the-art (SOTA) baseline for an efficient, small-scale diffusion purpose-built model, incorporating techniques such as deep-compression auto-encoder~\citep{chen2025deep} and linear attention~\cite{linear_attn_katharopoulos2020}. Intrinsically, SANA models require training of the transformer from scratch, which departs from compressing existing models. %

All comparisons to the baselines and SANA-Sprint-1.6B are made with identical experimental protocols. We ensure all methods use the same prompt sets, generation parameters, and hardware to isolate confounded variables and allow performance attribution of each approach to the specific methodologies.

\subsection{Evaluation Framework}
\textbf{Quantitative Metrics:} We adopt two representative and complementary image quality assessment metrics, capturing two different aspects of the quality of generation. First, \textbf{GenEval}~\citep{ghosh2023geneval} provides an all-encompassing evaluation across three areas, \textit{i.e.},~semantic fidelity, visual fidelity, and prompt adherence. GenEval is now widely used for evaluations of T2I models~\citep{xie2025sana,esser2024mmdit}. Second, \textbf{HPSv2}~\citep{wu2023human} provides a contrasting perspective on image quality assessment, with demonstrated effectiveness in estimating the perceptual preferences of humans about image quality. HPSv2 has been utilised in several recent works such as KOALA~\citep{lee2024koala}.

\noindent\textbf{Human Preference Metrics:}
We conduct a user study to assess human preferences for the generated images that are difficult to capture with the quantitative quality metrics (see Appendix~\ref{app:subsec:user_study} for further details regarding our user study design). Similarly to~\cite{sheynin2023emueditpreciseimage} and \cite{dai2023emuenhancingimagegeneration}, we evaluate two aspects: (1)~\textbf{Text Alignment}: How well a generated image follows the description of the text prompt, (2)~\textbf{Image Quality}: The overall visual quality considering clarity, colour, composition, and other factors.

\noindent\textbf{Hardware Testing Environment:} On the system resource side, we report the measured peak memory usage and latency of running the target models on GPUs. Our evaluation was conducted across three different GPUs, representing deployment scenarios on cloud (A100 with 80GB), high-end consumer (A6000 with 48GB), and resource-constrained low-end GPUs (GTX 3090 with 24GB).

\begin{figure*}[t]
    \centering
    \subfloat{
    \includegraphics[width=0.80\textwidth]{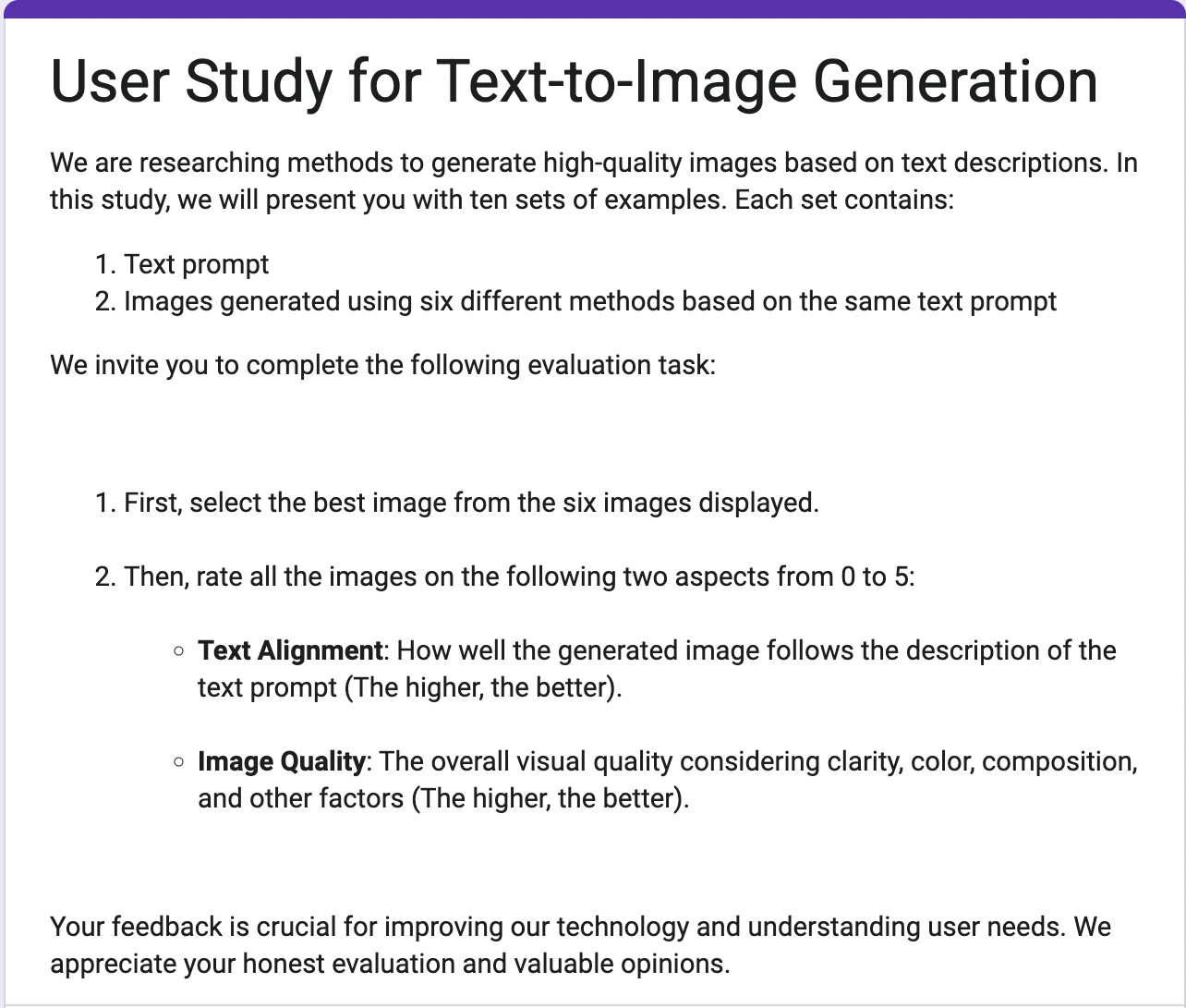}
    }
    \hfill
    \centering
    \subfloat{
    \includegraphics[width=0.41\textwidth]{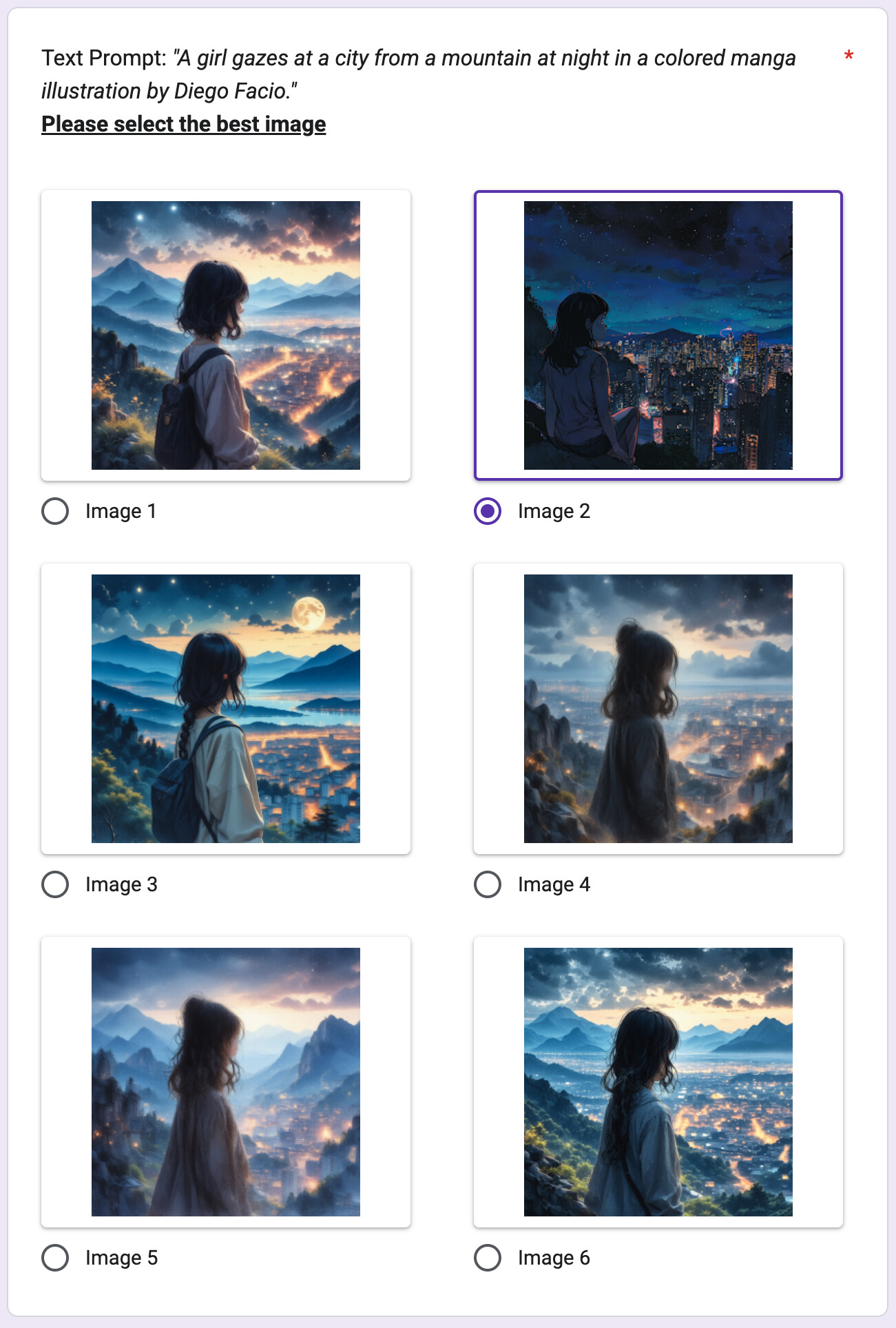}
    }
    \subfloat{
    \includegraphics[width=0.40\textwidth]{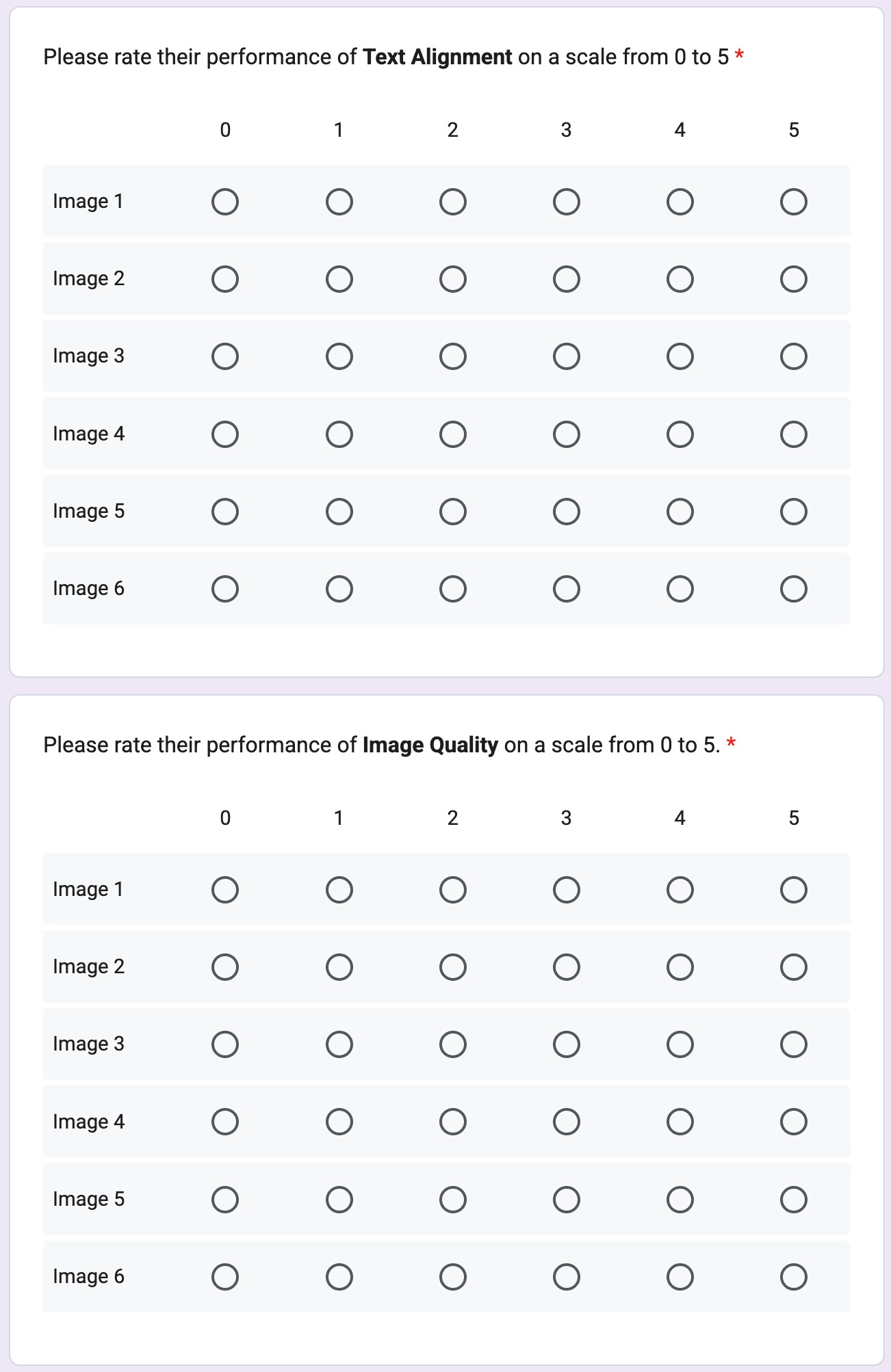}
    }
    \caption{The user study interface that describes the objectives, instructions, evaluation criteria, and example image samples.}
    \label{fig:user_study_interface}
\end{figure*}

\subsection{Detailed User Study Design}\label{app:subsec:user_study}
Acknowledging the inconsistency of automated metrics, especially in discerning fine aspects of image quality and preferences for aesthetics, we choose to conduct a full user study of methods following the protocol of \cite{sheynin2023emueditpreciseimage} and \cite{dai2023emuenhancingimagegeneration}. In our user evaluation method, we are focused on measuring two aspects of user experience:
\begin{itemize}
\item \textbf{Text Alignment}: Evaluators measure the extent to which the generated images reflected correctly the content item or detail in the input text prompts. Within this measure, evaluators take into account whether the required object appeared and its spatial relationship with other objects, about detail examined before the invocation of the process, and attributes/stylistic characteristics specified in the prompts.
\item \textbf{Image Quality}: We ask the evaluator to judge the quality on a multi-factor basis, \textit{e.g.}, clarity of image, colour accuracy and vibrancy, balance of composition (or lack thereof), uniformity of lighting, preservation of texture detail, and the absence of artefacts or distortion.
\end{itemize}

\noindent\textbf{Experimental Protocol and Interface Design:} Based on these evaluation criteria, we designed an online questionnaire that rigorously collects systematic and unbiased data. Figure~\ref{fig:user_study_interface} shows the user study interface, which provides a consistent and structured layout and enables a progressive evaluation workflow.

First (see bottom-left figure), participants were shown six images generated by six different methods using the same text prompt (displayed in grid format with adequate visual separation) and required to select the single best image based on their holistic review of the image quality. This can create an overall perceived quality ranking among the evaluated generation methods. Second (see bottom-right figure), participants rated images individually based on two aspects of user experience as discussed above using structured rating scales. Specifically, each image was rated in isolation, on both the Text Alignment and Image Quality dimensions using a 6-point Likert scale (0-5). Zero = Poor performance, and 5 = Excellent performance.

\noindent\textbf{Randomisation Strategy and Bias Mitigation:} Randomisation is necessary for the performance differences to be based on the quality of the generated images, not due to positional bias, order, or pattern recognised by the evaluator. Therefore, to mitigate bias and avoid participants discovering patterns in how we generated the order of images, we randomly ordered the presentation of images across all evaluation sessions. Specifically, each participant would see the images in different random sequences, and there were no fixed positional relationships between methods across prompt sets. Randomisation occurred in the image grid as well as the rating phases, thus maintaining the fair assessment of generated image quality across different methods.

\noindent\textbf{Quality Control:} To ensure that our assessment criteria are interpreted in the same way, participants receive introduction materials before starting the formal evaluation. These materials have detailed descriptions of both the Text Alignment and Image Quality dimensions. The questionnaire has some built-in systems to ensure data quality and participant attention across multiple evaluation sessions. The platform sets out clear progress indicators to orientate the participants and maintain their attention. Also, the platform requires fully completed ratings for all images in both dimensions before the participant can move on to the next prompt set.

\begin{figure*}[t!]
  \centering
  \subfloat[Block Removal]{
    \includegraphics[width=0.325\textwidth]{figures/stable-diffusion-3.5-large-turbo_hpsv2_All_extended_2.pdf}
  \label{subfig:block_removal_supp}}
  \subfloat[Multi-modal Attention]{
    \includegraphics[width=0.325\textwidth]{figures/stable-diffusion-3.5-large-turbo_hpsv2_Attn_extended_2.pdf}
  \label{subfig:attn_removal_supp}}
  \subfloat[MLP]{
    \includegraphics[width=0.325\textwidth]{figures/stable-diffusion-3.5-large-turbo_hpsv2_FF_extended_2.pdf}
  \label{subfig:mlp_removal_supp}}
  \hfill
  \centering
  \subfloat[Norm]{
    \includegraphics[width=0.325\textwidth]{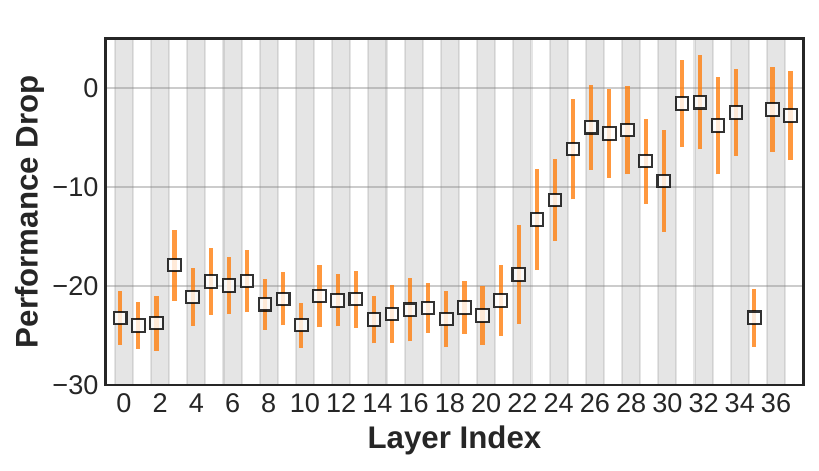}
  \label{subfig:norm_removal_supp}}
  \subfloat[Context Norm]{
    \includegraphics[width=0.325\textwidth]{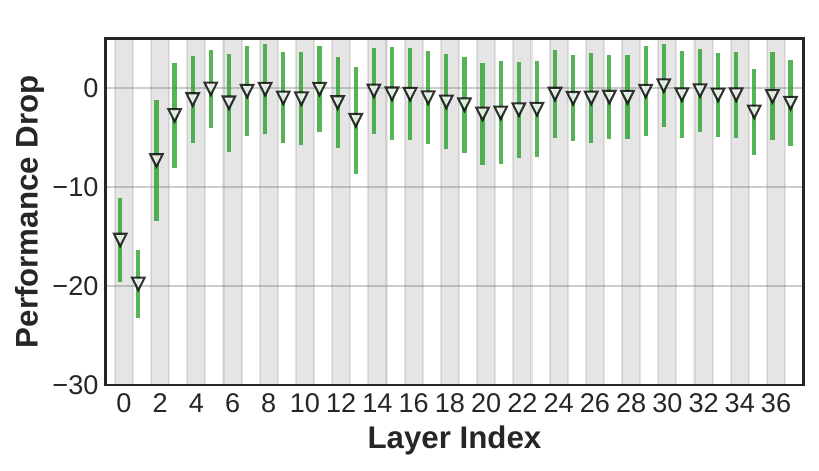}
  \label{subfig:norm_context_removal_supp}}
  \subfloat[Context MLP]{
    \includegraphics[width=0.325\textwidth]{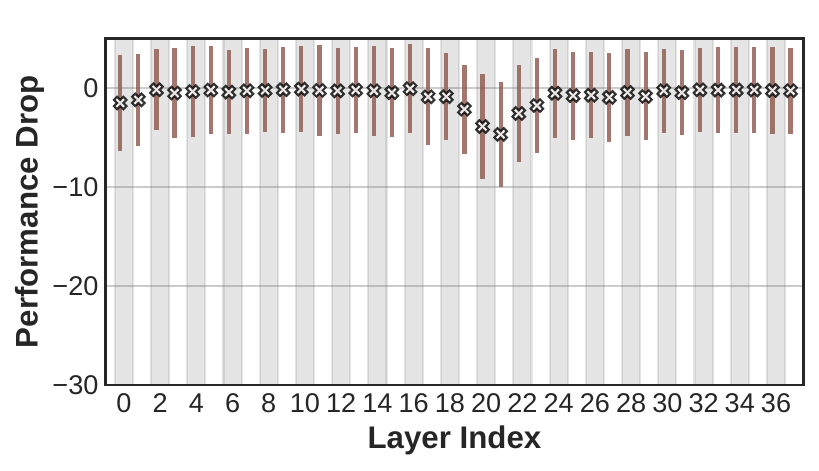}
  \label{subfig:mlp_context_removal_supp}}
  \caption{
  Fine-grained contribution analysis of SD3.5 Large Turbo on the HPSv2 dataset by removing either an entire MMDiT block or an intra-block subcomponent. We report the performance drop compared to the original model. (a) An entire MMDiT block is removed following prior depth pruning approaches~\citep{lee2024koala,kim2024bksdm,fang2024tinyfusion}. (b,c,d,e,f) Each subcomponent of a MMDiT block is removed, revealing the different patterns of importance of each subcomponent.
  }
  \label{fig:contribution_analysis_full}
\end{figure*}

\begin{figure*}[t!]
  \centering
  \subfloat[Norm+MLP]{
    \includegraphics[width=0.325\textwidth]{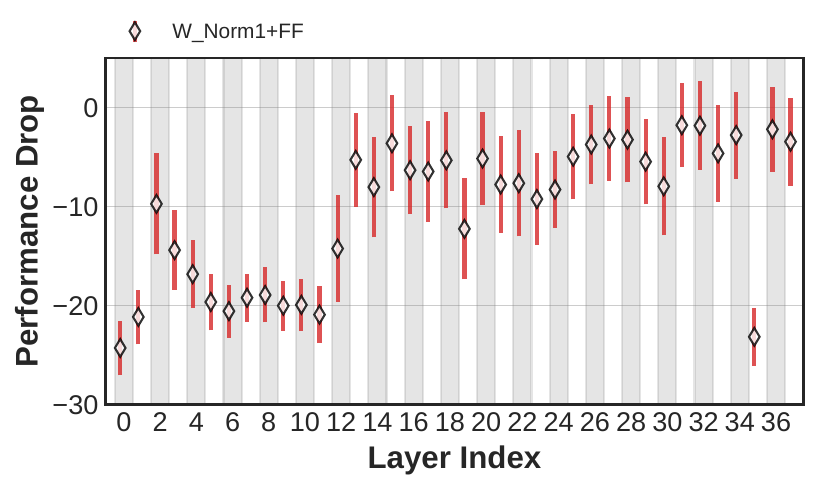}
  \label{subfig:fg_anal_1}}
  \subfloat[MLP \& Context MLP]{
    \includegraphics[width=0.325\textwidth]{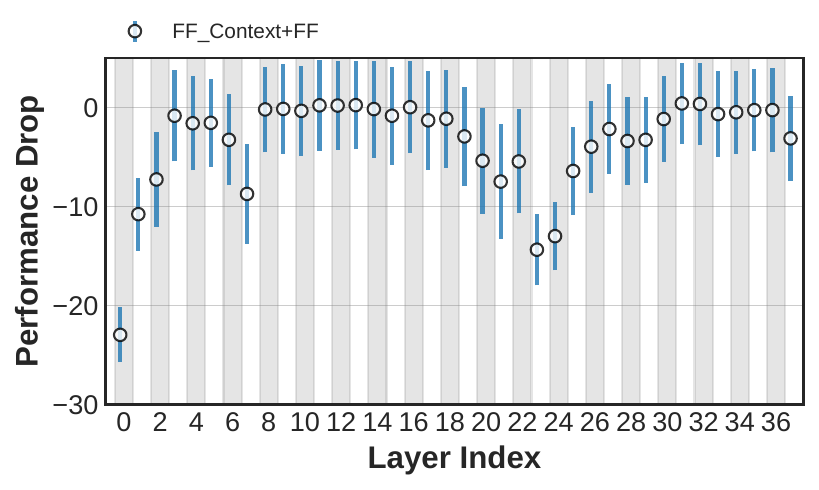}
  \label{subfig:fg_anal_2}}
  \subfloat[Norm \& Context Norm]{
    \includegraphics[width=0.325\textwidth]{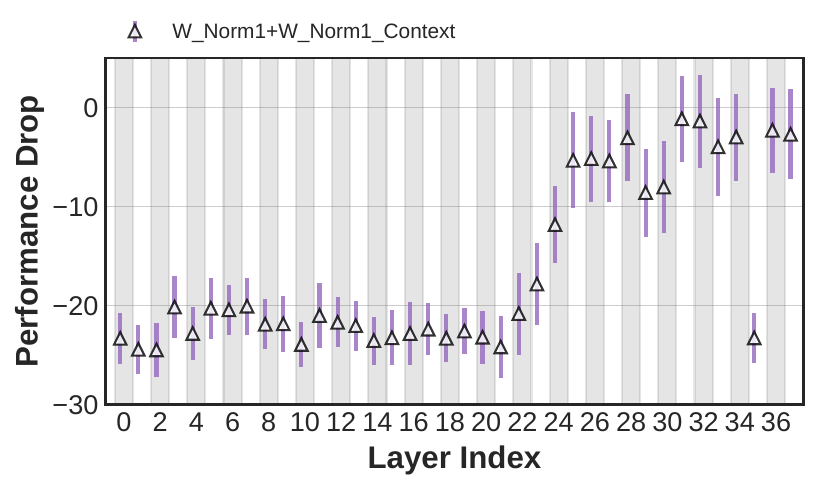}
  \label{subfig:fg_anal_3}}
  \hfill
  \centering
  \subfloat[Context Norm \& Context MLP]{
    \includegraphics[width=0.325\textwidth]{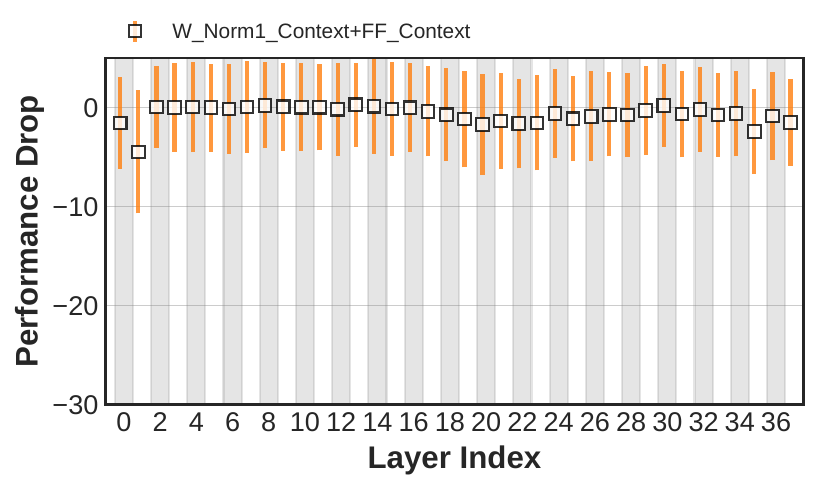}
  \label{subfig:fg_anal_4}}
  \subfloat[Context Norm \& All MLP]{
    \includegraphics[width=0.325\textwidth]{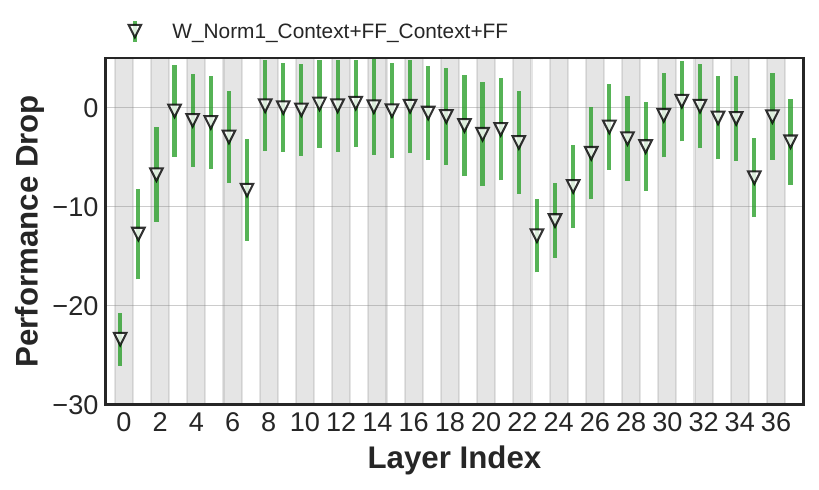}
  \label{subfig:fg_anal_5}}
  \subfloat[All Norm \& All MLP]{
    \includegraphics[width=0.325\textwidth]{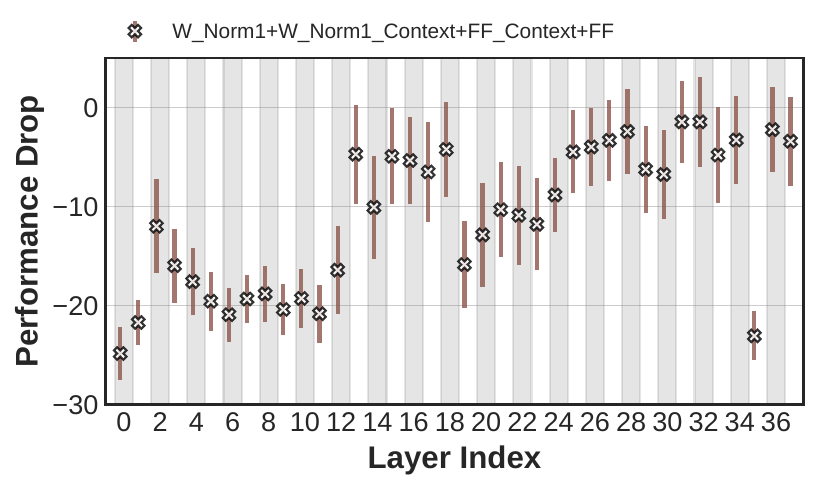}
  \label{subfig:fg_anal_6}}
  \caption{
  Fine-grained contribution analysis of SD3.5 Large Turbo on the HPSv2 dataset by jointly removing multiple subcomponent types. The performance drop follows different patterns for different pairs, with most type combinations leading to high performance degradation in earlier parts of the network, except for Context Norm + Context MLP (d), which has minimal impact.
  }
  \label{fig:fg_contribution_analysis}
\end{figure*}

\section{Additional Analysis}\label{app:additional_analysis}

In this section, we provide additional analysis results that are not included in the main content of the paper due to the page limit.

\subsection{SD3.5 Large Turbo}
\noindent\textbf{Individual Subcomponent Removal Analysis:}
In additional to Figure~\ref{fig:contribution_analysis}, where we showed full-block removal and two types of subcomponent removals, we include in Figure~\ref{fig:contribution_analysis_full} three other types of subcomponents (d-f).

\noindent\textbf{Joint Subcomponent Removal Analysis:}
To further understand the interdependencies between different MMDiT subcomponents, we conducted an analysis examining the effect of jointly removing multiple subcomponents from SD3.5 Large Turbo. Figure~\ref{fig:fg_contribution_analysis} shows the performance degradation patterns when pairs of subcomponents are simultaneously removed from the model.

The results reveal distinct interaction patterns between different subcomponent combinations. Many of the paired subcomponent removals exhibit significant performance degradation concentrated in the earlier blocks of the model, consistent with the critical role of early processing stages in establishing semantic structures for image generation. However, a notable exception emerges with the combination of Context Norm and Context MLP, which demonstrates remarkably minimal impact across the entire network depth. This suggests that these two components may have complementary or redundant functions within the context processing pathway, allowing the model to maintain performance when both are simultaneously removed.

The varied degradation patterns across different subcomponent pairs indicate that the MMDiT architecture contains both critical interdependent components and potentially redundant pathways, providing insights for targeted pruning strategies that preserve model quality while reducing computational overhead.

\begin{figure*}[t!]
  \centering
  \subfloat[Block Removal]{
    \includegraphics[width=0.325\textwidth]{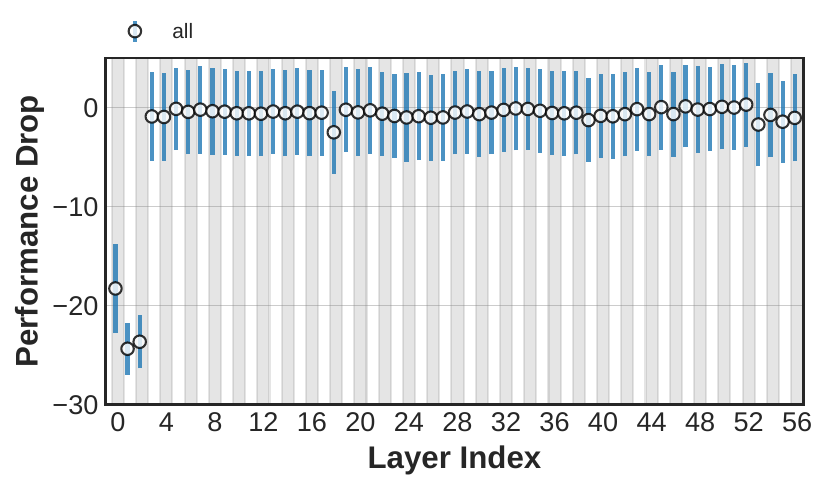}
  \label{subfig:flux_block_removal}}
  \subfloat[Multi-modal Attention]{
    \includegraphics[width=0.325\textwidth]{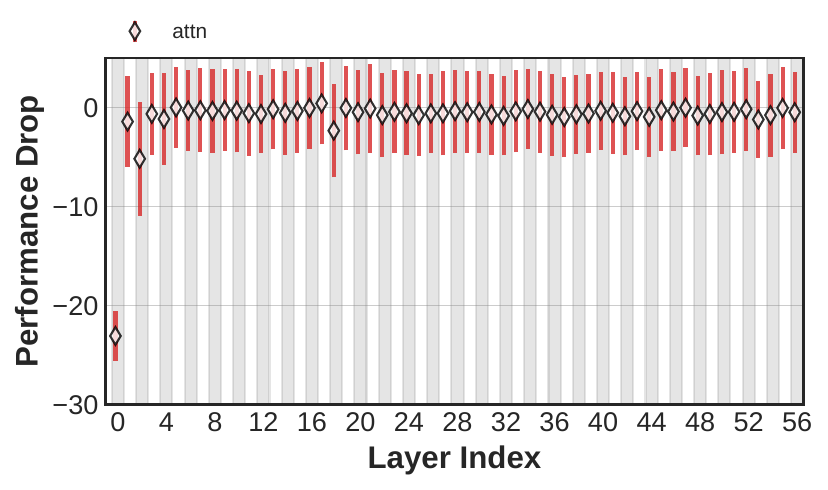}
  \label{subfig:flux_attn_removal}}
  \subfloat[MLP]{
    \includegraphics[width=0.325\textwidth]{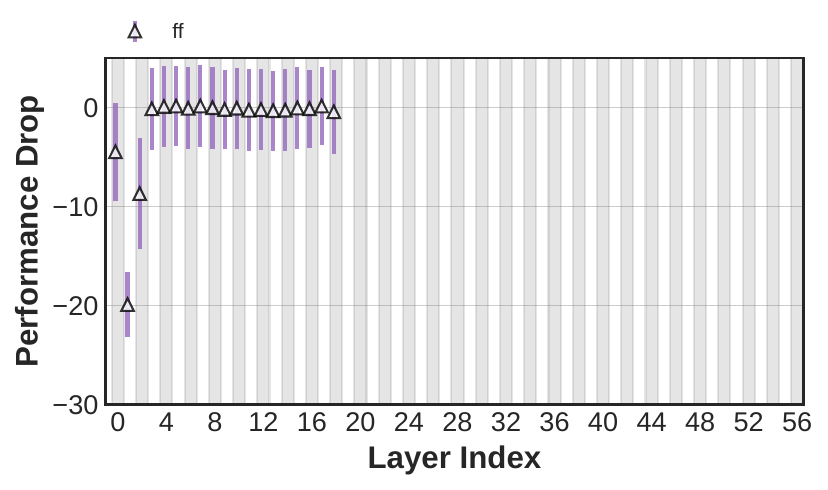}
  \label{subfig:flux_mlp_removal}}
  \hfill
  \centering
  \subfloat[Norm]{
    \includegraphics[width=0.325\textwidth]{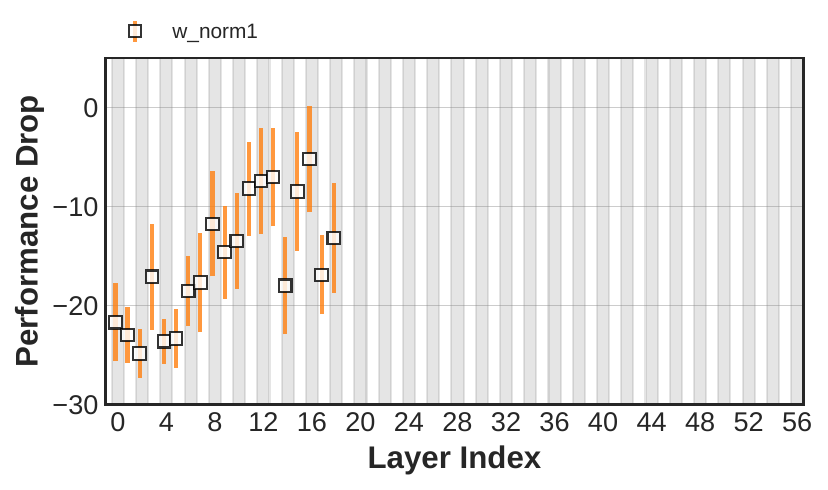}
  \label{subfig:flux_norm1_removal}}
  \subfloat[Context Norm]{
    \includegraphics[width=0.325\textwidth]{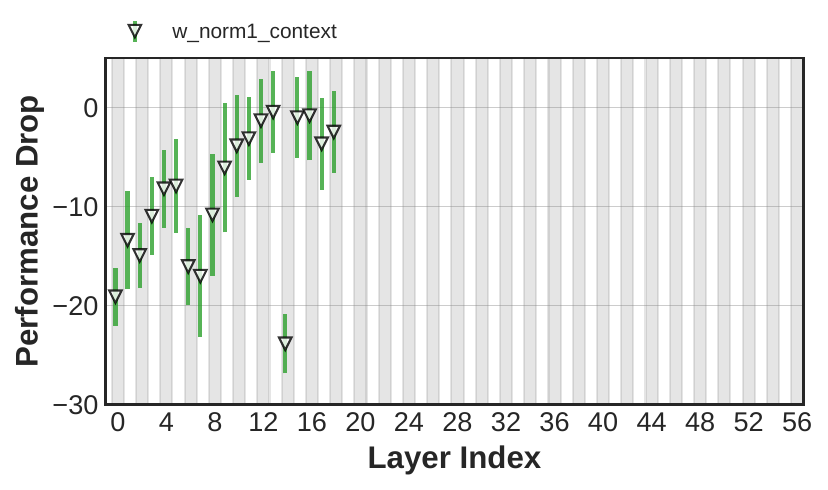}
  \label{subfig:flux_norm_context_removal}}
  \subfloat[Context MLP]{
    \includegraphics[width=0.325\textwidth]{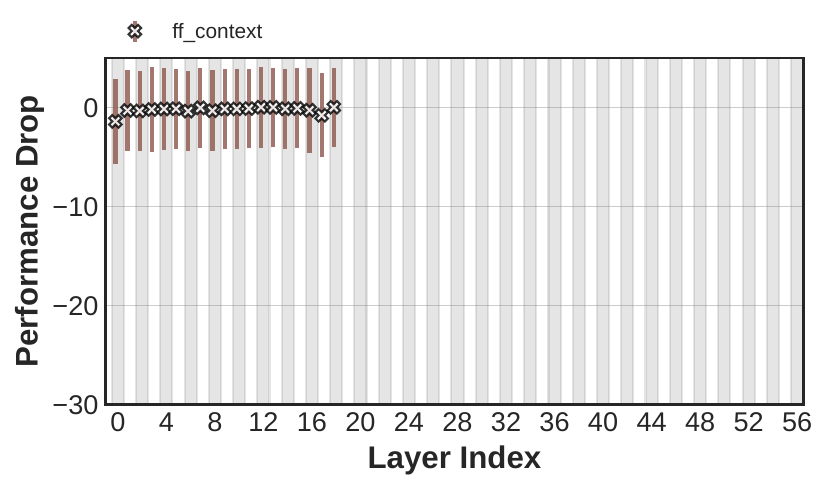}
  \label{subfig:flux_mlp_context_removal}}
  \hfill
  \centering
  \subfloat[(Single Transformer) Norm]{
    \includegraphics[width=0.325\textwidth]{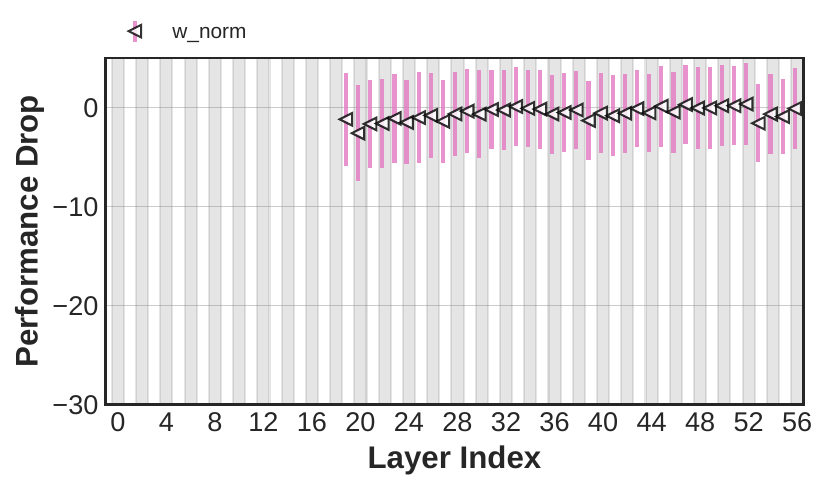}
  \label{subfig:flux_norm_removal}}
  \subfloat[(Single Transformer) MLP \& Proj]{
    \includegraphics[width=0.325\textwidth]{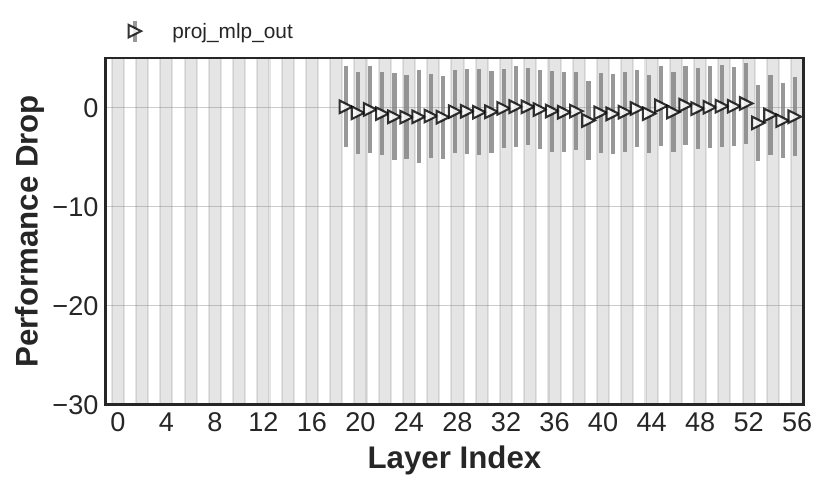}
  \label{subfig:flux_proj_mlp_out_removal}}
  \caption{
  Fine-grained contribution analysis of FLUX.1-Schnell on the HPSv2 dataset by removing either an entire MMDiT block or an intra-block subcomponent. We report the performance drop compared to the original model. (a) An entire MMDiT block is removed following prior depth pruning approaches~\citep{lee2024koala,kim2024bksdm,fang2024tinyfusion}. (b,c,d,e,f,g,h) Each subcomponent of a MMDiT block is removed, revealing the different patterns of importance of each subcomponent. Note that the first part of the model from block 0 to 18 is composed of the transformer blocks with five components (b-f) and the second part of the model is made of single transformer blocks with three components (b,g,h). \vspace{-0.3cm}
  }
  \label{fig:flux_contribution_analysis}
\end{figure*}

\begin{figure*}[t!]
  \centering
  \subfloat[Norm+MLP]{
    \includegraphics[width=0.325\textwidth]{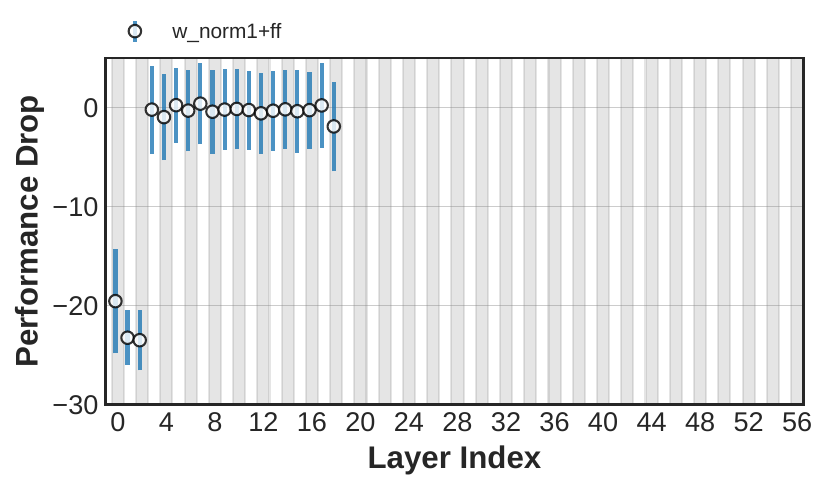}
  \label{subfig:flux_fg_anal_1}}
  \subfloat[Norm \& Context Norm]{
    \includegraphics[width=0.325\textwidth]{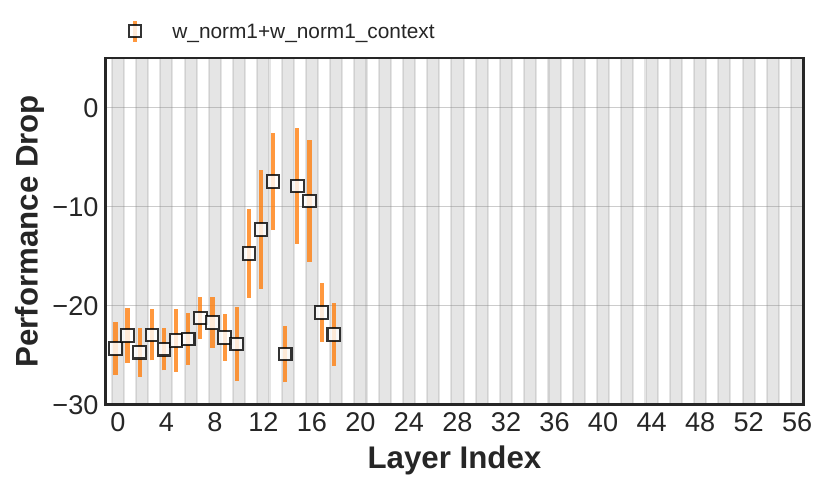}
  \label{subfig:flux_fg_anal_2}}
  \subfloat[Norm Context \& Context MLP]{
    \includegraphics[width=0.325\textwidth]{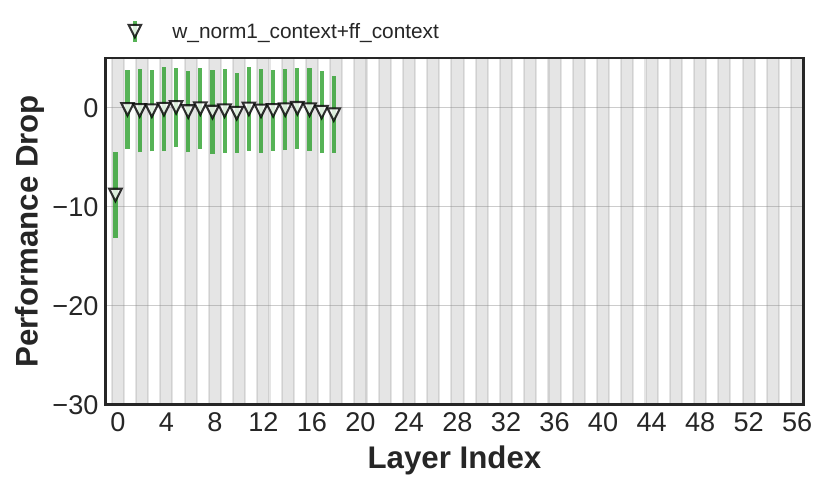}
  \label{subfig:flux_fg_anal_3}}
  \hfill
  \centering
  \subfloat[(Single Transformer) Attn \& MLP]{
    \includegraphics[width=0.325\textwidth]{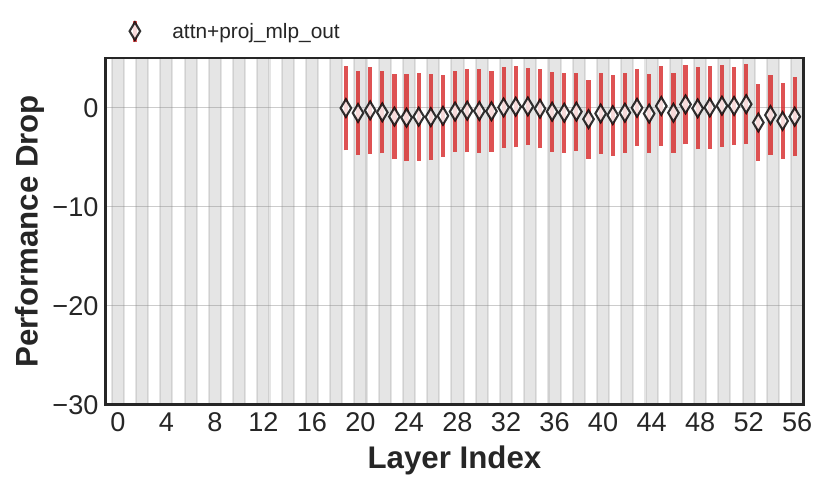}
  \label{subfig:flux_fg_anal_4}}
  \subfloat[(Single Transformer) Norm \& Attn]{
    \includegraphics[width=0.325\textwidth]{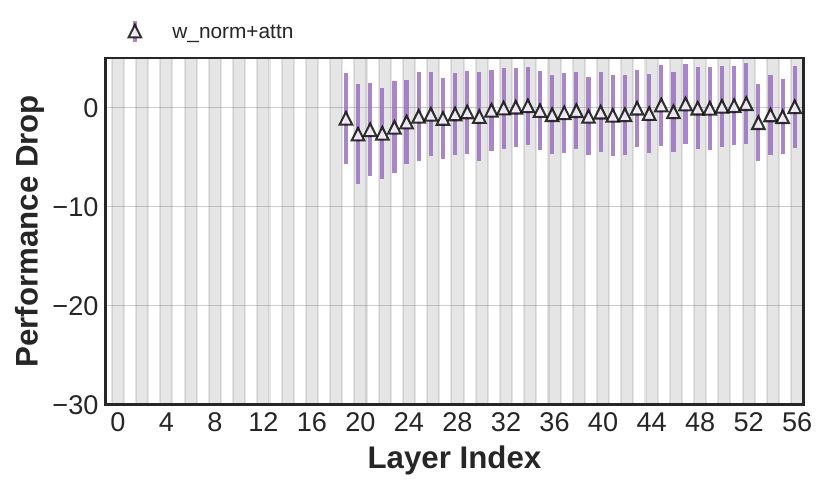}
  \label{subfig:flux_fg_anal_5}}
  \subfloat[(Single Transformer) Norm \& MLP]{
    \includegraphics[width=0.325\textwidth]{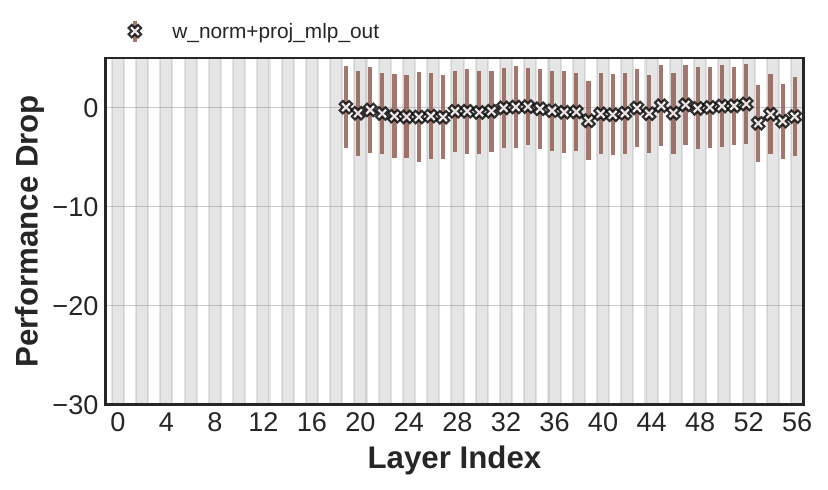}
  \label{subfig:flux_fg_anal_6}}
  \caption{
  Fine-grained contribution analysis of FLUX.1-Schnell on the HPSv2 dataset by jointly removing multiple subcomponent types. The performance drop follows different patterns for different pairs for the transformer blocks (a,b,c) and for the single transformer blocks (e,d,f). For the transformer blocks (a,b,c), it shows that removing the earlier part of the network leads to huge degradation of the quality for all the combinations of the subcomponents and removing all the norms (b) affects negatively across all the portion of the network, except for Norm + MLP (a) and Context Norm + Context MLP (c), which has smaller impact.
  For the single transformer blocks, removing subcomponents have minimal impacts on the final outputs as shown in (e,d,f).
  }
  \label{fig:flux_fg_contribution_analysis}
\end{figure*}

\subsection{FLUX.1-Schnell}

\noindent\textbf{Individual Subcomponent Removal Analysis:}
We extended our analysis to FLUX.1-Schnell, which exhibits a unique two-stage architecture design. The model consists of blocks 0-18 containing transformer blocks with five subcomponents, followed by single transformer blocks with three subcomponents in the latter portion of the network.

Figure~\ref{fig:flux_contribution_analysis} presents the contribution analysis results for FLUX.1-Schnell. The architectural heterogeneity of this model provides additional insights into the role of different subcomponents across varying block types. The transformer blocks (Figures~\ref{subfig:flux_block_removal}-\ref{subfig:flux_mlp_context_removal}) show sensitivity patterns similar to those observed in SD3.5 Large Turbo, especially for Norm subcomponent, while the single transformer blocks (Figures~\ref{subfig:flux_norm_removal}-\ref{subfig:flux_proj_mlp_out_removal}) shows that the performance degradation is largely small.

\noindent\textbf{Joint Subcomponent Removal Analysis:}
The joint removal analysis for FLUX.1-Schnell reveals architecture-specific patterns that differ markedly between the two block types. As shown in Figure~\ref{fig:flux_fg_contribution_analysis}, for transformer blocks, the results demonstrate that removing components from earlier network portions leads to substantial quality degradation across all subcomponent combinations. Particularly notable is the impact of removing all normalisation components, which negatively affects performance throughout the network depth, with the exception of Norm + MLP and Context Norm + Context MLP combinations, which show reduced impact.

In contrast, the single transformer blocks exhibit remarkable resilience to subcomponent removal, with minimal impacts on final output quality. This robustness suggests that the single transformer blocks may contain significant redundancy or that their simpler architecture naturally provides more fault tolerance. The differential sensitivity between block types reinforces the hierarchical nature of the FLUX.1-Schnell architecture, where early complex processing is critical while later simplified processing is more robust to subcomponent removal.

\section{Additional Results}\label{app:additional_results}
We present additional results that are not included in the main content of the paper due page limit.

\subsection{Computational Overhead Analysis}\label{app:additional_results:cost_analysis}
We investigate the computational overhead of using \sysname. The overhead is primarily composed of two phases: (1) the contribution analysis for establishing hierarchical importance patterns and (2) the subsequent distillation process.

\noindent\textbf{Contribution Analysis Overhead:}
The contribution analysis phase represents the foundational step of our methodology, systematically evaluating architectural components to establish hierarchical importance patterns. This analysis involves removing individual blocks or subcomponents and measuring the resulting performance degradation. The computational requirements varied across architectures:

\begin{itemize}
    \item \textbf{SD3.5 Large Turbo:} Block-wise analysis required 6.3 hours (400 images per block). Fine-grained subcomponent analysis took an additional 6.3 hours (80 images per subcomponent).
    \item \textbf{FLUX.1-Schnell:} Block-wise analysis required 14.9 hours (400 images per block) and fine-grained analysis took an additional 12.2 hours (80 images per subcomponent).
\end{itemize}

The contribution analysis takes around 12.6 A100 GPU hours for SD3.5 Large Turbo and 27.1 A100 GPU hours for FLUX.1-Schnell. It systematically examines 38 blocks and 190 subcomponent combinations for SD3.5 Large Turbo, and 57 blocks with multiple subcomponent configurations for FLUX.1-Schnell (e.g., 19 transformer blocks with 5 components each, plus 38 single transformer blocks with 3 components each). The increased overhead for FLUX.1-Schnell is due to its higher inference cost and architectural complexity.

Overall, the contribution analysis phase requires 12.6-27.1 GPU hours for comprehensive architectural profiling. While substantial, this represents a one-time cost that is amortised across all deployments. Furthermore, training small-scale DMs (1-2B parameter-sized) from scratch requires 140,000 and/or 200,000 A100 GPU hours for smaller models such as SD1.4 and SD2.1, respectively, as reported in~\cite{2024ICLRgU58d5QeGv_cascade}, let alone large-scale DMs (8-11B parameter-sized), which could require significantly more GPU hours to train from scratch. Yet, our contribution analysis cost equals <0.006\% of typical DM training.

\noindent\textbf{Distillation Training Overhead:}
Following the contribution analysis, the distillation training process represents the most computationally intensive component of our methodology, requiring approximately 603 A100 GPU hours for SD3.5 Large Turbo and 1260 A100 GPU hours for FLUX.1-Schnell. The computational requirement reflects the complexity of preserving image quality during aggressive compression, as our hierarchical distillation approach meticulously balances knowledge transfer between original and compressed models while respecting the positional sensitivity patterns identified through contribution analysis.

\begin{table*}[t]
  \centering

  \resizebox{0.8\columnwidth}{!}{%
  \begin{tabular}{ p {1.7cm} l | c | c c | c }
    \toprule
     \textbf{Model} & \textbf{Method} & \textbf{Memory (\%)} & \textbf{GenEval $\Uparrow$} & \textbf{HPSv2 $\Uparrow$} & \textbf{Reduction $\Downarrow$} \\
        \cmidrule(l){1-6}
   Linear DiT & SANA-Sprint & (100\%)             & 0.77 & 29.61 ($\pm$ 0.071) & - \\
        \cmidrule(l){1-6}
    & Original                       & (100\%) & 0.71 & 30.29 ($\pm$ 0.074) & - \\
        \cmidrule(l){2-6}
          & KOALA                    & (79.4\%)	 & 0.37 & 19.99 ($\pm$ 0.074) & 41.2\% \\
          & KOALA (+Quant)           & (22.5\%)	 & 0.33 & 18.44	($\pm$ 0.075) & 46.4\% \\
           & BK-SDM                   & (79.4\%)	 & 0.38 & 21.21 ($\pm$ 0.077)	& 38.2\% \\
          & BK-SDM (+Quant) & (22.5\%)	 & 0.34 & 19.83 ($\pm$ 0.079)	& 43.3\% \\
        \cmidrule(l){2-6}
     SD3.5 & Ours  (HPP)     & (79.4\%)              & 0.03 & 11.08 ($\pm$ 0.042) & 79.4\% \\
     Large Turbo  & Ours  (+PWP) & (79.4\%)  & \textbf{0.71} & \textbf{28.97} ($\pm$ 0.077) & \textbf{2.5}\% \\
      & Ours  (+Quant)     & (\textbf{22.5}\%)  & 0.69          & 28.15 ($\pm$ 0.078) & 4.8\% \\
    \cmidrule(l){2-6}
      & Ours  (HPP)     & (71.5\%)              & 0.0 & 7.00 ($\pm$ 0.041) & 88.4\% \\
      & Ours  (+PWP)     & (71.5\%)              & 0.46 & 21.74 ($\pm$ 0.075) & 31.9\% \\
       & Ours  (+SGDistill)     & (71.5\%)	           & \textbf{0.64} & \textbf{27.29} ($\pm$ 0.079) & \textbf{10.1}\% \\
       & Ours (+Quant, All)         & (\textbf{20.5}\%)	           & 0.62 & 26.29 ($\pm$ 0.082) & 13.3\% \\
        \cmidrule(l){1-6}
    & Original     & (100\%)             & 0.66 & 29.71 ($\pm$ 0.073) & - \\
        \cmidrule(l){2-6}
    FLUX.1  & KOALA          & (70.5\%)	     & 0.38 & 25.24 ($\pm$ 0.075)	& 28.7\% \\
    Schnell & BK-SDM         & (70.5\%)	     & 0.45 & 27.38	($\pm$ 0.074) & 19.8\% \\
        \cmidrule(l){2-6}
       & Ours (All)             & (\textbf{19.6}\%)        & \textbf{0.64} & \textbf{28.69} ($\pm$ 0.079) & \textbf{3.2}\% \\
    \bottomrule
  \end{tabular}
  }
    \caption{Full quantitative evaluation of image quality with mean and standard error. Note that the standard error is presented only for HPSv2, as GenEval is the metric that reports the total number of correctly classified images with different attributes (standard error is not applicable to the GenEval metric). KOALA and BK-SDM experience substantial degradation of image quality when reducing memory usage by 20-30\%. \sysname achieves significantly reduced memory usage while maintaining the image quality close to the original models.}
  \label{app:tab:quality_metric}
\end{table*}

\noindent\textbf{Overall Computational Requirement:}
When considered in its entirety, the contribution analysis phase requires 12.6-27.1 GPU hours for comprehensive architectural profiling across different model architectures, and the subsequent distillation pipeline demands 603-1260 A100 GPU hours. While substantial, these represent one-time costs that amortise across all subsequent deployments. To place this computational investment in perspective, training small-scale DMs (1-2B parameters) from scratch requires 140,000-200,000 A100 GPU hours for SD1.4 and SD2.1 models, respectively, as reported~\cite{2024ICLRgU58d5QeGv_cascade}, let alone large-scale DMs (8-11B parameters), which would require considerably more computational resources. Notably, our contribution analysis cost equals less than 0.006\% of typical DM training overhead, while our distillation training process represents less than 0.30-0.63\% of the computational cost required to train comparable models from scratch.

This perspective emphasises the efficiency of our compression approach: rather than training new compact models that could sacrifice quality, \sysname leverages existing high-quality models and transforms them into deployable versions through targeted compression. Furthermore, as shown in Section~\ref{sec:evaluation}, the reduced computational requirement of the compressed DMs shows practical benefits, enabling 77.5-80.4\% memory reduction and 27.9-38.0\% latency improvements while preserving the superior image quality that large-scale DMs provide. Overall, this cost-benefit strongly favours compression-based approaches over training compact models from scratch, particularly when deployment accessibility and quality preservation are paramount considerations.

\subsection{Comprehensive Experimental Results}
In this subsection, we provide the comprehensive experimental results.

Table~\ref{app:tab:quality_metric} shows all the results presented in Section~\ref{sec:evaluation}, aggregating main quantitative results and ablation study results, regarding quantitative quality metrics such as GenEval and HPSv2. Note that we also present standard error to provide statistical evidence for robust performance evaluation. The standard error is presented only for HPSv2 as GenEval is the metric that reports the total number of correctly classified images with different attributes, hence the standard error is not applicable to the GenEval metric.

In terms of computational resources, as shown in Tables~\ref{tab:quality_metric}, ~\ref{tab:latency_measurements} and~\ref{app:tab:latency_full}, our compression pipeline achieves 79.5\% reduction in peak memory for SD3.5 Large Turbo i.e., from 15.8 GB to just 3.24 GB, and 80.4\% for FLUX.1.Schnell, i.e. from 22.6GB to 4.44GB.

For latency measurements, we used three testing environments as described in Appendix~\ref{app:detailed_exp_setup}, and we ran inference on the HPSv2 dataset, using 400 prompts sampled equally from all four categories. We compare \sysname with KOALA and BK-SDM, each incorporated with SD3.5 Large Turbo and FLUX.1-Schnell model. We have also included SANA-Sprint-1.6B as a strong baseline.

In Table~\ref{app:tab:latency_full}, we report the mean and standard error of the per-step latency of the MMDiT inferences, as well as the reduction compare to each base model.
We observe that \sysname constantly achieves the highest latency reduction for both base models across representative hardware platforms for cloud (A100), high-end professional (A6000), and consumer-grade edge (GTX3090) GPUs.

\begin{table}[t]
  \centering

  \resizebox{1.02\columnwidth}{!}{%

\begin{tabular}{llc|cc|cc|cc} \toprule
\textbf{}                                               & \textbf{} & \textbf{}      & \multicolumn{2}{c|}{\textbf{A100}}      & \multicolumn{2}{c|}{\textbf{A6000}}     & \multicolumn{2}{c}{\textbf{GTX 3090}}                         \\

\textbf{Model}                                          & \textbf{Method} & \textbf{Memory $\Downarrow$} & \textbf{Latency $\Downarrow$} & \textbf{Reduction $\Uparrow$} & \textbf{Latency $\Downarrow$} & \textbf{Reduction $\Uparrow$} & \textbf{Latency $\Downarrow$} & \textbf{Reduction $\Uparrow$} \\\hline

Linear DiT   & SANA-Sprint   & 3.14 GB  & 37$\pm$0.5 ms       & -       & 54$\pm$0.4 ms    & -       & 54$\pm$0.7 ms       & - \\ \hline
\multirow{4}{*}{\begin{tabular}[c]{@{}l@{}}SD3.5 \\ Large Turbo\end{tabular}} & Original     & 15.8 GB   & 444$\pm$0.4 ms       & -       & 823$\pm$0.7 ms  &    & 922$\pm$3.8 ms     & - \\ \cline{2-9}
& KOALA         & 11.3 GB  & 348$\pm$0.3 ms         &    21.6\%       & 637$\pm$0.6 ms  & 22.6\%  & 816$\pm$0.7 ms         & 11.5\% \\
& BK-SDM         & 11.3 GB & 348$\pm$0.3 ms         &    21.6\%    & 633$\pm$0.6 ms  & 23.1\%  & 816$\pm$0.7 ms       & 11.5\% \\ \cline{2-9}
  & Ours (All)    & \textbf{3.24 GB} & \textbf{341$\pm$0.3 ms}                    & \textbf{23.2\%}                    & \textbf{593$\pm$0.6 ms}               & \textbf{27.9\%}               & \textbf{754$\pm$0.7 ms}                    & \textbf{18.2\%}                    \\ \midrule
\multirow{4}{*}{FLUX.1 Schnell}                         & Original       & 22.6 GB & 430$\pm$2.7 ms      & - & 756$\pm$4.0 ms  & -       & 997$\pm$4.6 ms      & - \\\cline{2-9}
& KOALA           & 15.9 GB& 221$\pm$2.0 ms & 48.6\%       & 432$\pm$3.9 ms & 42.9 \% & 562$\pm$4.4 ms      & 43.6\% \\
& BK-SDM          & 15.9 GB & 221$\pm$2.0 ms   & 48.6\%      & 432$\pm$3.9 ms   & 42.9 \% & 562$\pm$4.4 ms      & 43.6\%                        \\\cline{2-9}
  & Ours (All)  & \textbf{4.44 GB}  & 245$\pm$1.9 ms                   & 43.0\%                    &469$\pm$3.6 ms              & 38.0 \%              & 562$\pm$4.4 ms      & 43.6\%                   \\
\bottomrule
\end{tabular}
  }
    \caption{Comparison of the peak memory and mean and standard error of the per-step latency of the DM Transformer measured on representative hardware from three tiers, \textit{i.e.},~NVIDIA A100 (80 GB VRAM), A6000 (48 GB VRAM) and GTX 3090 24 GB (VRAM).}
  \label{app:tab:latency_full}
\end{table}

\subsection{Additional Qualitative Image Results}
In this subsection, we include image samples generated from \sysname and baselines, to showcase the quality of outputs (see Figures~\ref{app:fig:image_quality} and~\ref{app:fig:image_quality2}).

\section{Limitations \& Societal Impacts}\label{app:limilations}
\noindent\textbf{Technical Limitations:}
While \sysname demonstrates impressive results, several considerations remain. First, our approach requires initial profiling to identify hierarchical importance patterns. Yet, this is a one-time cost that gets amortised when applying the pruning/distillation method repeatedly to DMs with similar architectures. Second, while we have validated our dual-hierarchy insights across multiple MMDiT-based architectures (SD3.5 and FLUX), fundamentally different architectures might exhibit different patterns requiring adaptation.

\noindent\textbf{Broader Impact:} By reducing hardware requirements, our work enables broader access to state-of-the-art generative tools for researchers, artists, and educators with limited computing resources. Efficient inference aligns with sustainable AI practices, potentially lowering the carbon footprint of large-scale generative tasks. However, automation of creative workflows may impact industries reliant on human-generated art. We emphasise collaboration with policymakers to ensure equitable transitions. Moreover, like any research in efficient foundation models, faster and cheaper generation could amplify harmful applications (\textit{e.g.}, spam and deepfake). We advocate for strict ethical guidelines, developer accountability, educational campaigns to raise awareness of generative AI’s capabilities and risks and tools like watermarking to trace AI-generated content.

\newcommand\appImgA{\adjustbox{valign=m,vspace=0pt,margin=0pt}{\includegraphics[width=.33\linewidth]{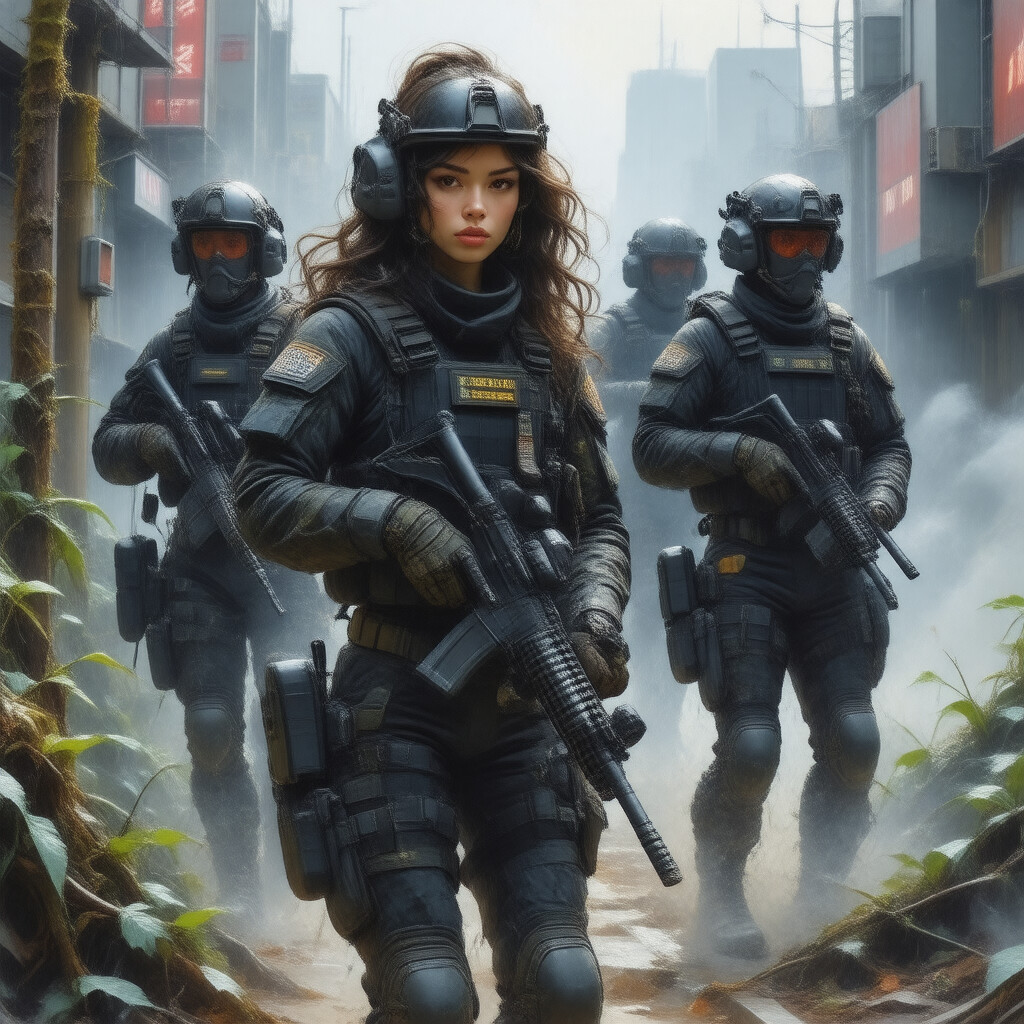}}}
\newcommand\appImgB{\adjustbox{valign=m,vspace=0pt,margin=0pt}{\includegraphics[width=.33\linewidth]{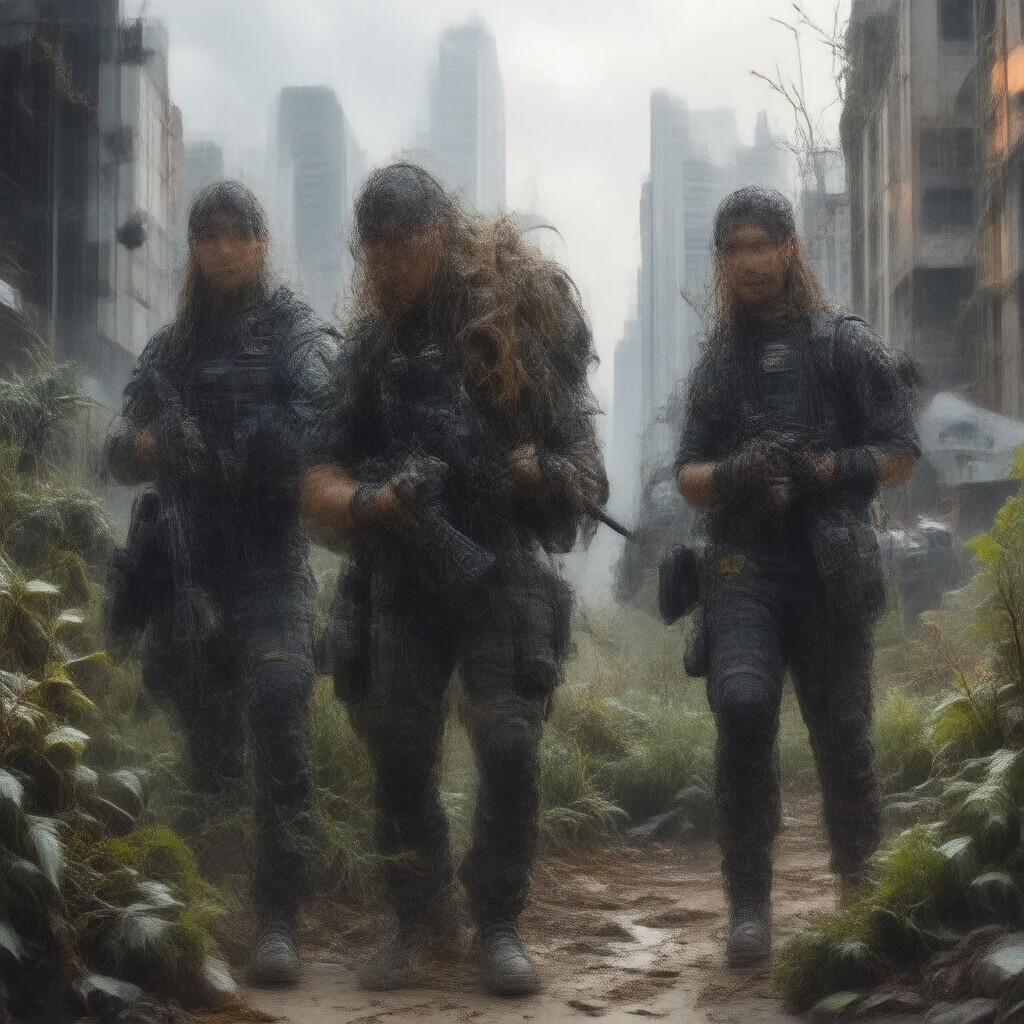}}}
\newcommand\appImgC{\adjustbox{valign=m,vspace=0pt,margin=0pt}{\includegraphics[width=.33\linewidth]{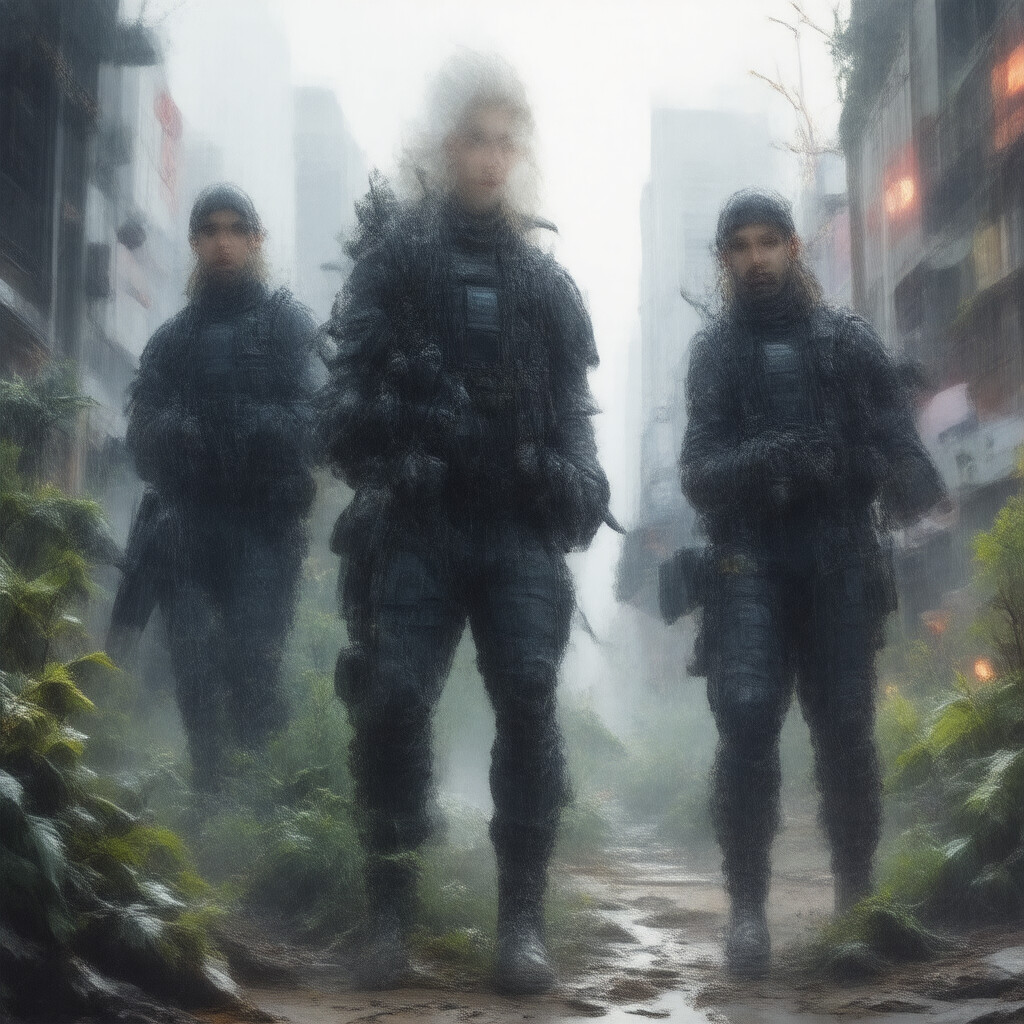}}}
\newcommand\appImgD{\adjustbox{valign=m,vspace=0pt,margin=0pt}{\includegraphics[width=.33\linewidth]{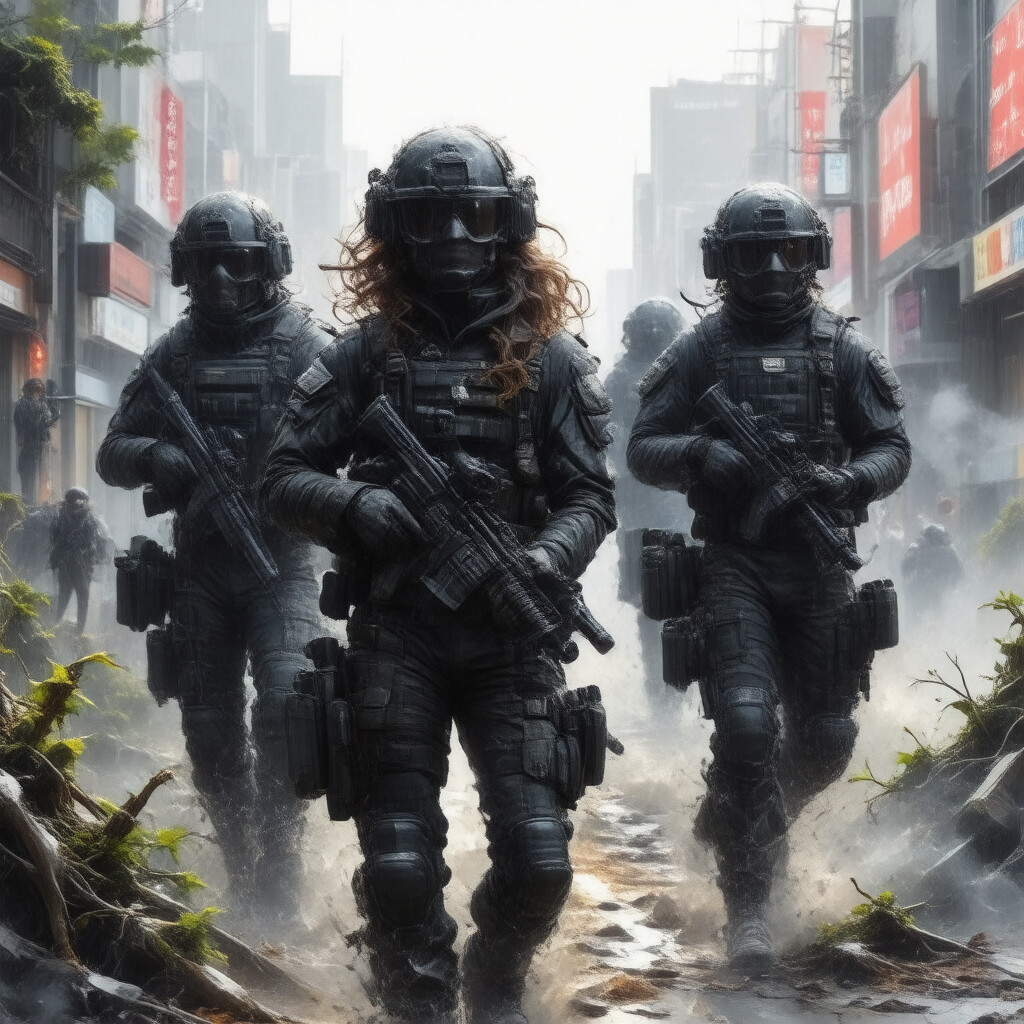}}}
\newcommand\appImgE{\adjustbox{valign=m,vspace=0pt,margin=0pt}{\includegraphics[width=.33\linewidth]{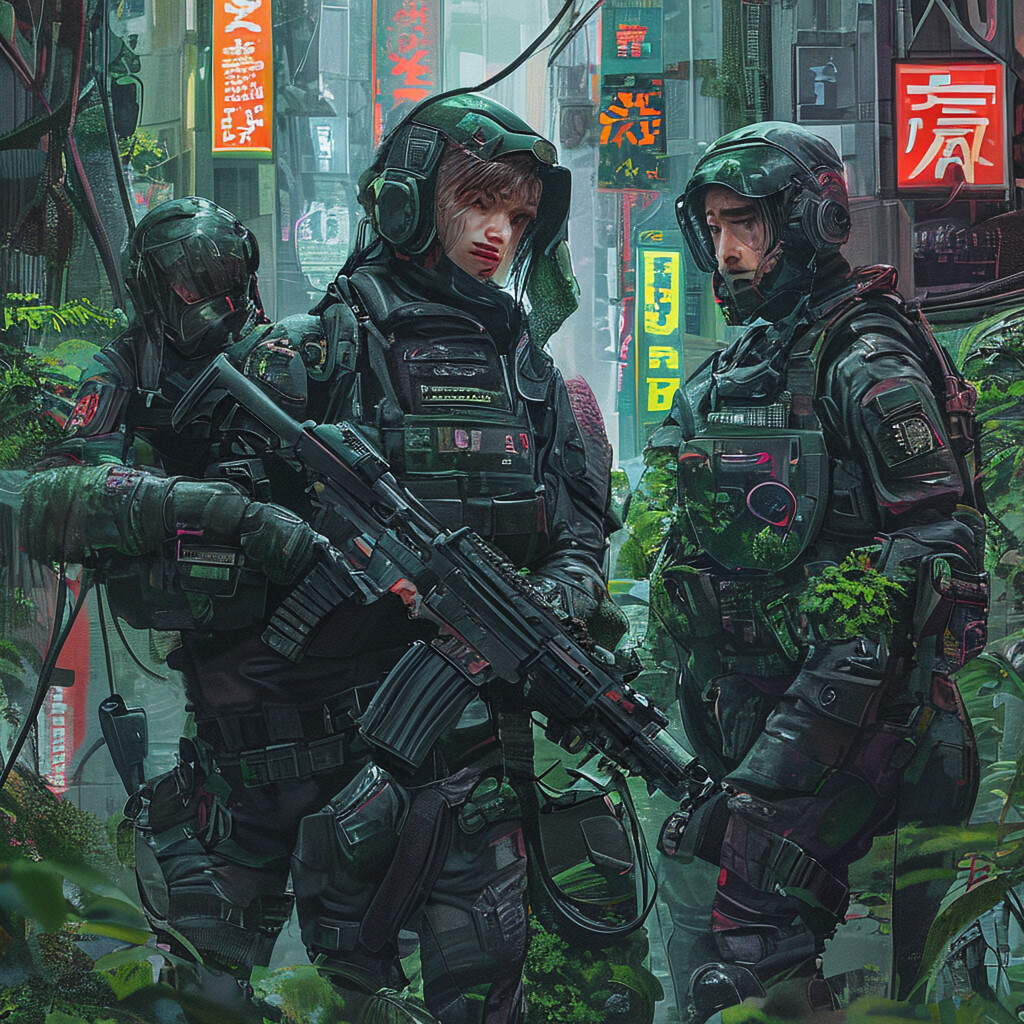}}}

\newcommand\appImgAA{\adjustbox{valign=m,vspace=0pt,margin=0pt}{\includegraphics[width=.33\linewidth]{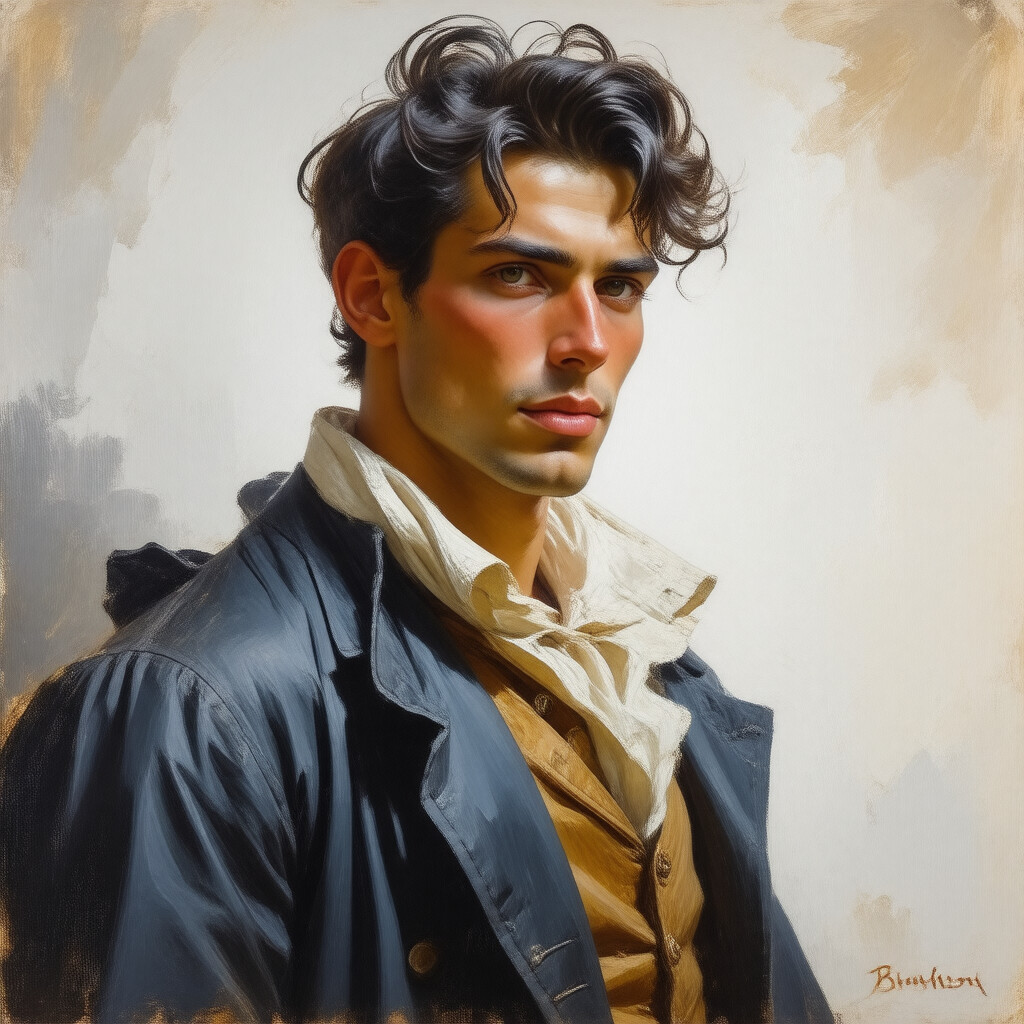}}}
\newcommand\appImgBB{\adjustbox{valign=m,vspace=0pt,margin=0pt}{\includegraphics[width=.33\linewidth]{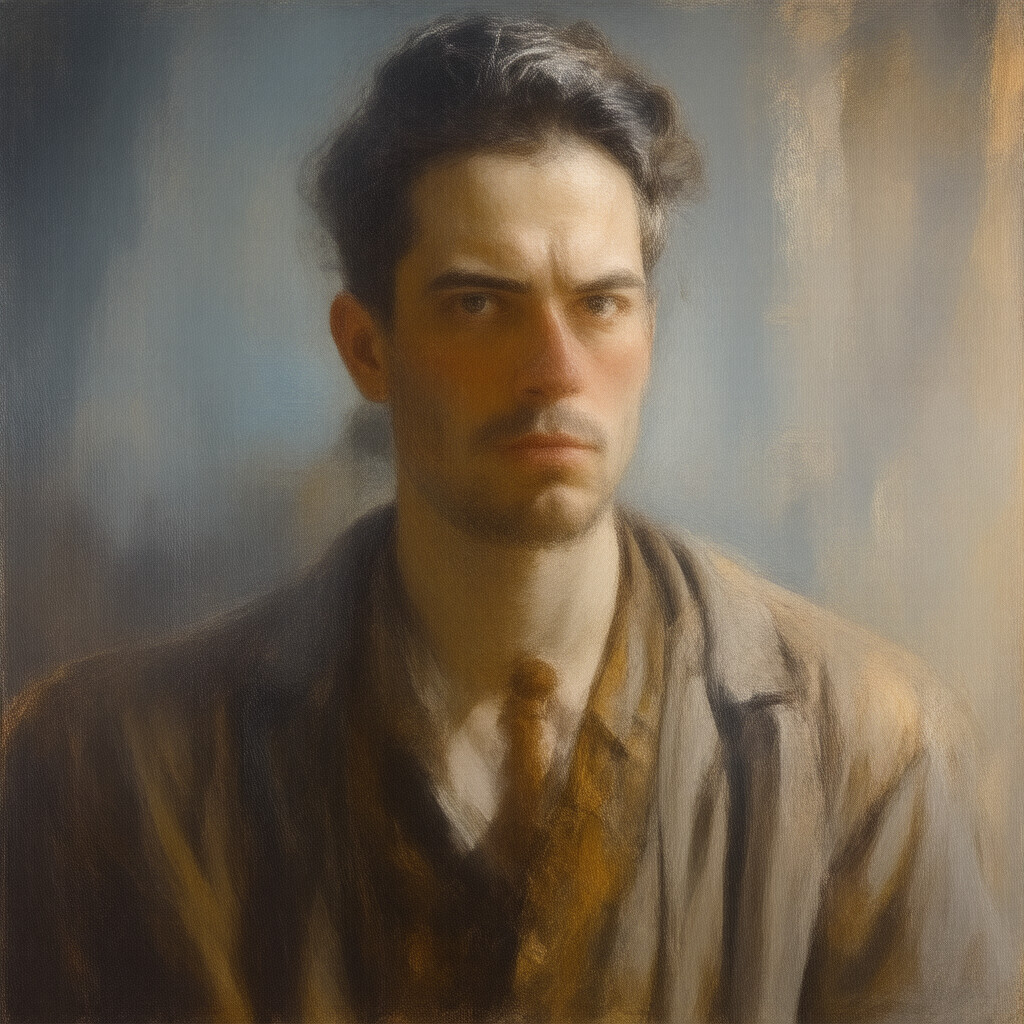}}}
\newcommand\appImgCC{\adjustbox{valign=m,vspace=0pt,margin=0pt}{\includegraphics[width=.33\linewidth]{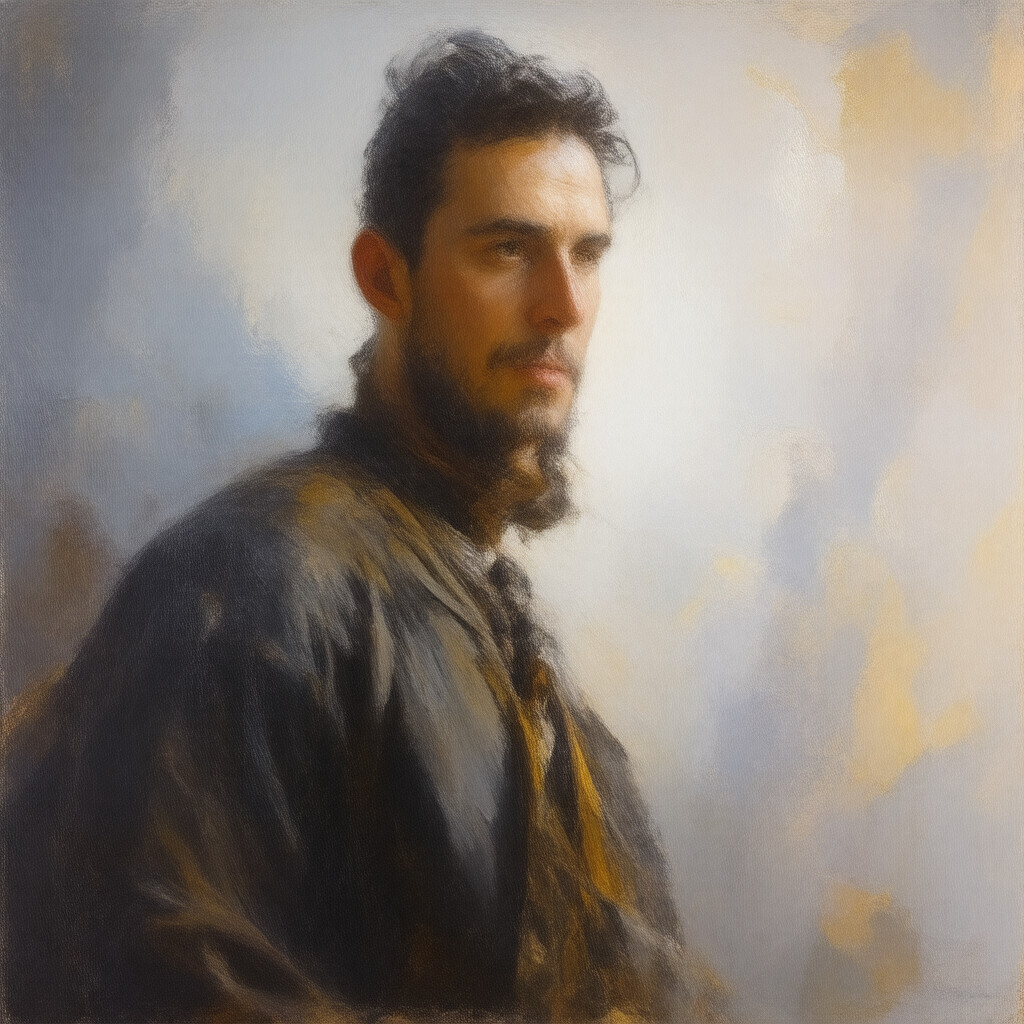}}}
\newcommand\appImgDD{\adjustbox{valign=m,vspace=0pt,margin=0pt}{\includegraphics[width=.33\linewidth]{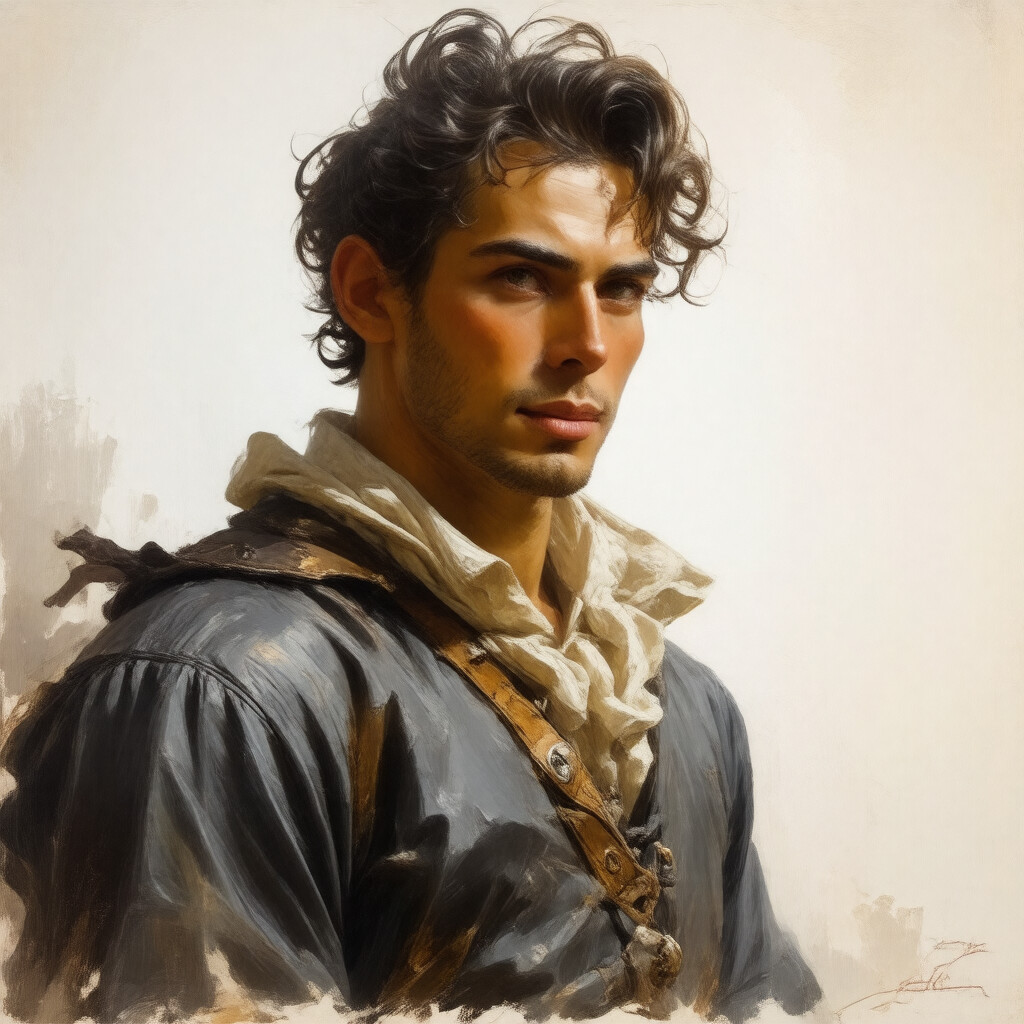}}}
\newcommand\appImgEE{\adjustbox{valign=m,vspace=0pt,margin=0pt}{\includegraphics[width=.33\linewidth]{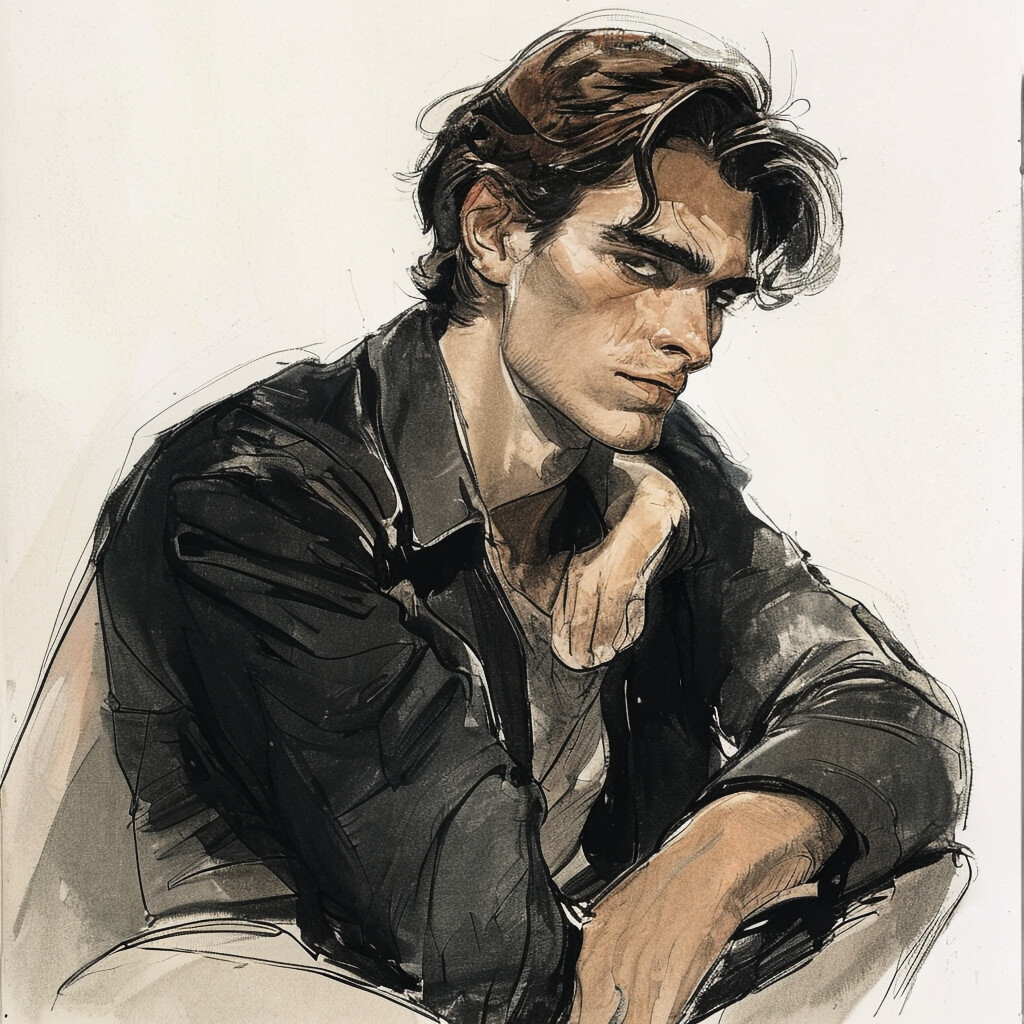}}}

\newcommand\imgA{\adjustbox{valign=m,vspace=0pt,margin=0pt}{\includegraphics[width=.33\linewidth]{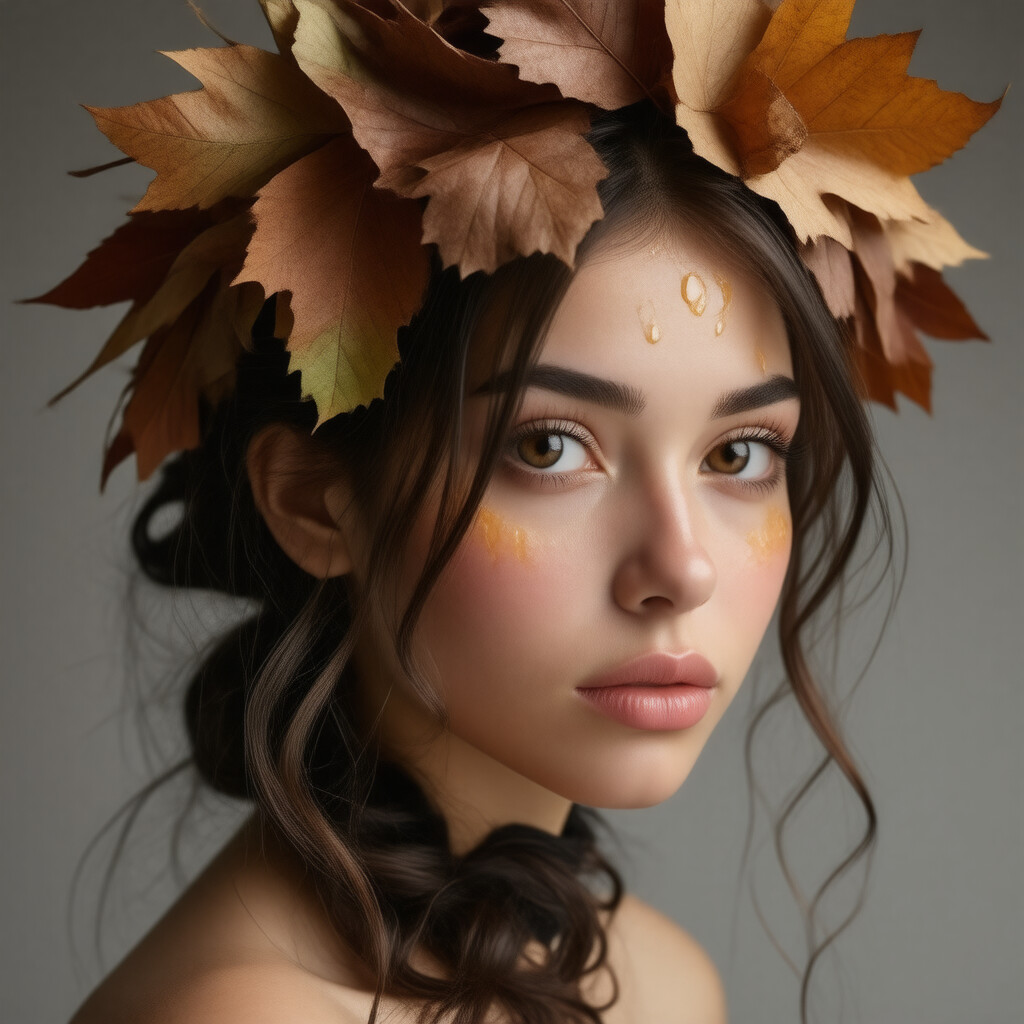}}}
\newcommand\imgB{\adjustbox{valign=m,vspace=0pt,margin=0pt}{\includegraphics[width=.33\linewidth]{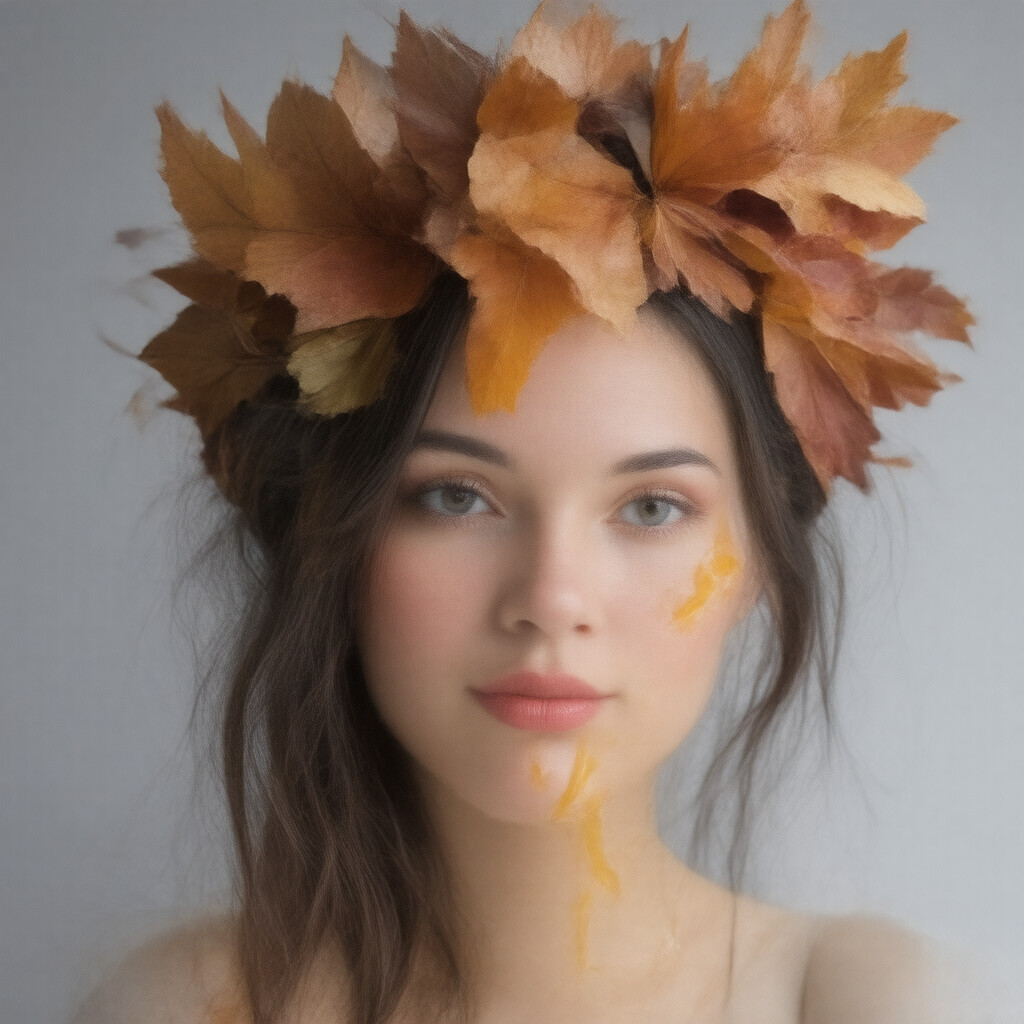}}}
\newcommand\imgC{\adjustbox{valign=m,vspace=0pt,margin=0pt}{\includegraphics[width=.33\linewidth]{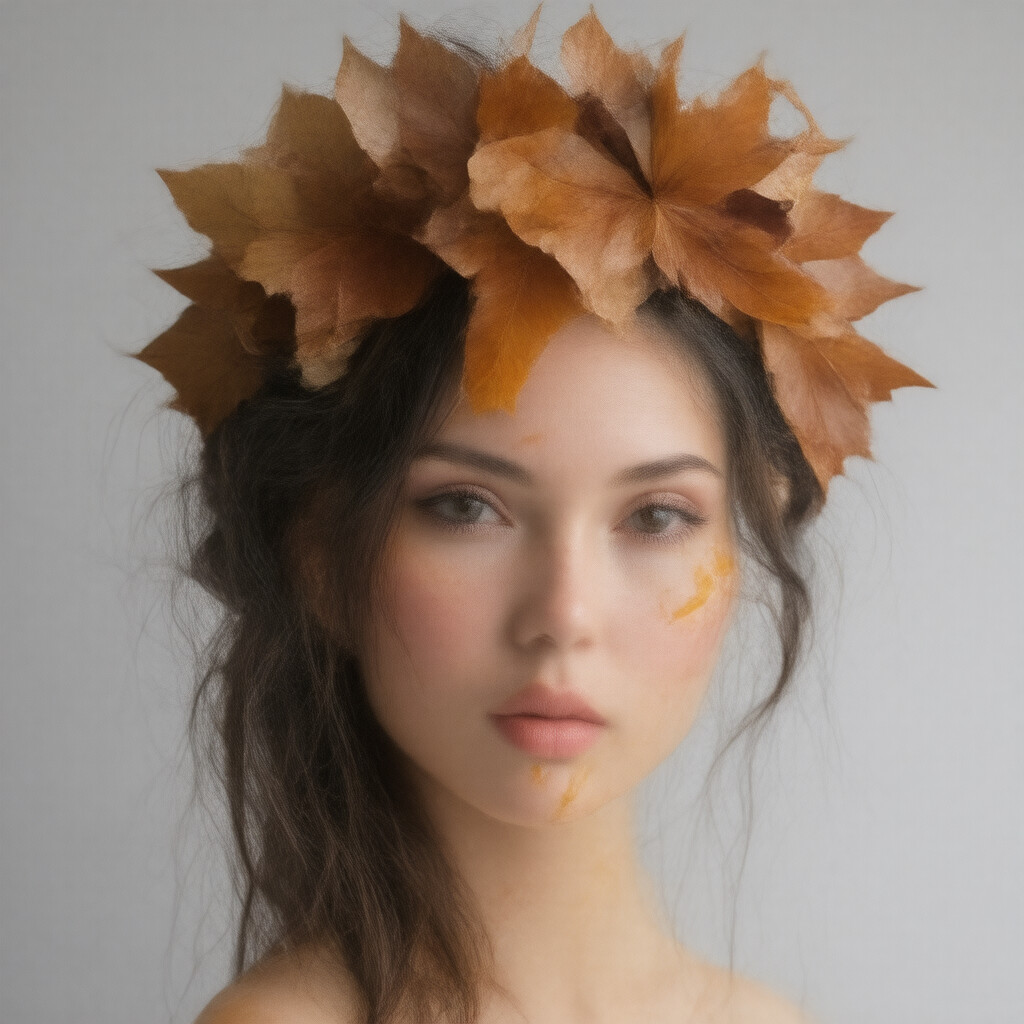}}}
\newcommand\imgD{\adjustbox{valign=m,vspace=0pt,margin=0pt}{\includegraphics[width=.33\linewidth]{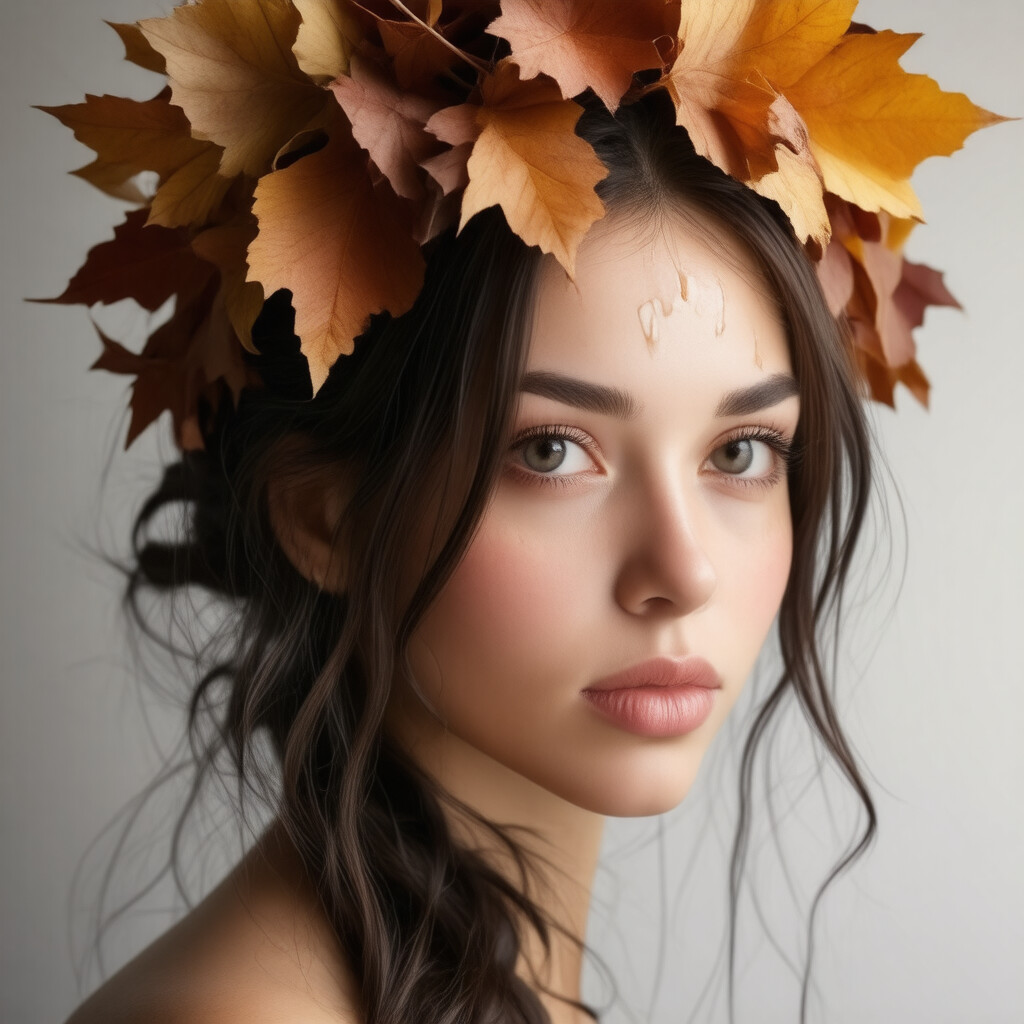}}}
\newcommand\imgE{\adjustbox{valign=m,vspace=0pt,margin=0pt}{\includegraphics[width=.33\linewidth]{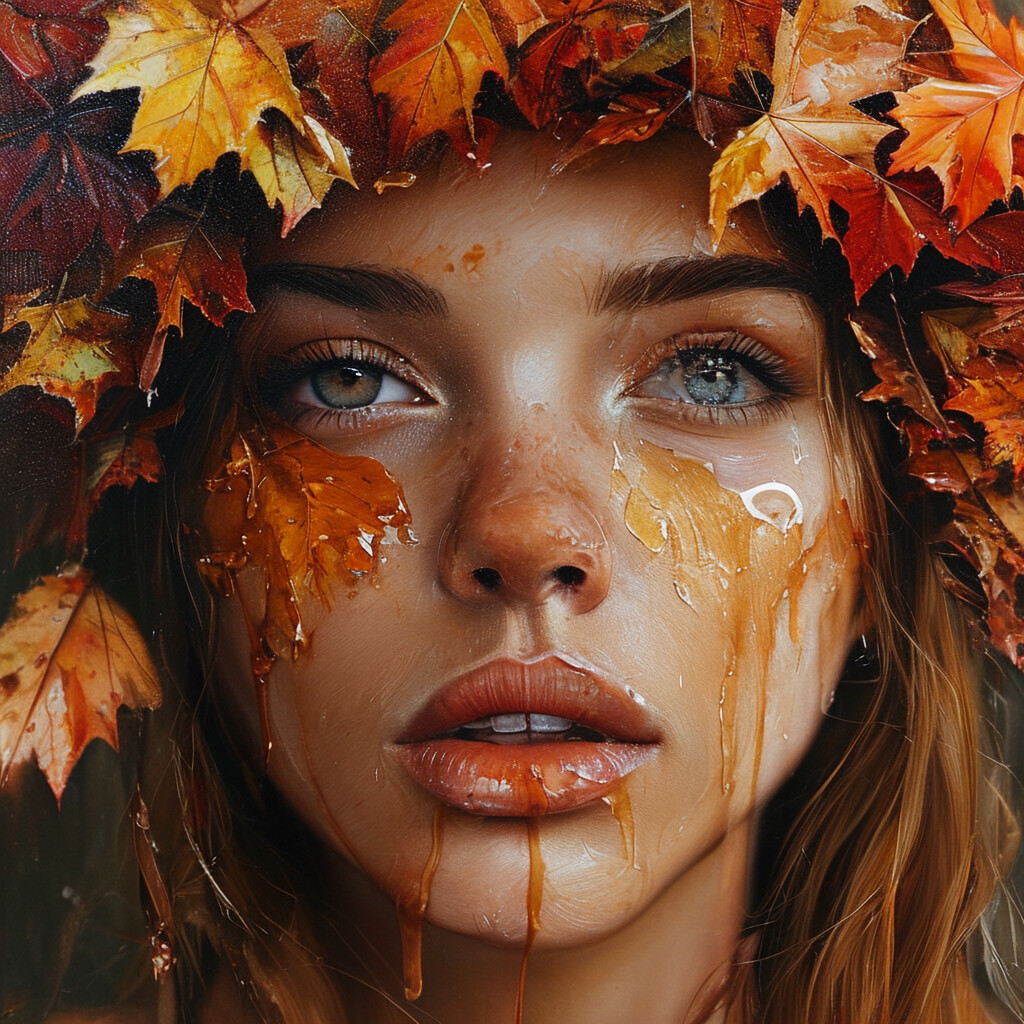}}}

\newcommand\appImgAAAA{\adjustbox{valign=m,vspace=0pt,margin=0pt}{\includegraphics[width=.33\linewidth]{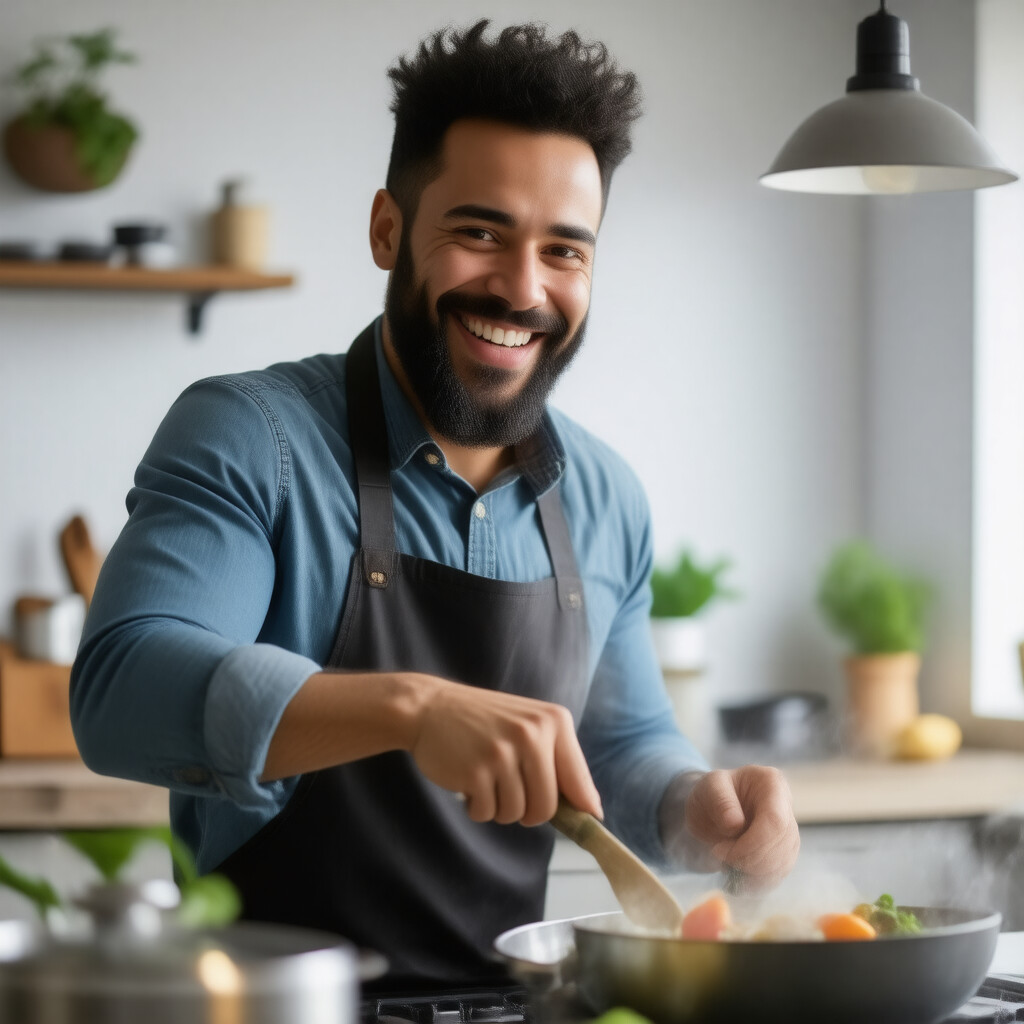}}}
\newcommand\appImgBBBB{\adjustbox{valign=m,vspace=0pt,margin=0pt}{\includegraphics[width=.33\linewidth]{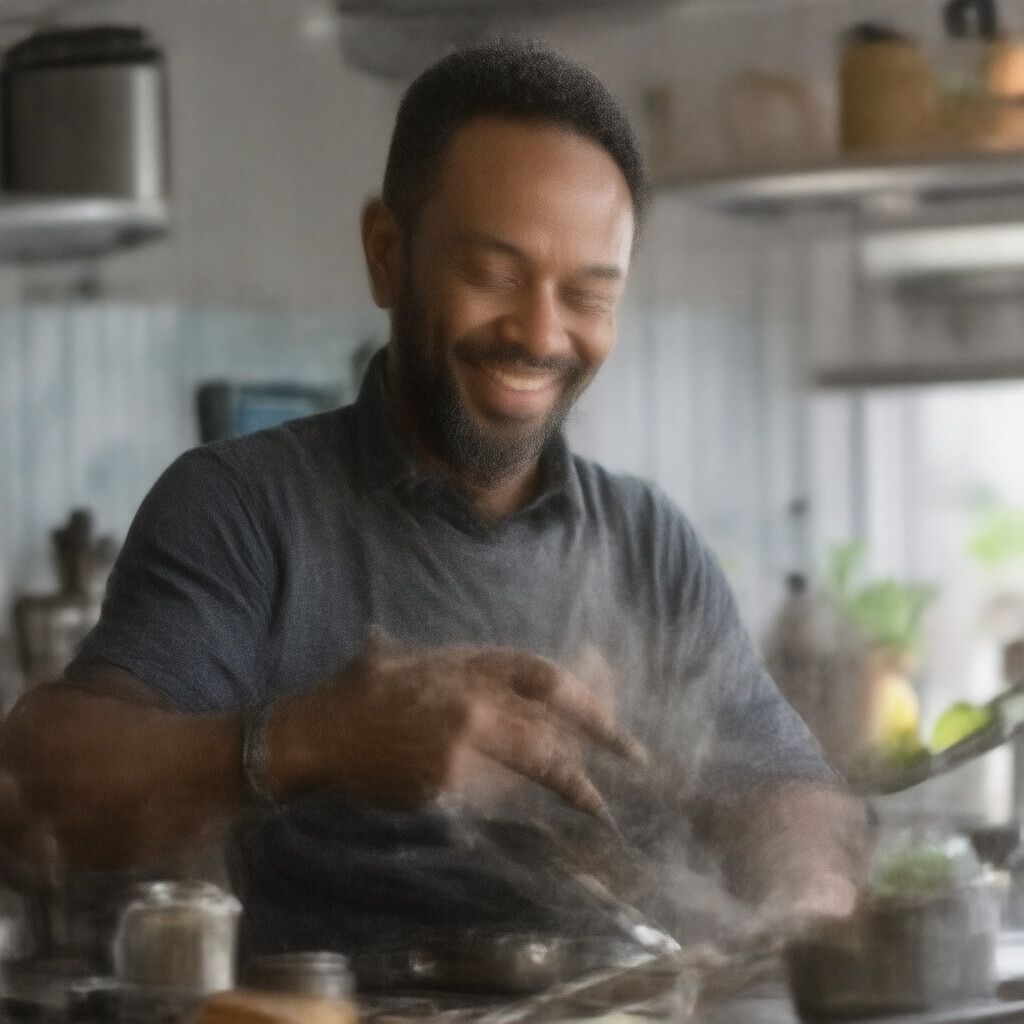}}}
\newcommand\appImgCCCC{\adjustbox{valign=m,vspace=0pt,margin=0pt}{\includegraphics[width=.33\linewidth]{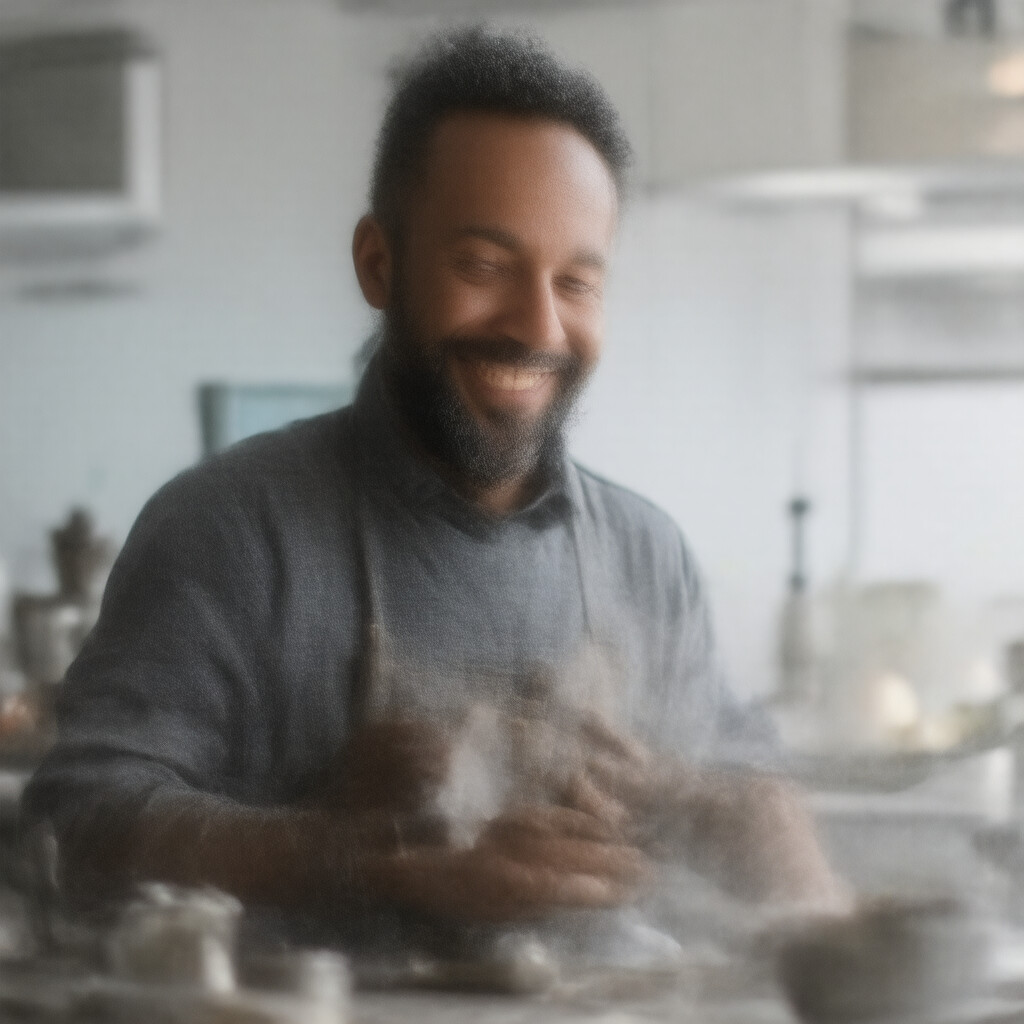}}}
\newcommand\appImgDDDD{\adjustbox{valign=m,vspace=0pt,margin=0pt}{\includegraphics[width=.33\linewidth]{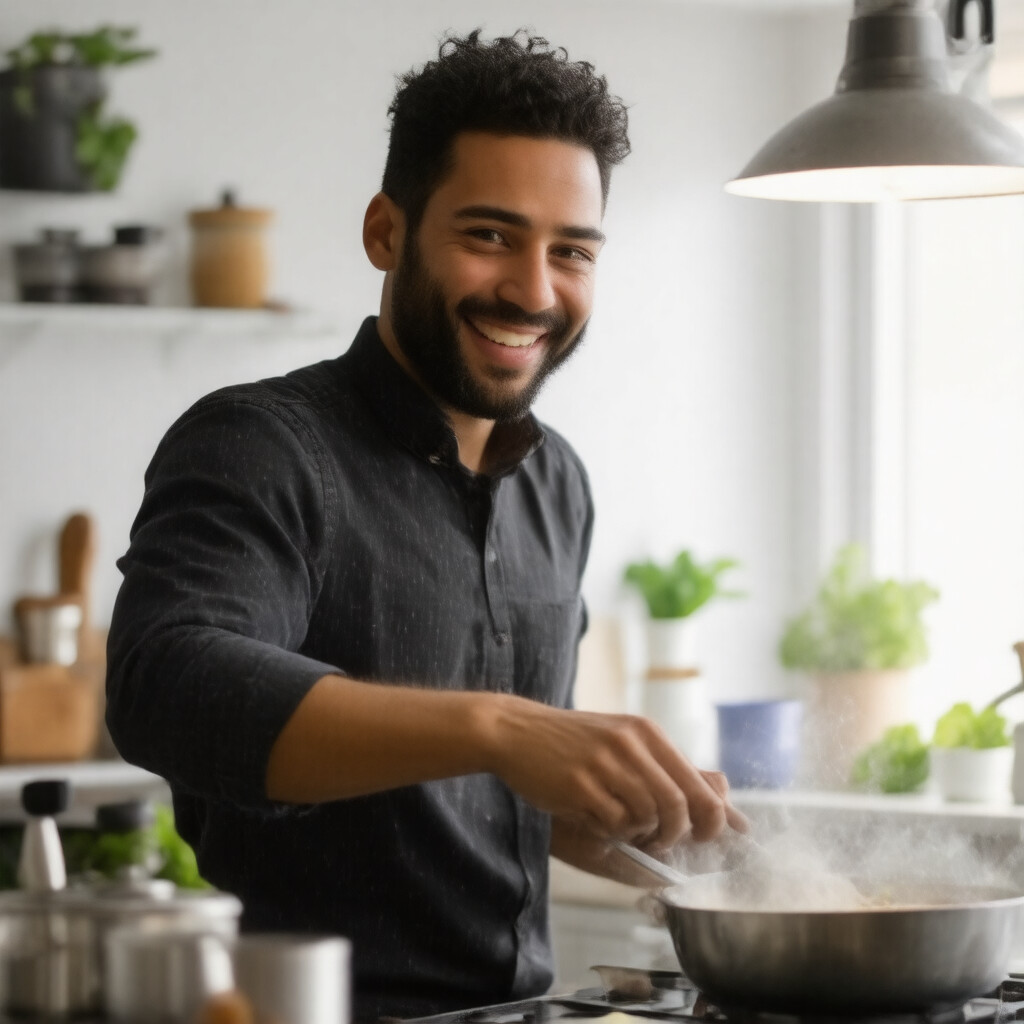}}}
\newcommand\appImgEEEE{\adjustbox{valign=m,vspace=0pt,margin=0pt}{\includegraphics[width=.33\linewidth]{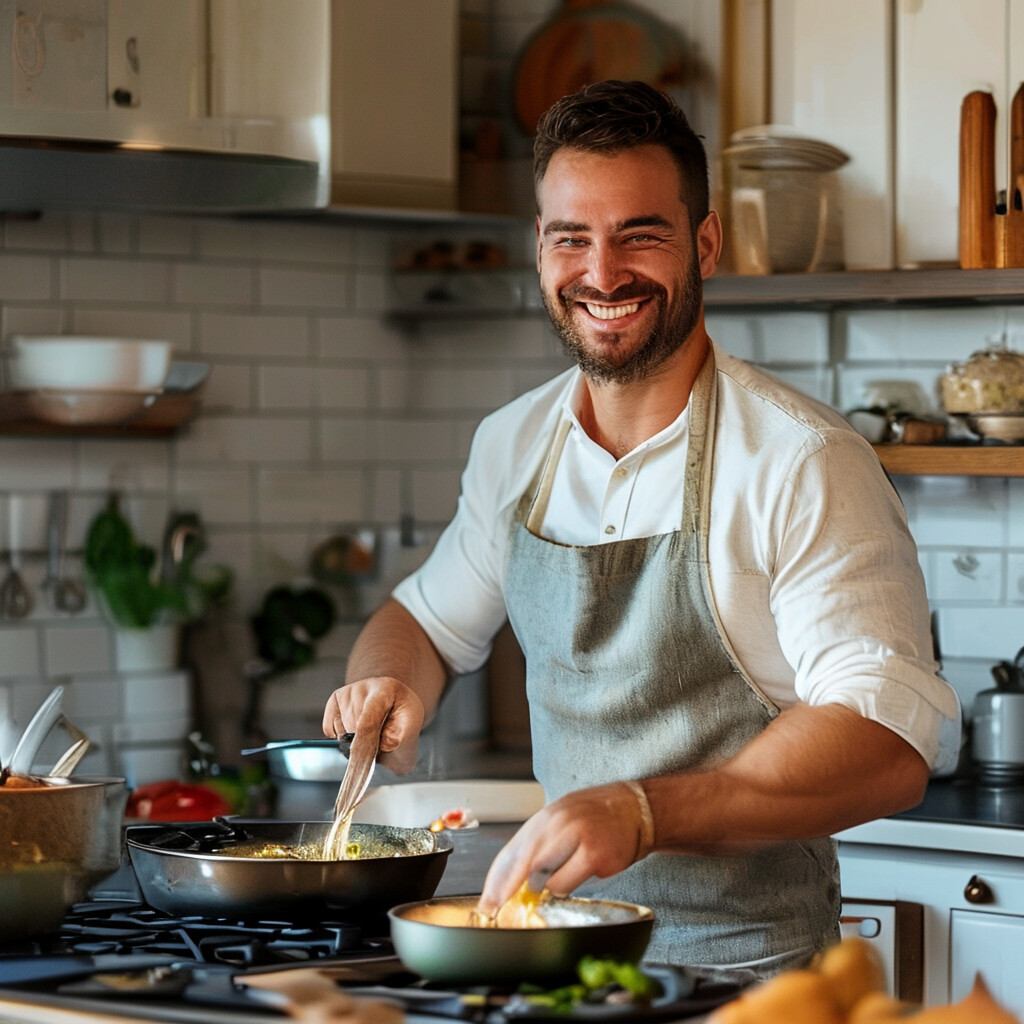}}}

\begin{figure*}[t]
    \centering
    \resizebox{1.01\linewidth}{!}{%
    \begin{tabular}{@{}c| @{\hspace{0.2cm}}c @{\hspace{0.1cm}}c @{\hspace{0.1cm}}c@{} @{\hspace{0.2cm}} |c}
    \LARGE \textbf{Original Model} &\multicolumn{3}{c|}{\LARGE \textbf{Compression Methods}} & \LARGE \textbf{Non-Compression} \\
        \toprule
                \LARGE SD3.5 Large Turbo & \LARGE BK-SDM & \LARGE KOALA  & \LARGE HierarchicalPrune (Ours) & \LARGE SANA-Sprint-1.6B \\
        \appImgA & \appImgB & \appImgC &
        \appImgD & \appImgE \vspace{0.2cm} \\
        \multicolumn{5}{c}{\textit{\LARGE "A digital illustration of a beautiful and alluring American SWAT team in dramatic poses"}} \vspace{0.2cm} \\
        \appImgAA & \appImgBB & \appImgCC &
        \appImgDD & \appImgEE \vspace{0.2cm} \\
        \multicolumn{5}{c}{\textit{\LARGE "Male character illustration by Gaston Bussiere.}} \vspace{0.1cm} \\
        \imgA & \imgB & \imgC &
        \imgD & \imgE \vspace{0.2cm} \\
        \multicolumn{5}{c}{\textit{\LARGE "A close-up portrait of a beautiful girl with an autumn leaves headdress and melting wax."}} \vspace{0.2cm} \\
        \appImgAAAA & \appImgBBBB & \appImgCCCC &
        \appImgDDDD & \appImgEEEE \vspace{0.2cm} \\
        \multicolumn{5}{c}{\textit{\LARGE "A smiling man is cooking in his kitchen."}} \vspace{-0.1cm} \\
    \end{tabular}
    }
    \caption{Visual comparison demonstrating the quality difference between the original model (column~1), depth pruning based on BK-SDM (column~2), KOALA (column~3), and our proposed \sysname (column~4), as well as SOTA small-scale diffusion model, SANA-Sprint-1.6B (column~5). Our approach successfully maintains visual quality while delivering 79.5\% memory reduction over the original model. Notably, our method preserves the text drawing capability of the original SD3.5 Large Turbo model where SANA-Sprint-1.6B is limited.}
    \label{app:fig:image_quality}
\end{figure*}

\newcommand\appImgintroA{\adjustbox{valign=m,vspace=0pt,margin=0pt}{\includegraphics[width=.33\linewidth]{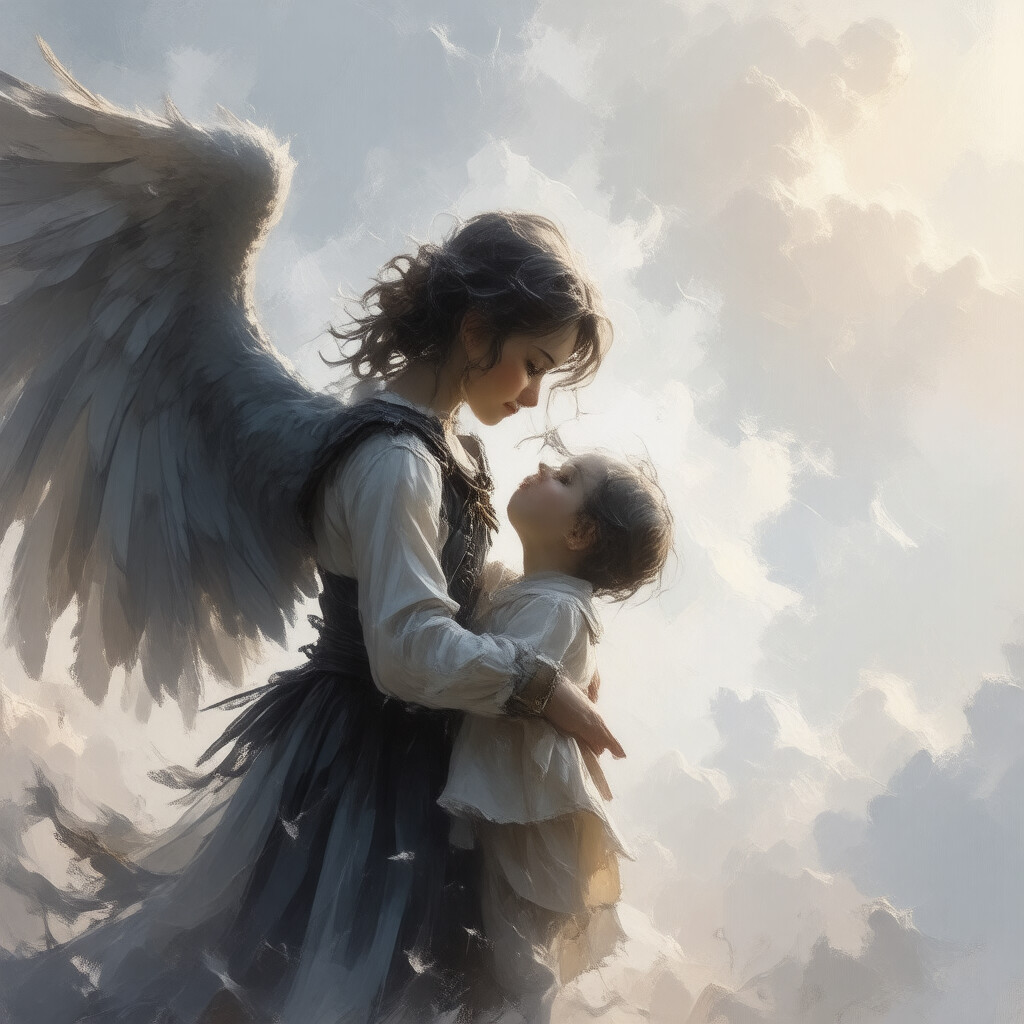}}}
\newcommand\appImgintroB{\adjustbox{valign=m,vspace=0pt,margin=0pt}{\includegraphics[width=.33\linewidth]{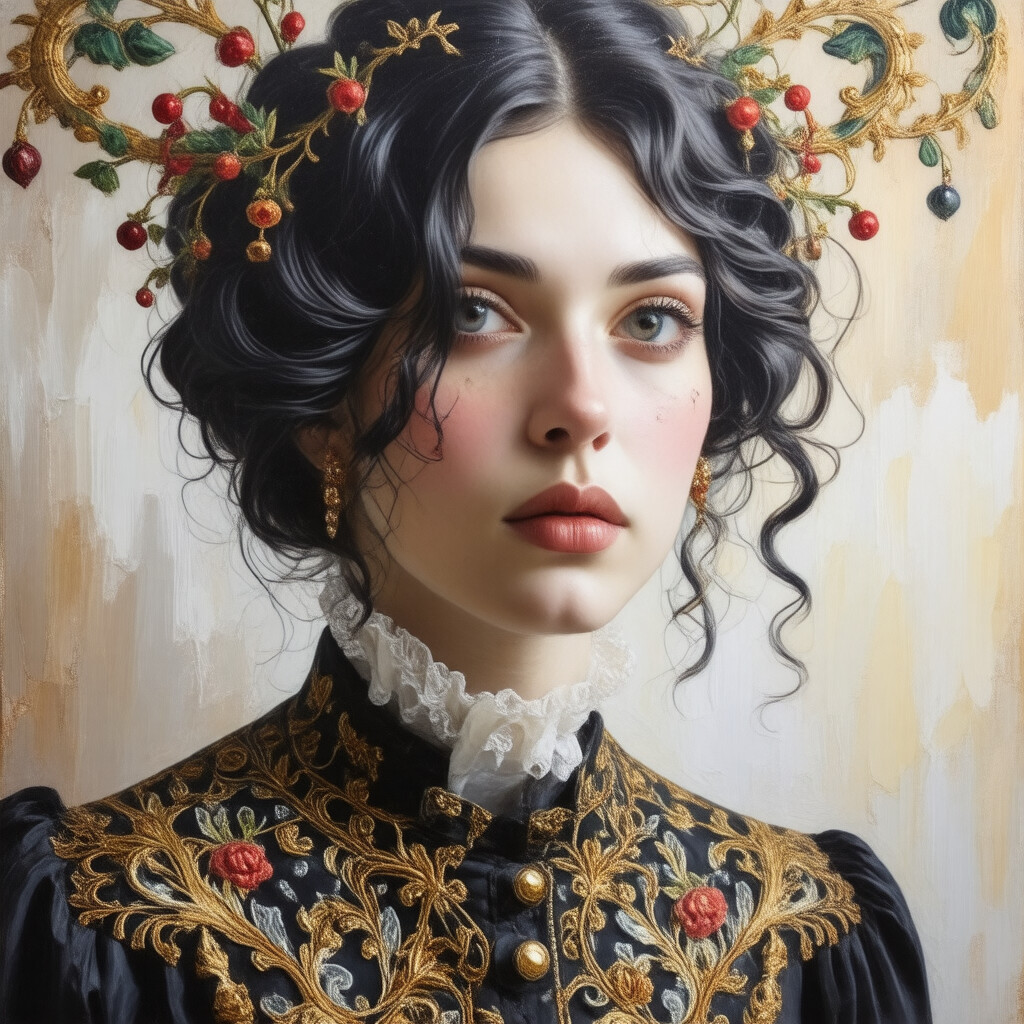}}}
\newcommand\appImgintroC{\adjustbox{valign=m,vspace=0pt,margin=0pt}{\includegraphics[width=.33\linewidth]{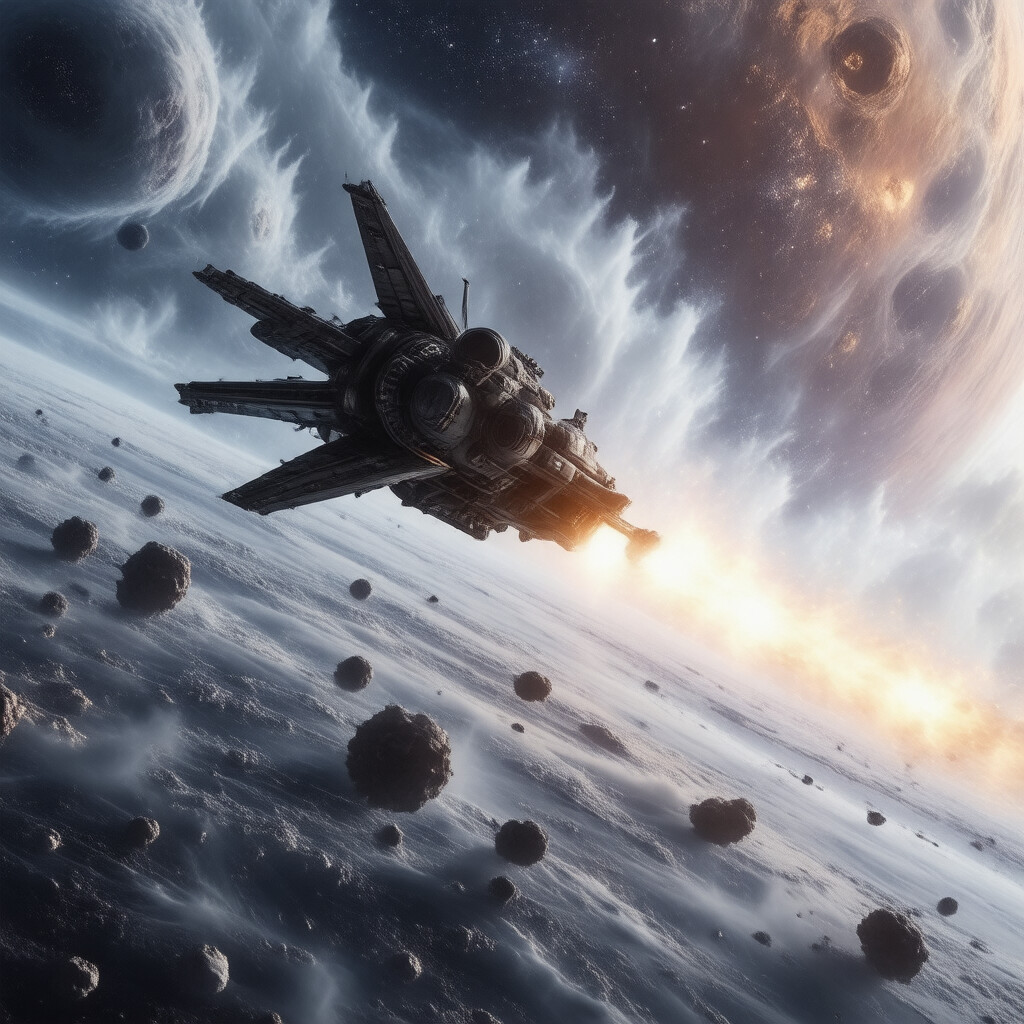}}}
\newcommand\appImgintroD{\adjustbox{valign=m,vspace=0pt,margin=0pt}{\includegraphics[width=.33\linewidth]{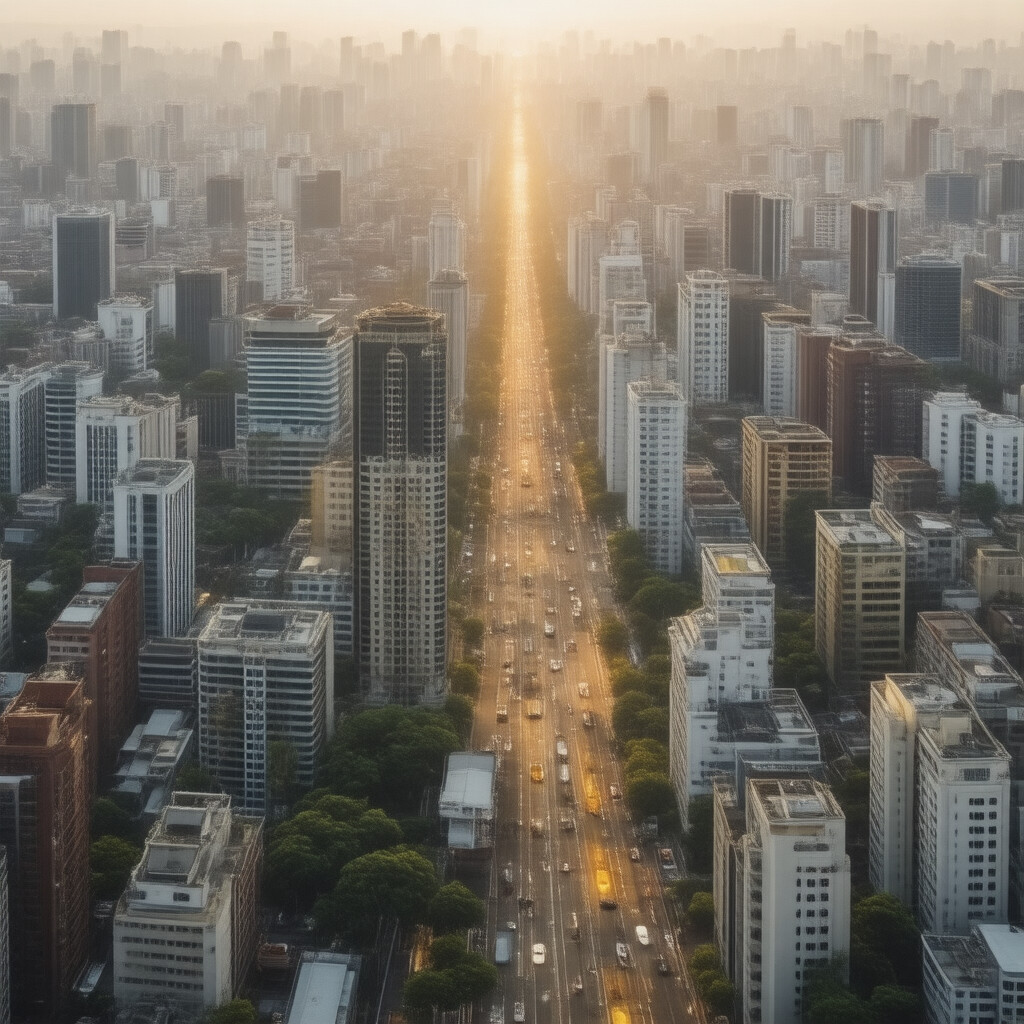}}}

\newcommand\appImgintroAA{\adjustbox{valign=m,vspace=0pt,margin=0pt}{\includegraphics[width=.33\linewidth]{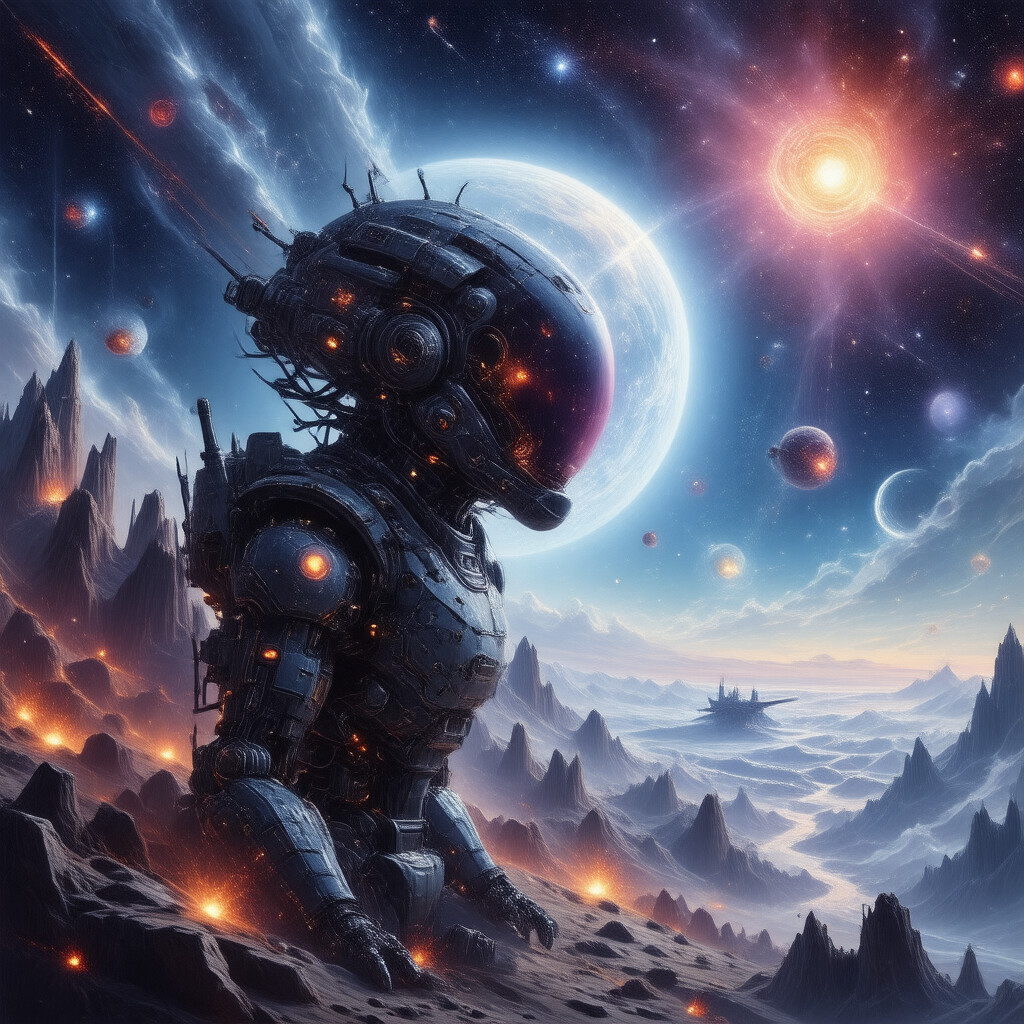}}}
\newcommand\appImgintroBB{\adjustbox{valign=m,vspace=0pt,margin=0pt}{\includegraphics[width=.33\linewidth]{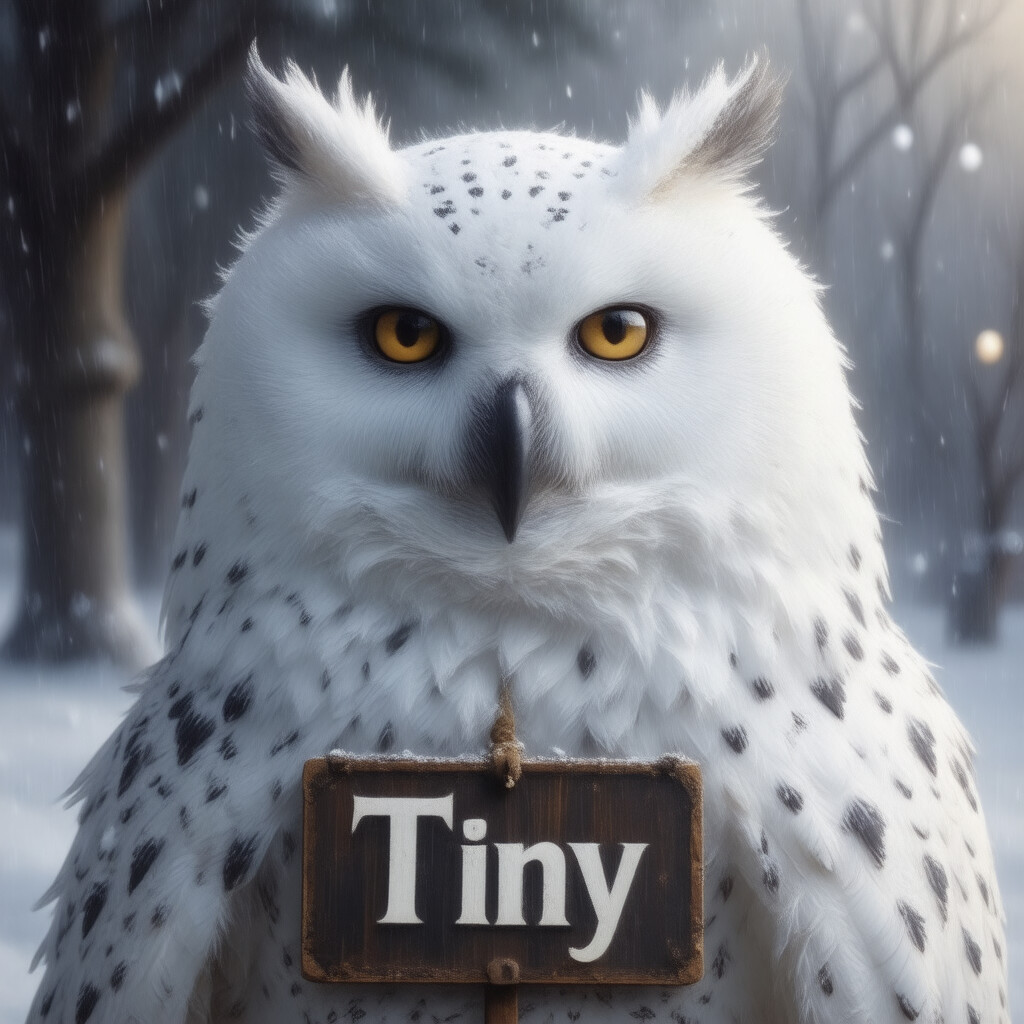}}}
\newcommand\appImgintroCC{\adjustbox{valign=m,vspace=0pt,margin=0pt}{\includegraphics[width=.33\linewidth]{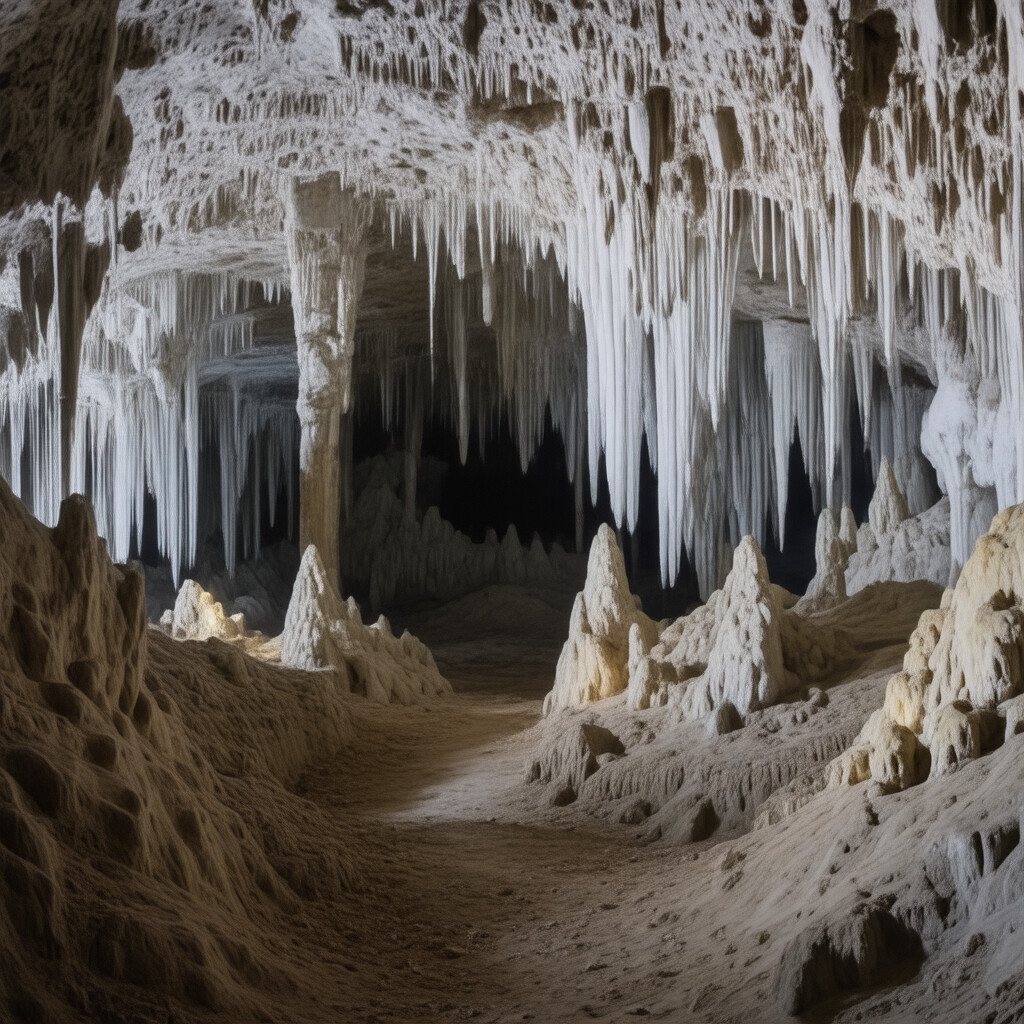}}}
\newcommand\appImgintroDD{\adjustbox{valign=m,vspace=0pt,margin=0pt}{\includegraphics[width=.33\linewidth]{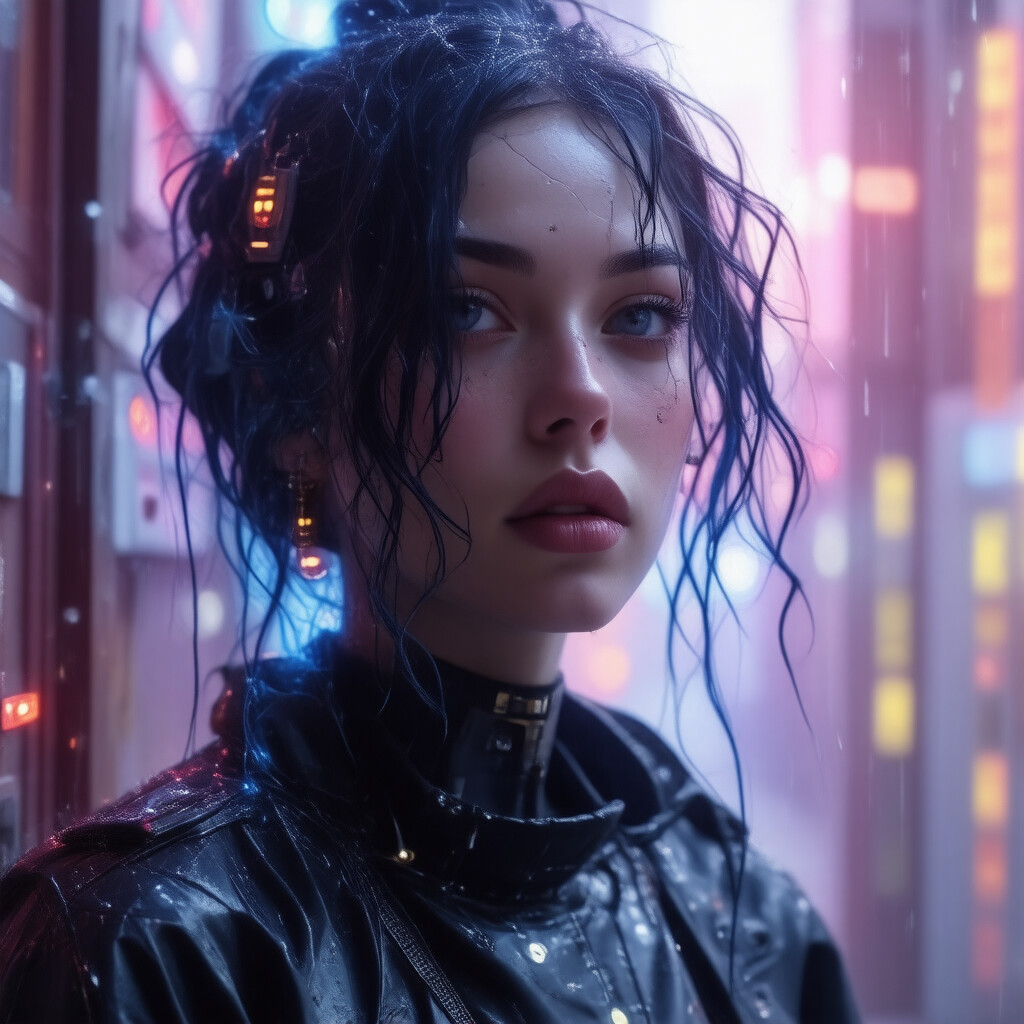}}}

\newcommand\appImgintroE{\adjustbox{valign=m,vspace=0pt,margin=0pt}{\includegraphics[width=.33\linewidth]{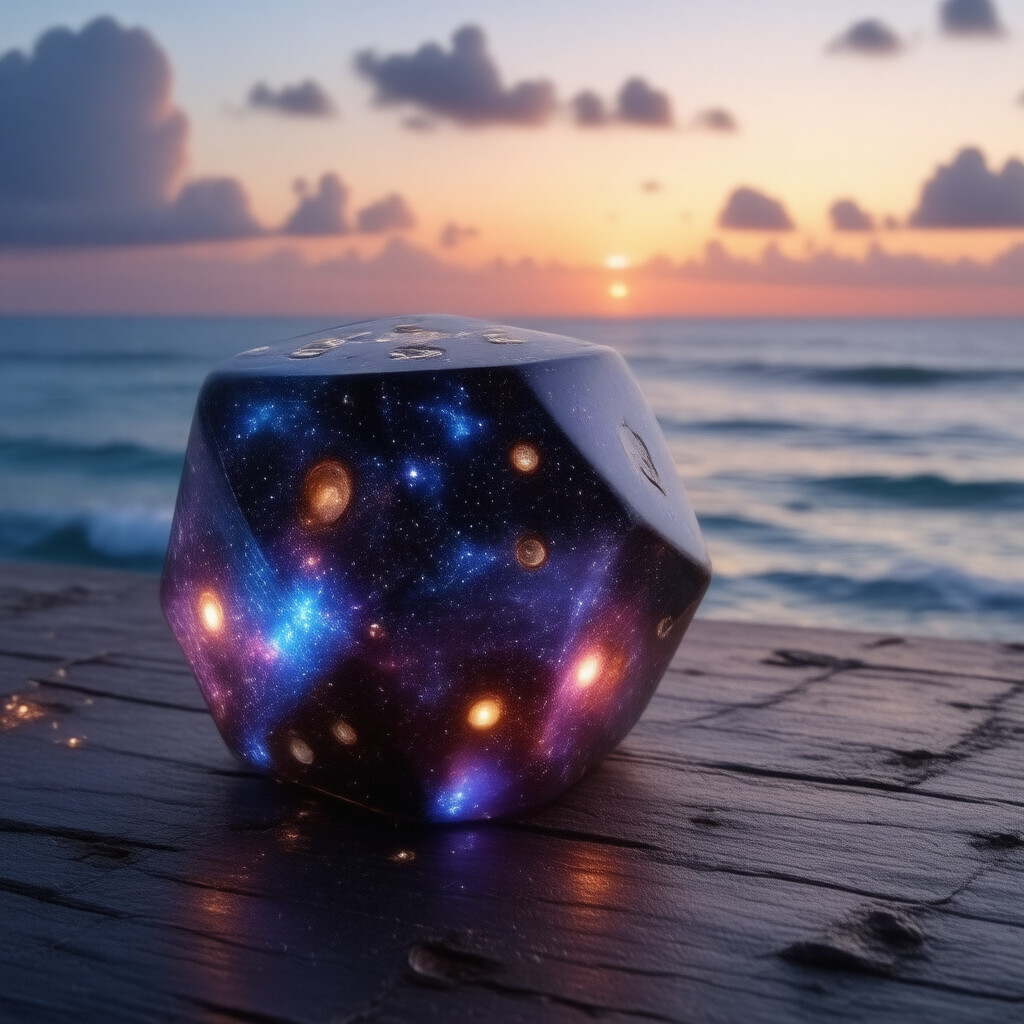}}}

\begin{figure*}[h]
    \centering
    \resizebox{1.01\linewidth}{!}{%
    \begin{tabular}{@{}c @{\hspace{0.05cm}}c @{\hspace{0.05cm}}c@{}}
        \appImgintroA & \appImgintroB & \appImgintroC
         \vspace{0.1cm} \\
        \appImgintroAA & \appImgintroBB & \appImgintroCC
        \vspace{0.1cm} \\
        \appImgintroD & \appImgintroDD  & \appImgintroE \vspace{0.1cm} \\
    \end{tabular}
    }
    \caption{High-resolution image samples generated by our pruned/distilled model using our proposed method, \sysname, showcasing its superior visual quality across various visual styles, precisely following text prompts, and preserving the ability to draw typography.}
    \label{app:fig:image_quality2}
\end{figure*}

\end{document}